\documentclass{article} 
\usepackage{iclr2027_conference,times}

\usepackage{amsmath,amsfonts,bm}

\def\eqref#1{equation~\ref{#1}}

\def\1{\bm{1}}

\DeclareMathAlphabet{\mathsfit}{\encodingdefault}{\sfdefault}{m}{sl}
\SetMathAlphabet{\mathsfit}{bold}{\encodingdefault}{\sfdefault}{bx}{n}

\usepackage{hyperref}
\usepackage{url}
\usepackage{graphicx}

\usepackage[most]{tcolorbox}
\usepackage{enumitem}
\usepackage{xcolor}
\usepackage{multirow}
\usepackage{booktabs}
\usepackage{placeins}
\usepackage{inconsolata}

\definecolor{predefinedcolor}{RGB}{230,245,255}
\definecolor{sequentialcolor}{RGB}{235,250,235}
\definecolor{hierarchicalcolor}{RGB}{255,245,225}
\definecolor{searchcolor}{RGB}{248,235,255}
\definecolor{predefinedframe}{RGB}{50,110,170}
\definecolor{sequentialframe}{RGB}{50,130,60}
\definecolor{hierarchicalframe}{RGB}{190,110,20}
\definecolor{searchframe}{RGB}{120,60,160}
\definecolor{declarationcolor}{RGB}{242,242,242}
\definecolor{declarationframe}{RGB}{90,90,90}

\tcbset{  predefinedprompt/.style={colback=predefinedcolor, colframe=predefinedframe, coltitle=white},  sequentialprompt/.style={colback=sequentialcolor, colframe=sequentialframe, coltitle=white},  hierarchicalprompt/.style={colback=hierarchicalcolor, colframe=hierarchicalframe, coltitle=white},  searchprompt/.style={colback=searchcolor, colframe=searchframe, coltitle=white},  declarationprompt/.style={colback=declarationcolor, colframe=declarationframe, coltitle=white},
} 
\newtcolorbox{promptbox}[2][]{    
enhanced,    
breakable,    
title={#2},    
colback=gray!4,    
colframe=gray!55,    
coltitle=black,    
fonttitle=\bfseries,    
boxrule=0.6pt,    
arc=2pt,    
left=6pt,    
right=6pt,    
top=6pt,    
bottom=6pt,    
before skip=8pt,    
after skip=8pt,    
#1
}

\newtcolorbox{planexample}[2][]{
    enhanced,
    breakable,
    colback=#2,
    colframe=black!45,
    boxrule=0.6pt,
    arc=2pt,
    left=5pt,
    right=5pt,
    top=4pt,
    bottom=4pt,
    before skip=6pt,
    after skip=6pt,
    fonttitle=\bfseries,
    title=#1
}

\usepackage{listings} 

\lstdefinestyle{promptstyle}{
    basicstyle=\ttfamily\scriptsize,
    backgroundcolor=\color{gray!3},
    frame=single,
    rulecolor=\color{gray!40},
    framerule=0.4pt,
    framesep=4pt,
    breaklines=true,
    breakatwhitespace=false,
    columns=fullflexible,
    keepspaces=true,
    showstringspaces=false,
    xleftmargin=2pt,
    xrightmargin=2pt,
    aboveskip=3pt,
    belowskip=7pt
}

\newcommand{\tool}[1]{\texttt{#1}}

\title{Do LLM Agents Execute the Plans They Declare? From Planning-Mode Declaration to Pattern-Specific Execution}

\author{Subba Reddy Oota$^{1}$, Francisco Herrera$^{1,2}$, Jordi Cabot Sagrera$^{1,3}$ \\
\textbf{Marcos López de Prado$^{1,4,5}$, Shadab Khan$^1$} \\
\normalsize  $^1$ADIA Lab, Abu Dhabi, United Arab Emirates, $^2$University of Granada, Granada, Spain \\ $^3$ Luxembourg Institute of Science and Technology, Luxembourg $^4$Cornell University, Ithaca, USA \\ $^5$Lawrence Berkeley National Laboratory, Berkeley, CA\\
  \small \texttt{Subba.Oota@adialab.ae, herrera@decsai.ugr.es, jordi.cabot@list.lu} \\
  \texttt{Marcos.LopezDePrado@adia.ae, Shadab.Khan@adialab.ae} }

\iclrfinalcopy 
\usepackage{etoolbox}

\makeatletter
\patchcmd{\@maketitle}
  {\lhead{Published as a conference paper at ICLR 2027}}
  {\lhead{Preprint}}
  {}{}
\makeatother
\begin{document}

\maketitle

\begin{abstract}

Large language models (LLMs) enable agents to solve long-horizon tasks by generating a plan and then executing it in an environment. However, successful planning requires two distinct capabilities: selecting an appropriate plan for the task and executing it faithfully. Existing planner--executor systems can fail at either stage: the agent may deviate from a structured plan during execution, or it may faithfully execute a plan that is poorly matched to the task or environment. Final task success alone cannot distinguish between these two sources of failure. We therefore ask: can LLM agents be trusted to execute the plans they commit to, and do different tasks and environments benefit from different planning modes? To study this, we introduce a diagnostic framework for the \emph{Plan Declaration--Execution Gap} and propose \textsc{Planning-as-Routing}, in which the LLM declares one of four planning modes: \emph{Predefined, Sequential, Hierarchical, or Search}. A deterministic router sends the task to the corresponding pattern-specific executor. Across four benchmarks and three LLMs, we find three consistent patterns. First, generic Plan+ReAct often fails to preserve the declared planning structure, especially for longer plans: across three benchmarks, only \(22\)–\(45\%\) of trajectories preserve it, whereas pattern-specific executors enforce the intended structure. Second, planning-mode effectiveness varies across environments and models: Search performs best on ALFWorld, Hierarchical on SWE-bench, and the strongest pattern can vary across models within the same benchmark. Third, the largest gains come from plan execution: pattern-specific executors raise task success from \(0.48\) to \(0.92\) on ALFWorld and from \(0.36\) to \(0.44\) on SWE-bench Verified over Plan+ReAct. However, current LLMs do not reliably select the strongest planning mode for a task, while few-shot examples can improve planning-mode selection for some benchmark--model combinations. Overall, reliable agent planning requires both selecting an effective planning mode and executing it with a matching executor: routing substantially closes the execution gap, while selecting the right mode for each task remains open. \footnote{Project page: \url{https://subbareddy248.github.io/projects/declaration-gap-site/}}

\end{abstract}

\section{Introduction}

The rapid advancement of large language models (LLMs) has accelerated the development of interactive agents~\cite{liu2025advances,wang2024survey}, which are designed to solve complex real-world tasks through multi-turn interactions with external environments such as web browsing~\cite{zhou2024webarena,deng2023mind2web}, computer use~\cite{xie2024osworld,merrill2026terminal}, and embodied task execution~\cite{shridharalfworld}. To solve such tasks, an agent often needs to decompose a high-level goal into structured steps and perform the necessary actions; planning provides this structure by turning an objective into a course of action that guides the agent toward the goal~\cite{yao2023react}. Consequently, planning has become a central component of both agentic frameworks~\cite{shen2023hugginggpt,webb2025brain} and world-model-based approaches that simulate or reason about possible future states before acting~\cite{qiao2024agent,maes2026leworldmodel,wang2026adajepa}. This structure is what sustains goal-directed behavior in complex, multi-step tasks, where later decisions depend on the outcomes of earlier actions. However, generating a coherent plan is not sufficient for reliable agent behavior. A plan can fail at two levels. At the execution level, the agent may declare one plan but deviate from it during execution. At the selection level, the declared plan itself may be poorly matched to the task: even faithful execution can fail if the chosen plan does not fit the environment. \emph{This raises two fundamental questions: when an agent proposes a plan for solving a task, does it execute the plan it commits to, and is the proposed plan appropriate for the task?}


Existing agent evaluation harnesses offer limited insight into how agents plan and execute their tasks, because they focus on final task success~\cite{liu2024agentbench,zhou2024webarena,mialon2024gaia,xie2024osworld,pan2024webcanvas}. Final task success does not reveal whether an agent followed a deliberate plan or reached the goal through an inefficient trajectory, memorization, or chance~\cite{liu2026plan,sun2026agent}. Recent work has begun to evaluate plan quality and whether agents follow instructed plans~\cite{sun2026agent,liu2026plan,jia2025your}. However, these studies largely focus on diagnosing planning behavior rather than jointly examining whether the selected planning mode is appropriate for the task and preserved during execution. They also treat a plan as a sequence of steps to be generated or followed, rather than as a choice among different planning modes. Planning requirements differ across tasks and environments: some tasks can be solved with a fixed plan, whereas others require adaptation, decomposition, or exploration. In this work, we study whether LLM agents select a planning mode for a task, whether that mode is preserved during execution, and whether matching modes to corresponding executors improves task success.




\begin{figure*}[t]
    \centering
    \begin{minipage}{0.49\textwidth}
        \centering
        \includegraphics[width=\linewidth]{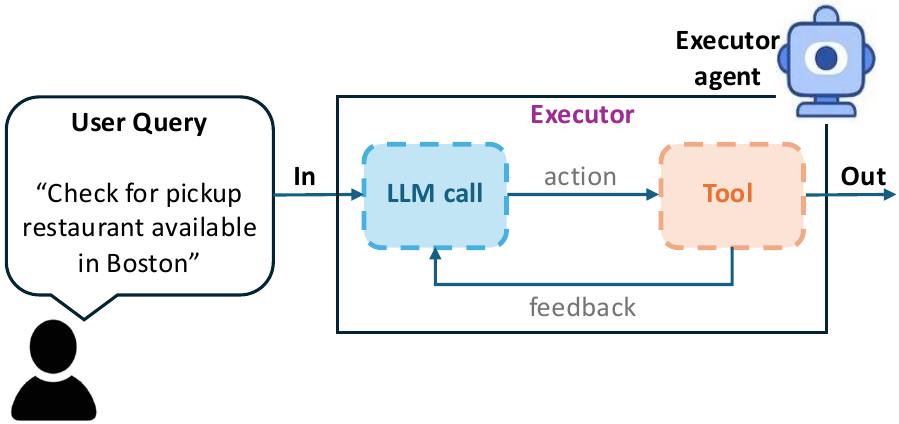} \\
        (a) Flat ReAct
    \end{minipage}
    \hfill
    \begin{minipage}{0.49\textwidth}
        \centering
        \includegraphics[width=\linewidth]{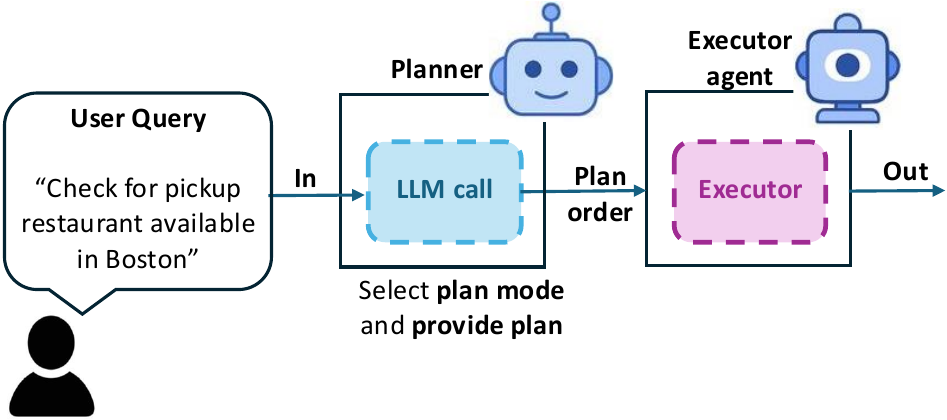} \\
        (b) Planning in the Prompt + ReAct
    \end{minipage}
    \hfill
    \begin{minipage}{\textwidth}
        \centering
        \includegraphics[width=\linewidth]{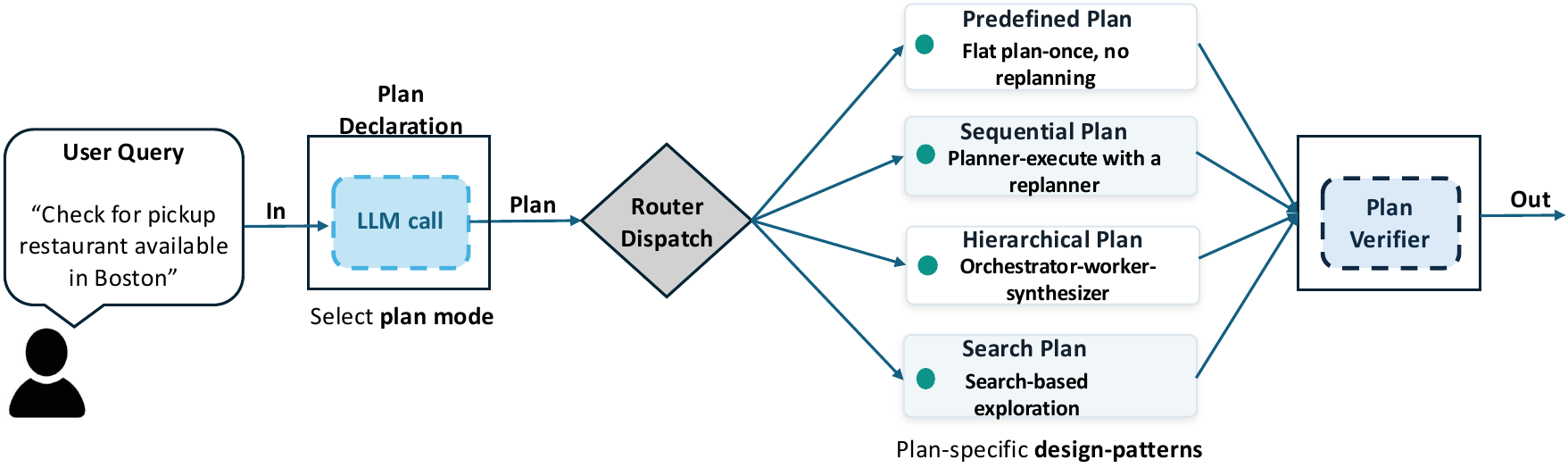} \\
        (c) Planning-as-Routing
    \end{minipage}
    \caption{Overview of the three execution conditions.
(a) \textbf{Flat ReAct}: a single ReAct-style executor selects actions from environment feedback without an explicit plan.
(b) \textbf{Planner--Executor (Plan+ReAct)}: a planner first selects a planning mode and produces a plan, which is then passed to a generic ReAct executor; however, the declared plan is not structurally enforced during execution. 
(c) \textbf{Planning-as-Routing}: the LLM first declares a planning mode, and a deterministic router dispatches the task to a matching pattern-specific executor: predefined, sequential, hierarchical, or search. A verifier checks whether execution preserves the declared plan structure.
}
    \label{fig:workflow_agents}
\end{figure*}

A useful perspective on this problem comes from human planning: humans do not rely on a single fixed strategy for all tasks. A familiar task calls for a routine plan, a structured but unfamiliar one for step-by-step planning, a multi-stage one for decomposition into subgoals, and an uncertain one for weighing alternative routes before acting~\cite{daw2005uncertainty,daw2011model,botvinick2008hierarchical,mattar2018prioritized,mattar2022planning}. This flexibility suggests that planning in LLM agents should not be treated as a uniform behavior. We call a mismatch between the declared plan and its execution the \emph{Plan Declaration--Execution Gap}. This distinction helps distinguish failures of planning-mode selection from failures of plan execution. This motivates the following research questions:

\begin{itemize}
    \item \textbf{RQ1:} As agents take on increasingly long-horizon and multi-step tasks, can we rely on them to execute the planning approach they commit to, or do they fall back to reacting one step at a time?
    \item \textbf{RQ2:} Do different environments and models benefit from different planning approaches, or is a single planning approach sufficient across web, coding, and embodied tasks?
    \item \textbf{RQ3:} When an agent declares a planning approach, does executing it through the corresponding planning pattern improve task success compared with passing the same plan to a generic ReAct executor?
    \item \textbf{RQ4:} When multiple planning approaches are available, can an LLM choose the one that is most appropriate for the task it is trying to solve?
\end{itemize}

Together, these questions clarify whether agent failures arise from choosing the wrong planning mode or from failing to execute the chosen mode faithfully.
To address these questions, we systematically study how LLM agents select and execute different planning approaches, and whether matching a declared approach to its corresponding execution pattern improves task performance. For the purpose of this work, we evaluate agents built on multiple LLM families (Qwen3.6~\cite{qwen3.6-35b-a3b}, DeepSeek-V4~\cite{deepseek2026v4}, Gemma-4-26B~\cite{Abd2026Gemma4T}) across four benchmarks spanning diverse and complex environments: web browsing (WebArena~\cite{zhou2024webarena}, Mind2Web~\cite{deng2023mind2web}), software engineering (SWE-Bench Verified~\cite{jimenez2024swe}), and embodied navigation (ALFWorld~\cite{shridharalfworld}). 
We compare three conditions that differ by one factor at a time: (i) a no-planning Flat ReAct baseline, (ii) a plan-prompted Flat ReAct baseline (Plan+ReAct), and (iii) our Planning-as-Routing framework with pattern-specific executors. Figure~\ref{fig:workflow_agents} illustrates these three conditions.

Our contributions are threefold: 
(1) We introduce a diagnostic framework for measuring the \emph{Plan Declaration--Execution Gap} in LLM agents, and show that a ReAct agent told to plan a certain way often does not follow the declared plan, instead reacting locally to intermediate observations, a divergence invisible to final-success metrics. Our analysis goes beyond task success by measuring plan quality, plan adherence, and structural faithfulness, while separately examining whether the declared planning approach is effective for the task.
(2) We introduce \emph{Planning-as-Routing}, an architecture in which the
LLM declares a planning mode and a deterministic router sends the task to
the corresponding pattern-specific executor: \{predefined, sequential,
hierarchical, or search\}. This design makes the declared planning structure explicit in execution and allows its behavior to be verified from the resulting trajectory.
(3) Across web navigation, software engineering, and embodied tasks, and across multiple LLM families, we show that planning pattern effectiveness
varies across environments and models, while planning-mode selection remains a key bottleneck: current LLM declarations trail the best fixed planning pattern on every benchmark--model pair. Few-shot demonstrations improve task success through better planning-mode selection, with gains of $+0.004$ to $+0.16$ across benchmarks. We will release the code, baseline and planning trajectories across seeds upon publication of this paper.

\vspace{-0.1cm}
\section{Related Work}
\vspace{-0.1cm}


\noindent\textbf{LLMs as Planning Agents.}
Prior work has proposed many planning mechanisms for LLM agents, but these methods often instantiate planning behavior through a specific agent architecture. Earlier approaches used ReAct-style reasoning and acting~\cite{yao2023react}, search over reasoning paths~\cite{yao2023tree}, symbolic planning~\cite{liu2023llm+}, planner--executor architectures~\cite{erdoganplan}, multi-agent workflows~\cite{shen2023hugginggpt,hong2024metagpt}, and adaptive replanning~\cite{liu2026adaplanbench,dong2026pear,wu2026planner}. These approaches can generate and revise task-specific plans, but the underlying planning mechanism is typically fixed by the agent framework. Our work instead treats the planning mode as a per-task choice and asks whether the declared mode is preserved during execution.

\noindent\textbf{Agent Evaluation and Trajectory Diagnosis.}
Agent benchmarks span web navigation~\cite{deng2023mind2web,zhou2024webarena,pan2024webcanvas}, computer use~\cite{xie2024osworld,merrill2026terminal}, software engineering~\cite{jimenez2024swe}, embodied tasks~\cite{shridharalfworld}, and general assistants~\cite{mialon2024gaia}, but primarily evaluate final task outcomes. Recent work moves toward process-level evaluation through progress metrics, plan-compliance analysis, trajectory diagnosis, and failure taxonomies~\cite{ma2024agentboard,liu2026plan,ou2025agentdiagnose,kong2025aegis,cemrimulti}. Our work complements these approaches by explicitly modeling the planner--executor handoff, allowing us to distinguish failures of planning-mode selection from failures to preserve the declared mode during execution.
Discussion of planning architectures and process-level agent evaluation is provided in Appendix~\ref{app:additional-related-work}.

\section{Methodology}

\subsection{Planning Patterns.}
We consider four planning patterns inspired by common forms of human planning~\citep{mattar2018prioritized,mattar2022planning}: \textit{predefined}, \textit{sequential}, \textit{hierarchical}, and \textit{search}. These patterns capture different ways an agent can organize task execution. A \textit{predefined} plan follows a fixed plan generated before execution, without replanning. A \textit{sequential} plan executes one step at a time and updates the remaining plan based on intermediate observations. A \textit{hierarchical} plan decomposes the task into subgoals and coordinates their execution through an orchestrator--worker structure. A \textit{search} plan generates multiple candidate plans, executes each candidate independently, and selects the most promising resulting trajectory using a \emph{rubric-based judge}. The workflow of each planning pattern is illustrated in Appendix~\ref{app:planning_patterns}, Figs.~\ref{fig:planning_patterns} and~\ref{fig:search_planning_patterns}. Representative plan structures and execution traces for all four planning modes are provided in Appendix~\ref{app:planning-examples}.

\subsection{Problem Formulation.}
We study planning under three conditions that share the same backbone LLM and action space, differing only in how planning is produced and executed.

\noindent\textbf{Condition 1: Flat ReAct (no explicit plan declaration).} The LLM receives only the task description and goal, and acts through a standard Flat ReAct loop, one step at a time. No planning strategy is explicitly declared before execution. This condition serves as our implicit-planning baseline, where any planning behavior must emerge locally through the interaction loop.

\noindent\textbf{Condition 2: Plan+ReAct (declared but unenforced).} The LLM first selects a planning approach and produces a task-specific plan. The plan is then provided to the same generic ReAct executor used in Condition 1. Because the executor retains a flat step-by-step control loop, the requested planning structure is available as context but is not structurally enforced. This condition tests whether prompting alone is sufficient for the intended planning approach to appear during execution.

\noindent\textit{Condition 3: Planning-as-Routing.} 
The LLM first declares one planning mode \vspace{-0.4em}
\[
P \in \{\textit{predefined},\textit{sequential},
\textit{hierarchical},\textit{search}\}.
\]  A deterministic router maps the declaration to the corresponding pattern-specific executor. Unlike Condition 2, the selected planning structure therefore determines the execution control flow.

\noindent\textbf{Planning-as-Routing components.} 

\noindent\textit{(1) Plan declaration.} Given a task, the backbone LLM selects one planning mode (P). We evaluate this declaration step under both \emph{zero-shot} and \emph{few-shot} settings. In the \emph{zero-shot setting}, the model selects a mode from the task description alone. In the \emph{few-shot setting}, the declaration prompt additionally includes example tasks paired with planning modes, allowing us to test whether demonstrations improve task-conditioned mode selection.

\noindent\textit{(2) Router.} A deterministic mapping dispatches $P$ to its corresponding executor, with no additional model inference.

\noindent\textit{(3) Pattern-specific executors.} Each executor imposes a distinct planning and control-flow pattern, following established agent-planning architectures. The \textit{predefined} executor generates a complete plan before execution and follows it without replanning, similar to plan-then-solve approaches~\cite{wang2023plan}. The \textit{sequential} executor follows a planner--executor--replanner loop, where the agent executes the current step, observes the outcome, and revises the remaining plan when necessary~\cite{sun2023adaplanner}. The \textit{hierarchical} executor follows an orchestrator--worker structure, where an orchestrator decomposes the task into sub-goals, delegates them to specialized workers, and aggregates their outputs~\cite{zhang2025agentorchestra,choi2025reactree}. Finally, the \textit{search} executor generates multiple candidate plans,
executes each independently, and selects the most promising trajectory using a rubric-based judge, following prior search-based agent planning approaches~\citep{zhou2023language}.


\textit{(4) Execution-structure verifier.} A rule-based verifier checks whether a Plan+ReAct trajectory executes the declared plan steps in order. We first clean the declared steps by removing non-actionable text, then match each remaining step to the corresponding trajectory actions using benchmark-specific rules. Structure is maintained only when all scorable steps are matched in the declared order, while allowing extra actions between them. Routed runs are not scored this way because their executor dispatch records establish order fidelity by construction. Implementation details and representative matching examples are provided in
Appendix~\ref{app:verifier}.

\subsection{Planning Process Metrics}
\label{sec:planning-metrics}

Following prior work~\cite{jia2025your}, we evaluate \textit{plan adherence} and \textit{plan quality}, together with \textit{plan-order faithfulness}. \textit{Plan adherence} measures whether the executed actions complete the declared plan steps. \textit{Plan quality} measures whether the generated plan is appropriate for the task goal and environment. \textit{Plan-order faithfulness} measures whether the plan steps or subgoals are executed in their declared order. \textit{Plan quality} and \textit{plan adherence} are empirical metrics, whereas plan-order faithfulness is a structural verification for predefined, sequential, and hierarchical; it is not applicable to search, where candidates represent competing alternatives rather than an ordered sequence.

\subsection{Pattern-Ceiling Analysis}
\label{sec:ceiling}

Task success alone cannot distinguish whether an executor is incapable of solving a task or whether the declaration module selected an unsuitable planning pattern. To separate execution limitations from planning-mode selection, we use \emph{forced dispatch}: for each benchmark--model pair, every planning mode is executed on every task, bypassing the declaration module. This yields a task--mode success matrix \(m_{i,p}\), where \(m_{i,p}=1\) if planning mode (p) solves task (i), and (0) otherwise. From this matrix, we compute each fixed mode's success \(S(p)\), the best fixed-mode performance \(S^\star=\max_p S(p)\), and a per-task oracle ceiling \(\mathrm{DSR}_{\max}=\frac{1}{N}\sum_i\max_p m_{i,p}\). Their difference \(H=\mathrm{DSR}_{\max}-S^\star\) measures the available routing headroom.

For a declaration policy \(\pi\), task success is computed as \(\mathrm{TSR}(\pi)=\frac{1}{N}\sum_i m_{i,\pi(i)}\), using the same forced-dispatch matrix without re-running executors. Together with the trajectory verifier, this separates \emph{execution failure} from \emph{selection failure}: whether the declared mode is preserved during execution versus whether the selected mode is effective for the task. Additional controls and estimation details are provided in Appendix~\ref{app:ceiling-analysis}.

\section{Experimental Setup}

\noindent\textbf{Environments.} We evaluate our agent planning strategies across four agent benchmarks spanning web navigation (Mind2Web~\cite{deng2023mind2web}, WebArena~\cite{zhou2024webarena}), software engineering (SWE-bench Verified~\cite{jimenez2024swe}), and embodied tasks (ALFWorld~\cite{shridharalfworld}). We additionally group tasks using each benchmark's available task categories to examine whether planning-pattern preferences vary with task type; category definitions and counts are provided in Appendix~\ref{app:dataset-details} Table~\ref{tab:benchmark-composition}.


\noindent\textbf{Language Models.} We evaluate three backbone LLMs: Qwen3.6-35B-A3B~\cite{qwen3.6-35b-a3b}, DeepSeek-V4-Flash~\cite{deepseek2026v4}, and Gemma-4-26B~\cite{Abd2026Gemma4T}. Model details are provided in Appendix~\ref{app:dataset-details} Table~\ref{tab:model-details}. Each model is evaluated under the same task inputs, tool interfaces, action spaces, and execution budgets across all conditions. The exact environment-step and planning-structure budgets are reported in
Appendix~\ref{app:inference-config}, Table~\ref{tab:execution-budgets}.
The same backbone model is used for plan declaration and execution unless otherwise specified. 

\noindent\textbf{Repeated runs.}
We evaluate all baselines, declaration settings, and planning patterns with three random seeds (7, 13, and 42), reporting mean and standard deviation across seeds using the same tasks and inference configurations.

\noindent\textbf{Evaluation Metrics.}
We report both task- and process-level metrics. Task success follows each benchmark's standard protocol: task success rate (TSR) for ALFWorld and WebArena, TSR and step success rate (SSR) for Mind2Web, and patch success rate (PSR) for SWE-bench Verified. Process metrics include plan quality, plan adherence, and structural faithfulness (Section~\ref{sec:planning-metrics}). We also measure execution cost through environment interactions, LLM calls, generated tokens (thinking vs. content), and completed trajectories to control for differences in inference and interaction budget.


\noindent\textbf{Inference Settings.}
For each model, decoding parameters, thinking settings, tool-calling policies, and generation budgets are fixed across conditions for each model. Full inference and sampling configurations are provided in Appendix~\ref{app:inference-config}.


\noindent\textbf{Development and tuning protocol.}
The four planning-pattern executors, their prompts, and the declaration prompt were fixed before the final evaluation runs, and the same implementations and routing rules are applied to every task within a benchmark. The benchmark tasks were used for inference only; no model parameters were trained or fitted on them. For few-shot declaration, demonstration tasks are excluded from scoring. The null baselines and permutation tests were specified after the main results and are reported as post-hoc analyses.

\noindent\textbf{Plan verification and quality judging.} Structural fidelity is measured with a rule-based verifier that matches declared plan steps to executed actions and requires all scorable steps to be preserved in order; we validate it against two independent human annotators, with full matching and agreement results in Appendices~\ref{app:plan_structure_verifier} and \ref{app:human_validation_verifier}. 
Plan quality and adherence are scored separately with an LLM-as-judge
pipeline using a 0--3 GPA-style rubric~\citep{jia2025your}; full prompts
and judge configurations are provided in
Appendix~\ref{app:plan_quality_judge}.




\begin{table*}[t]
\centering
\scriptsize
\caption{
Plan-structure maintenance under generic Plan+ReAct, overall and by declared plan length. \emph{Overall} is the percentage of trajectories preserving the declared structure; \emph{Avg. steps} is the mean number of declared plan units. Pattern-specific executors are omitted because they enforce their execution structure by design. Values are mean percentages \(\pm\) SD across seeds 7, 13, and 42. \(^{\ddagger}\)Sparse bins should be interpreted cautiously.
}
\label{tab:structure-maintenance}
\setlength{\tabcolsep}{3.2pt}
\begin{tabular}{|l|l|cc|ccccc|}
\hline
\textbf{Benchmark} &
\textbf{Model} &
\textbf{Overall} &
\textbf{Avg. steps} &
\multicolumn{5}{c|}{\textbf{Structure maintained by declared plan length}} \\
\cmidrule(lr){5-9}
&
&
&
&
$\mathbf{\leq3}$ &
\textbf{4--5} &
\textbf{6--7} &
\textbf{8--10} &
$\mathbf{>10}$ \\
\hline

\multirow{3}{*}{ALFWorld}
& DeepSeek-V4
& $27.4{\pm}5.1$
& $6.2{\pm}0.3$
& $53.3{\pm}37.7$
& $36.7{\pm}7.0$
& $21.4{\pm}3.0$
& $0.0{\pm}0.0^{\ddagger}$
& $0.0{\pm}0.0^{\ddagger}$ \\

& Qwen3.6-35B
& $23.7{\pm}2.7$
& $6.1{\pm}0.3$
& $66.7{\pm}27.2$
& $36.5{\pm}1.3$
& $9.5{\pm}3.9$
& $1.2{\pm}1.7$
& $0.0{\pm}0.0^{\ddagger}$ \\

& Gemma-4-26B
& $21.5{\pm}1.9$
& $7.6{\pm}0.2$
& $91.7{\pm}8.3$
& $49.6{\pm}2.2$
& $4.8{\pm}4.8$
& $0.0{\pm}0.0$
& $0.0{\pm}0.0$ \\
\hline

\multirow{3}{*}{Mind2Web}
& DeepSeek-V4
& $44.6{\pm}0.9$
& $4.8{\pm}0.0$
& $73.3{\pm}0.5$
& $39.2{\pm}1.4$
& $31.2{\pm}1.7$
& $10.0{\pm}3.4$
& $5.9{\pm}3.9$ \\

& Qwen3.6-35B
& $36.1{\pm}0.9$
& $4.8{\pm}0.0$
& $62.4{\pm}1.8$
& $31.5{\pm}0.3$
& $17.9{\pm}1.1$
& $8.6{\pm}1.4$
& $4.2{\pm}3.1$ \\

& Gemma-4-26B
& $28.2{\pm}0.2$
& $5.7{\pm}0.0$
& $70.1{\pm}2.9$
& $24.8{\pm}0.8$
& $13.4{\pm}0.5$
& $12.7{\pm}1.8$
& $11.3{\pm}4.9$ \\
\hline

\multirow{2}{*}{SWE-bench}
& DeepSeek-V4
& $25.4{\pm}2.1$
& $4.6{\pm}0.1$
& $33.4{\pm}1.3$
& $25.6{\pm}1.2$
& $15.5{\pm}2.5$
& $21.2{\pm}12.1$
& $0.0{\pm}0.0^{\ddagger}$ \\

& Qwen3.6-35B
& $23.5{\pm}1.0$
& $4.8{\pm}0.1$
& $47.9{\pm}3.2$
& $22.3{\pm}0.5$
& $13.6{\pm}0.1$
& $5.9{\pm}0.3$
& $3.6{\pm}3.6$ \\

& Gemma-4-26B
& $12.1{\pm}0.9$
& $5.6{\pm}0.3$
& $33.3{\pm}1.2$
& $13.3{\pm}0.6$
& $4.1{\pm}0.2$
& $5.6{\pm}0.7$
& $0.0{\pm}0.0$ \\
\hline

\multirow{2}{*}{WebArena}
& DeepSeek-V4
& $67.9 {\pm} 4.4$
& $5.1 {\pm} 0.1$
& $84.5 {\pm} 1.2$
& $68.7 {\pm} 7.4$
& $64.1 {\pm} 1.9$
& $45.0 {\pm} 5.0$
& $24.7 {\pm} 17.0$\\

& Qwen3.6-35B
& $42.2 {\pm} 1.2$
& $6.7 {\pm} 0.2$
& $65.9 {\pm} 6.8$
& $59.1 {\pm} 3.6$
& $39.7 {\pm} 0.3$
& $18.4 {\pm} 4.1$ & $5.4 {\pm} 1.5$ \\

& Gemma-4-26B
& $61.4 {\pm} 2.5$
& $5.6 {\pm} 0.1$
& $76.1 {\pm} 1.5$
& $65.9 {\pm} 2.6$
& $53.9 {\pm} 18.5$
& $52.3 {\pm} 2.3$ & $32.5 {\pm} 9.8$ \\
\hline
\end{tabular}
\end{table*}

\section{Results}

\subsection*{\textbf{[RQ1]:} Pattern-specific execution preserves declared planning structure, while generic Plan+ReAct increasingly deviates on longer plans}
\label{sec:faithfulness-results}

\noindent\textbf{Generic Plan+ReAct does not reliably preserve the committed planning structure.}
To examine whether agents preserve their declared planning structure, we first measure structure maintenance under Plan+ReAct. Table~\ref{tab:structure-maintenance} reports overall plan-structure maintenance under Plan+ReAct and its variation with declared plan length. We make the following observations: (i) Plan+ReAct reveals a substantial Plan Declaration--Execution Gap across benchmarks. Across ALFWorld, Mind2Web, and SWE-bench, only about \(22\)--\(45\%\) of Plan+ReAct trajectories preserve the declared structure. (ii) Plan structure fidelity also decreases with plan length: on ALFWorld, maintenance falls from (36.5)--(49.6\%) for 4--5-step plans to (4.8)--(21.4\%) for 6--7 steps and approaches zero for longer plans; Mind2Web shows a similar decline, while SWE-bench shows the same overall pattern from short to medium-length plans. WebArena is a short-plan boundary case, with higher maintenance ((65.6)--(69.1\%)) and average plans of only (3.3)--(3.8) steps.

\noindent\textbf{Hierarchical/Search plans are particularly difficult for generic ReAct execution to preserve.} We next analyze structural maintenance by planning mode. The mode-level breakdown in Appendix~\ref{app:structure-bymode}, Table~\ref{tab:structure-bymode}, shows that the declaration--execution gap is most pronounced for richer planning structures. Hierarchical plans are difficult for generic Plan+ReAct to preserve, with structure maintenance typically below \(25\%\) across ALFWorld, Mind2Web, and SWE-bench. Search shows a similar pattern where enough declarations are available. Together with the plan-length results, this indicates that generic ReAct is  unreliable for multi-level or multi-candidate planning structures.

\noindent\textbf{Verifier agreement with human annotations.}
We further validate the rule-based verifier against two independent human annotators on 100 sampled Plan+ReAct ALFWorld trajectories. Verifier--human agreement (\(\kappa=0.31/0.34\)) is comparable to human--human agreement (\(\kappa=0.35\)); full validation results are reported in Appendix~\ref{app:human_validation_verifier}.

\begin{figure*}[t]
    \centering
    \includegraphics[width=0.8\textwidth]
    {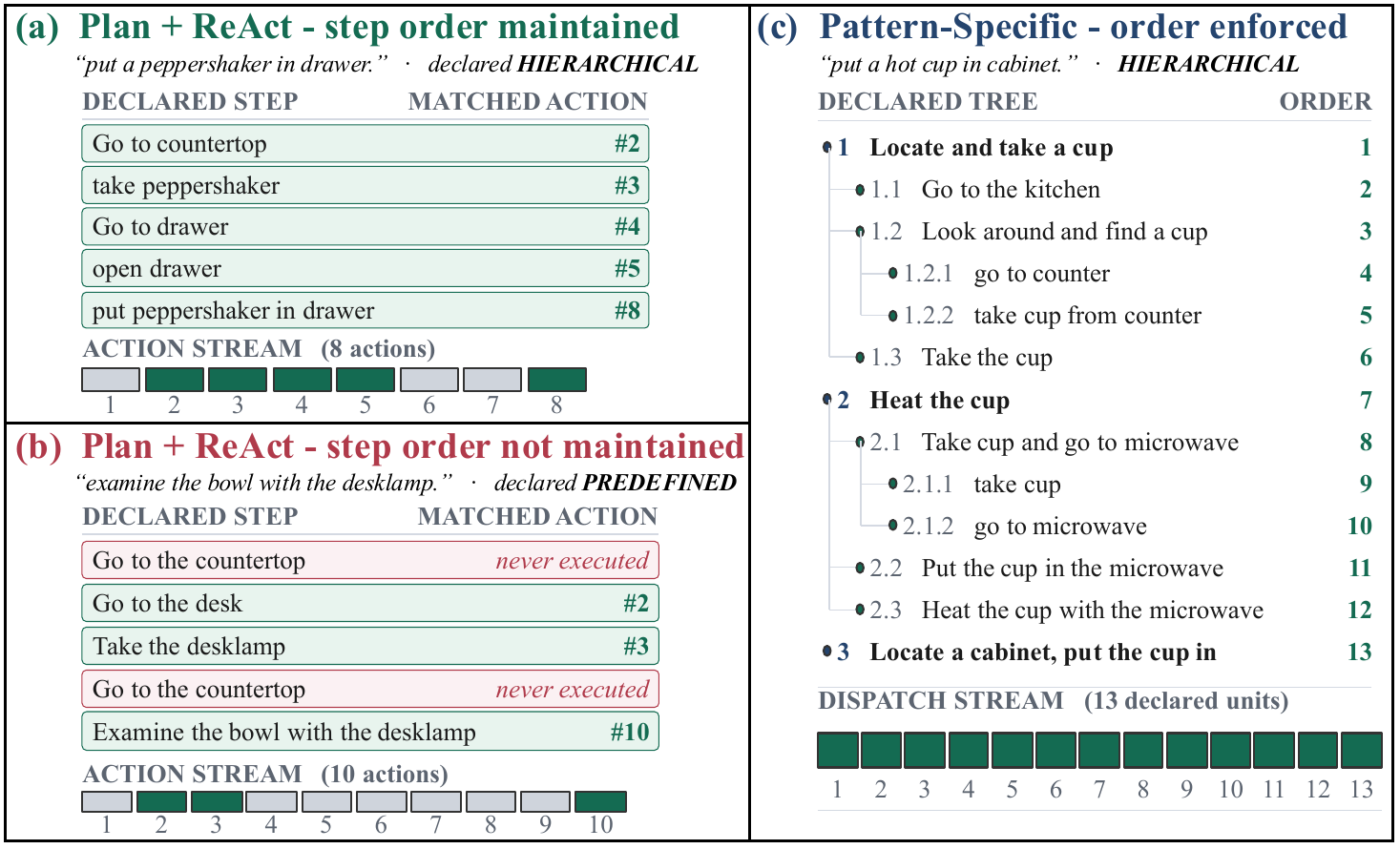}
    \caption{
    Qualitative examples of plan-structure preservation. (a) A Plan+ReAct trajectory that maintains the declared hierarchical
structure: declared units are reached in order despite intervening actions. (b) A Plan+ReAct trajectory that does not maintain the declared predefined structure, skipping two plan units before continuing with later ones. (c) A routed pattern-specific hierarchical executor, where the declared tree directly determines the dispatch sequence. Green denotes maintained/dispatched plan units and red denotes declared units that are not reached.
    }
    \label{fig:qualitative-structure}
\end{figure*}

\noindent\textbf{Qualitative trajectories illustrate the declaration--execution gap.}
Figure~\ref{fig:qualitative-structure} provides representative trajectories that complement the aggregate results. In the maintained Plan+ReAct example, all declared hierarchical units are reached in their original structural order, even though additional environment actions occur between them. In contrast, the non-maintained example skips multiple units of the declared predefined plan while continuing with later parts of the trajectory. This illustrates that a generic ReAct executor can depart from the committed plan structure while continuing to act in the environment. The pattern-specific executor shows a different behavior: the hierarchical plan is explicitly traversed through its declared tree, with all 13 units dispatched according to the prescribed structure. These examples illustrate the distinction captured quantitatively in Tables~\ref{tab:structure-maintenance} and \ref{tab:structure-bymode} (Appendix~\ref{app:structure-bymode}). This suggests that providing a plan as context does not enforce its structure, whereas a pattern-specific executor makes that structure part of the execution control flow.

\begin{table*}[t]
\centering
\scriptsize
\caption{
Plan adherence on ALFWorld under the fixed execution budget, evaluated on 134 tasks with three random seeds per model. Values are mean $\pm$ standard deviation across seeds. \emph{Adherence} is the fraction of declared executable plan units completed within the allocated budget; \emph{Full Plan} is the percentage of tasks for which all declared units were completed. Hierarchical planning is evaluated over executable leaf nodes.
}
\label{tab:alfworld-adherence}
\begin{tabular}{|l|l|c|c|c|c|}
\hline
\textbf{Pattern} & \textbf{Model} &
\textbf{Planned / Task} & \textbf{Completed / Task} &
\textbf{Adherence} & \textbf{Full Plan Completed (\%)} \\
\hline

\multirow{3}{*}{Predefined}
& DeepSeek-V4 & $5.917{\pm}0.023$ & $5.167{\pm}0.040$ & $0.871{\pm}0.009$ & $58.7{\pm}1.1$ \\

& Qwen3.6-35B & $5.867{\pm}0.132$ & $5.053{\pm}0.235$ & $0.864{\pm}0.018$ & $54.0{\pm}3.5$ \\

& Gemma-4-26B & $6.243{\pm}0.040$ & $5.797{\pm}0.121$ & $0.930{\pm}0.023$ & $72.9{\pm}7.5$ \\
\hline

\multirow{3}{*}{Sequential}
& DeepSeek-V4 & $4.070{\pm}0.056$ & $4.070{\pm}0.056$ & $1.000{\pm}0.000$ & $100.0{\pm}0.0$ \\

& Qwen3.6-35B & $4.023{\pm}0.057$ & $4.023{\pm}0.057$ & $1.000{\pm}0.000$ & $100.0{\pm}0.0$ \\

& Gemma-4-26B & $3.810{\pm}0.108$ & $3.810{\pm}0.108$ & $1.000{\pm}0.000$ & $100.0{\pm}0.0$ \\
\hline

\multirow{3}{*}{Hierarchical}
& DeepSeek-V4 & $13.063{\pm}0.193$ & $10.440{\pm}0.135$ & $0.812{\pm}0.005$ & $37.1{\pm}4.8$ \\

& Qwen3.6-35B & $10.480{\pm}0.087$ & $8.177{\pm}0.107$ & $0.793{\pm}0.006$ & $30.3{\pm}3.1$ \\

& Gemma-4-26B & $13.177{\pm}0.266$ & $11.010{\pm}0.263$ & $0.845{\pm}0.016$ & $41.3{\pm}3.0$ \\
\hline

\multirow{3}{*}{Search}
& DeepSeek-V4 & $3.337{\pm}0.055$ & $3.320{\pm}0.046$ & $0.996{\pm}0.004$ & $98.5{\pm}1.3$ \\

& Qwen3.6-35B & $3.223{\pm}0.047$ & $3.223{\pm}0.047$ & $1.000{\pm}0.000$ & $100.0{\pm}0.0$ \\

& Gemma-4-26B & $2.993{\pm}0.006$ & $2.993{\pm}0.006$ & $1.000{\pm}0.000$ & $100.0{\pm}0.0$ \\
\hline

\end{tabular}
\end{table*}

\noindent\textbf{Plan completion also depends on plan size and execution budget.}
Plan structural fidelity and plan adherence capture different properties: the former measures whether execution preserves the organization of the declared plan, whereas the latter measures how much of the plan is completed within the interaction budget. On ALFWorld (Table~\ref{tab:alfworld-adherence}), short Sequential plans are completed almost entirely, while larger Hierarchical plans contain \(10\)--\(13\) executable units and achieve lower adherence (\(\sim79\)--\(85\%\)). Predefined plans fall between these cases, while Search shows high candidate-level completion. Thus, structural fidelity and plan completion capture different properties: pattern-specific executors preserve the intended organization, but completion additionally depends on plan size and budget. Results for Mind2Web and SWE-bench are provided in Appendix~\ref{app:plan-adherence}.

\begin{table*}[t]
\centering
\scriptsize
\caption{
Pattern-ceiling analysis across benchmarks. \(S(p)\) is task success under forced execution of pattern \(p\); \(\mathrm{DSR}_{\max}\) is the per-task oracle; and \(H=\mathrm{DSR}_{\max}-\max_p S(p)\) is the headroom beyond the strongest fixed pattern. Values are mean \(\pm\) SD across seeds where available.
}
\label{tab:ceiling}
\resizebox{\textwidth}{!}{%
\begin{tabular}{|l|l|r|cccc|c|c|c|c|}
\hline
\textbf{Bench} &
\textbf{Model} &
\textbf{$N$} &
\textbf{SEQ} &
\textbf{PRED} &
\textbf{HIER} &
\textbf{SEARCH} &
\textbf{DSR$_{\max}$} &
\textbf{$H$} &
\textbf{DSR$_{\max}$ w/o Search} &
\textbf{$H$ w/o Search} \\
\hline

\multirow{3}{*}{ALFWorld}
& DeepSeek-V4
& 134 & $0.633{\pm}0.020$ & $0.556{\pm}0.008$ & $0.840{\pm}0.009$
& $\mathbf{0.918{\pm}0.026}$ & 
$0.953{\pm}0.013$ & $0.035{\pm}0.019$ & $0.898\pm0.004$ & $0.057\pm0.013$ \\

& Qwen3.6-35B
& 134
& $0.662{\pm}0.027$
& $0.550{\pm}0.023$
& $0.714{\pm}0.037$
& $\mathbf{0.796{\pm}0.004}$
& $0.925{\pm}0.016$
& $0.129{\pm}0.020$
& $0.843\pm0.030$ & $0.129\pm0.007$
\\

& Gemma-4-26B
& 134
& $0.550{\pm}0.014$
& $0.515{\pm}0.016$
& $\mathbf{0.600{\pm}0.013}$
& $0.495{\pm}0.058$
& $0.841{\pm}0.007$
& $0.241{\pm}0.013$
& $0.781\pm0.009$ & $0.182\pm0.021$
\\ \hline

\multirow{3}{*}{Mind2Web} & DeepSeek-V4 & 1341 & $0.053{\pm}0.005$
& $0.048{\pm}0.003$ & $0.041{\pm}0.001$ & $\mathbf{0.057{\pm}0.006}$ & 
$0.097{\pm}0.003$
& $0.037{\pm}0.005$ 
& $0.087{\pm}0.007$ & $0.034{\pm}0.003$
\\

& Qwen3.6-35B & 1341 & $0.032{\pm}0.001$ & $0.036{\pm}0.002$
& $0.029{\pm}0.002$ & $\mathbf{0.052{\pm}0.001}$ & 
$0.084{\pm}0.004$ & $0.031{\pm}0.004$ 
& $0.064{\pm}0.003$ & $0.028{\pm}0.002$
\\

& Gemma-4-26B & 1341 & $0.114 {\pm} 0.006$ & $\mathbf{0.147 {\pm} 0.005}$ & $0.083{\pm}0.002$
& $0.086 {\pm} 0.006$ & 
$0.282{\pm}0.002$ & $0.135{\pm}0.006$ 
& $0.253{\pm}0.002$ & $0.106{\pm}0.002$
\\ \hline
\multirow{3}{*}{SWE-bench}
& DeepSeek-V4 & 500 & $0.395{\pm}0.010$ & $0.372{\pm}0.030$ & $\mathbf{0.442{\pm}0.024}$
& $0.400{\pm}0.018$ &
$0.552{\pm}0.023$ & $0.110{\pm}0.000$
& $0.517{\pm}0.026$ & $0.075{\pm}0.002$ \\

& Qwen3.6-35B & 500 & $0.239{\pm}0.029$ & $0.303{\pm}0.007$ & $\mathbf{0.330{\pm}0.011}$
& $0.296{\pm}0.040$ &
$0.445{\pm}0.038$ & $0.115{\pm}0.027$
& $0.415{\pm}0.022$ & $0.085{\pm}0.011$ \\

& Gemma-4-26B & 500 & $0.170{\pm}0.019$ & $0.247{\pm}0.012$ & $\mathbf{0.290{\pm}0.015}$
& $0.211 {\pm}0.024$& $0.386{\pm}0.014$
 & $0.096{\pm} 0.021$
& $0.362{\pm}0.017$ & $0.072{\pm} 0.023$

\\ \hline
\multirow{3}{*}{WebArena}
& DeepSeek-V4 &204 
& $0.389 {\pm} 0.026$ & $0.372 {\pm} 0.031$ & $0.537 {\pm} 0.014$ & $\mathbf{0.580 {\pm} 0.006}$&
$0.736 {\pm} 0.020$ & $0.156{\pm}0.014$ & $0.676{\pm}0.011$ & $0.139{\pm}0.003$\\

& Qwen3.6-35B & 204 & $0.315 {\pm} 0.009$ & $0.364 {\pm} 0.009$ & $0.395 {\pm} 0.099$ &$\mathbf{0.577 {\pm} 0.026}$ 
&  $0.693 {\pm} 0.017$ & $0.119{\pm}0.011$ & $0.557{\pm}0.034$ & $0.162{\pm}0.065$ 
\\ 

& Gemma-4-26B & 204 & $0.364 {\pm} 0.019 $ & $0.290 {\pm} 0.005$&$\mathbf{0.378 {\pm} 0.011}$  &$0.306 {\pm} 0.033$ &$0.578 {\pm} 0.011$
& $0.200 {\pm} 0.022$ &$0.514 {\pm} 0.014$  & $0.136 {\pm} 0.003$ 
\\ 

\hline
\end{tabular}}
\end{table*}


\subsection*{\textbf{[RQ2]:} Planning-pattern effectiveness varies across environments and models.}

We next compute the forced-pattern ceiling described in Section~\ref{sec:ceiling}. Table~\ref{tab:ceiling} yields two main observations. First, the strongest planning pattern varies across benchmark--model pairs: Search is strongest for DeepSeek-V4 and Qwen3.6-35B on ALFWorld, Mind2Web, and WebArena, whereas Hierarchical is strongest on SWE-bench and for Gemma-4-26B on ALFWorld, and Predefined is strongest for Gemma-4-26B on Mind2Web. Thus, planning-pattern effectiveness depends on both the environment and backbone model. Second, the forced-pattern oracle exceeds the strongest fixed pattern across all benchmark--model pairs, indicating additional headroom when multiple planning modes are available. However, because the oracle also provides multiple independent execution attempts, it should be interpreted as an empirical ceiling rather than direct evidence of task-specific complementarity. This motivates the retry-matched control in Appendix~\ref{app:ceiling-analysis}; after three retries of the strongest fixed pattern, the residual oracle gap is only (-0.025) to (+0.027) across ALFWorld, Mind2Web, and SWE-bench (Table~\ref{tab:retry-null}).



\noindent\textbf{The oracle gap is not driven solely by Search.}
Because Search planning generates and executes multiple candidate plans before selecting among them with a rubric-based judge, it receives a larger multi-rollout execution budget than the other planning patterns. We therefore recompute the oracle after excluding Search (see Table~\ref{tab:ceiling}) and still observe a gap between the strongest fixed pattern and the forced-pattern oracle across benchmarks. On ALFWorld, for example, the no-Search oracle reaches (0.898), (0.843), and (0.781), while three retries of Hierarchical reach (0.963), (0.896), and (0.813), respectively, showing that much of the remaining gap can be explained by repeated execution rather than per-task pattern complementarity.

\begin{table}[t]
\centering
\scriptsize
\caption{
Plan quality (PQ) and task success on ALFWorld. Plans are evaluated by a Gemma-4-26B judge. Forced-pattern results report the mean $\pm$ half-range across seeds ($n=134$ tasks per seed). The final column reports the task-level Pearson correlation between plan quality and task success separately for seeds 13 and 7. $^\dagger$Plan+ReAct is reported for one seed. Bold indicates the highest PQ or TSR for each model.
}
\label{tab:plan-quality}
\begin{tabular}{|l|l|c|c|c|}
\hline
\textbf{Mode} & \textbf{Model} & \textbf{PQ} &
\textbf{TSR} & \textbf{PQ--success $r$} \\
\hline

\multirow{2}{*}{Predefined}
& DeepSeek-V4
& $2.75{\pm}0.01$
& $0.55{\pm}0.00$
& $-0.031,\,-0.005$ \\

& Qwen3.6-35B
& $2.44{\pm}0.05$
& $0.56{\pm}0.03$
& $+0.045,\,+0.095$ \\
\hline

\multirow{2}{*}{Sequential}
& DeepSeek-V4
& $\mathbf{2.84{\pm}0.03}$
& $0.64{\pm}0.02$
& $-0.025,\,-0.061$ \\

& Qwen3.6-35B
& $2.66{\pm}0.03$
& $0.68{\pm}0.01$
& $+0.090,\,-0.040$ \\
\hline

\multirow{2}{*}{Hierarchical}
& DeepSeek-V4
& $2.72{\pm}0.02$
& $0.85{\pm}0.00$
& $-0.060,\,-0.035$ \\

& Qwen3.6-35B
& $\mathbf{2.82{\pm}0.00}$
& $0.74{\pm}0.03$
& $+0.140,\,-0.031$ \\
\hline

\multirow{2}{*}{Search}
& DeepSeek-V4
& $2.53{\pm}0.03$
& $\mathbf{0.91{\pm}0.03}$
& $+0.026,\,+0.146$ \\

& Qwen3.6-35B
& $1.94{\pm}0.17$
& $\mathbf{0.79{\pm}0.00}$
& $+0.158,\,+0.181$ \\
\hline

\multirow{2}{*}{Plan+ReAct}
& DeepSeek-V4
& $\mathbf{2.84 \pm 0.02}$
& $0.48 \pm 0.01$
& $-0.017,\,-0.029$ \\

& Qwen3.6-35B
& $2.76 \pm 0.03$
& $0.54 \pm 0.04$
& $+0.104,\,-0.045$ \\
\hline



\end{tabular}
\end{table}

\begin{table*}[t]
\centering
\scriptsize
\setlength{\tabcolsep}{3.2pt}
\renewcommand{\arraystretch}{1.12}
\caption{
End-to-end comparison of planning and execution strategies. \emph{Best Fixed} (\(S^\star\)) is the strongest single planning pattern applied to all tasks; \emph{Routing@1} executes the top-1 declared pattern with its corresponding executor; and \emph{Oracle} is the per-task forced-pattern ceiling. Bold denotes the strongest non-oracle method. Routing@1 uses thinking disabled.
}
\label{tab:end-to-end-routing}
\resizebox{\textwidth}{!}{%
\begin{tabular}{|l|l|c|c|c|c|c|c|}
\hline
\textbf{Benchmark} & \textbf{Model} & \textbf{Metric} &
\textbf{Flat ReAct} & \textbf{Plan+ReAct} & \textbf{Best Fixed $S^\star$} & \textbf{Routing@1} & \textbf{Oracle} \\
\hline

\multirow{3}{*}{ALFWorld}

& DeepSeek-V4
& TSR & $0.440{\pm}0.012$ & $0.480{\pm}0.009$ & $\mathbf{0.918{\pm}0.026}$ (SEARCH) & $0.721{\pm}0.029$ & $0.953{\pm}0.013$ \\

& Qwen3.6-35B
& TSR & $0.535{\pm}0.007$ & $0.540{\pm}0.037$ & $\mathbf{0.796{\pm}0.004}$ (SEARCH) & $0.667{\pm}0.049$ & $0.925{\pm}0.016$ \\

& Gemma-4-26B & TSR & $0.112 {\pm} 0.040$ & $0.153{\pm}0.026$ & $\mathbf{0.600{\pm}0.013}$ (HIER) & $0.560{\pm}0.016$ & $0.841{\pm}0.007$ \\

\hline

\multirow{6}{*}{Mind2Web}

& \multirow{2}{*}{DeepSeek-V4}
& TSR & $\mathbf{0.058{\pm}0.003}$ & $0.038{\pm}0.005$
& $0.057{\pm}0.006$ (SEARCH) & $0.051{\pm}0.004$ & $0.097{\pm}0.003$ \\

&
& SSR & $\mathbf{0.434{\pm}0.001}$ & $0.407{\pm}0.002$
& $\mathbf{0.449{\pm}0.000}$ (SEARCH) & $0.425{\pm}0.007$ & $0.550{\pm}0.001$ \\

\cline{2-8}

& \multirow{2}{*}{Qwen3.6-35B}
& TSR & $0.042{\pm}0.002$ & $0.036{\pm}0.004$ & $\mathbf{0.052{\pm}0.001}$ (SEARCH) & $0.034{\pm}0.001$ & $0.084{\pm}0.004$ \\

&
& SSR & $0.366{\pm}0.003$ & $0.338{\pm}0.001$ & $\mathbf{0.411{\pm}0.003}$ (SEARCH) & $0.339{\pm}0.002$ & $0.505{\pm}0.001$ \\

\cline{2-8}

& \multirow{2}{*}{Gemma-4-26B}
& TSR & $0.051{\pm}0.002$ & $0.043{\pm}0.004$ & $\mathbf{0.147{\pm}0.005}$ (PRED) & $0.113{\pm}0.006$  & $0.282{\pm}0.002$ \\

&
& SSR & $0.395{\pm}0.001$ & $0.365{\pm}0.002$ & $\mathbf{0.388 {\pm} 0.006}$ (SEARCH) & $0.320{\pm}0.002$ & $0.585{\pm}0.002$ \\
\hline

\multirow{3}{*}{SWE-bench}

& DeepSeek-V4
& PSR & $0.384{\pm}0.010$ & $0.360{\pm}0.005$ & $\mathbf{0.442{\pm}0.024}$ (HIER) & $0.415{\pm}0.017$ & $0.552{\pm}0.023$ \\

& Qwen3.6-35B
& PSR & $0.179{\pm}0.034$ & $0.223{\pm}0.005$ & $\mathbf{0.330{\pm}0.011}$ (HIER) & $0.313{\pm}0.006$ & $0.445{\pm}0.038$ \\

& Gemma-4-26B
& PSR & $0.127{\pm}0.030$ & $0.126{\pm}0.004$ & $\mathbf{0.290{\pm}0.015}$ (HIER) & $0.257{\pm}0.004$ & $0.386{\pm}0.014$ \\
\hline

\multirow{3}{*}{WebArena}

& DeepSeek-V4
& TSR & $0.455 {\pm} 0.000$ & $0.472 {\pm} 0.006$ & $\mathbf{0.580{\pm}0.006}$ (SEARCH) & $0.460{\pm}0.011$ & $0.736{\pm}0.020$ \\

& Qwen3.6-35B
& TSR & $0.332 {\pm} 0.009$ & $0.338 {\pm} 0.037$ & $\mathbf{0.577{\pm}0.026}$ (SEARCH) & $0.404{\pm} 0.019$ & $0.693 {\pm} 0.017$ \\

& Gemma-4-26B & TSR & $0.318 {\pm} 0.023$ & $0.304 {\pm} 0.020$ & $\mathbf{0.378 {\pm} 0.011}$ (HIER) & $0.432 ± 0.014$ & $0.578{\pm}0.011$ \\

\hline

\end{tabular}}
\end{table*}

\noindent\textbf{Judged plan quality is only weakly related to task success.}
We now test whether higher judged plan quality is associated with higher task success. As shown in Table~\ref{tab:plan-quality}, the highest-quality plan on ALFWorld is not consistently associated with the highest-performing planning pattern. 
For DeepSeek-V4, Sequential receives the highest judged plan quality ($2.84{\pm}0.03$), yet Search achieves substantially higher task success ($0.91{\pm}0.03$ versus $0.64{\pm}0.02$) despite a lower quality score ($2.53{\pm}0.03$). Similarly, for Qwen3.6-35B, Hierarchical receives the highest plan-quality score ($2.82{\pm}0.00$), whereas Search achieves the
highest task success ($0.79{\pm}0.00$) despite the lowest score
($1.94{\pm}0.17$). At the task level, plan-quality and success are only weakly correlated
($|r|\leq0.181$). Thus, the advantage of a planning pattern cannot be explained simply by how good its plan appears in isolation; performance also depends on how well the planning structure fits the task and its execution environment.

Together, these results show that planning effectiveness varies across
environments and models, but that the raw per-task oracle overstates the
value of task-specific selection: once repeated execution is controlled,
little additional advantage remains over the strongest fixed pattern. Moreover, pattern-specific success cannot be explained by generic judgments of plan quality, which remain only weakly associated with task success. The remaining question is whether realizing these gains requires executing the selected planning structure through its corresponding executor, which we examine next.

\begin{figure*}[t]
    \centering
    \includegraphics[width=0.9\textwidth]
    {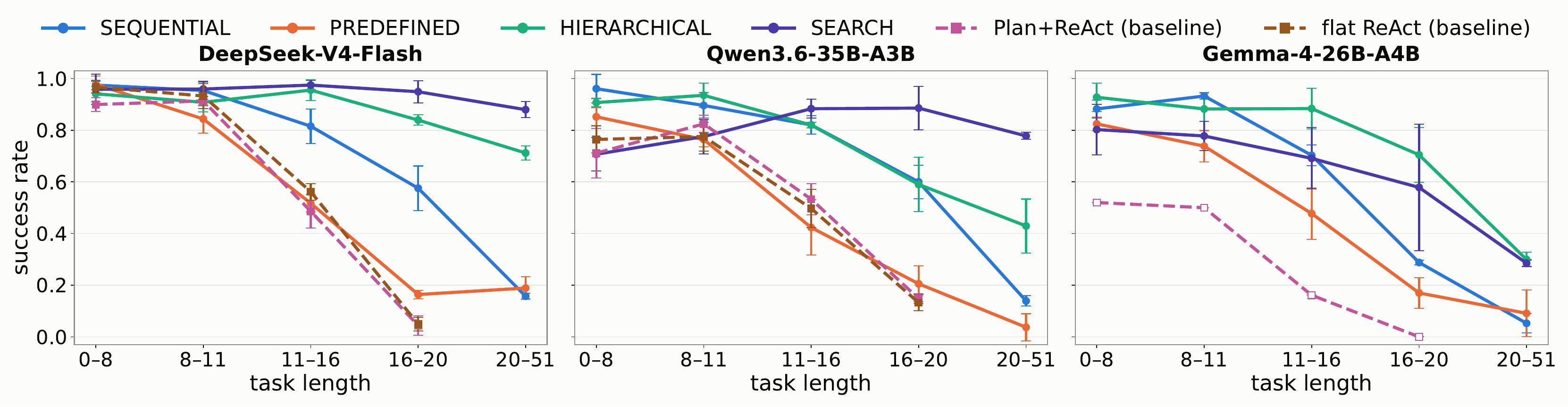}
    \hfill
    \includegraphics[width=0.9\textwidth]
    {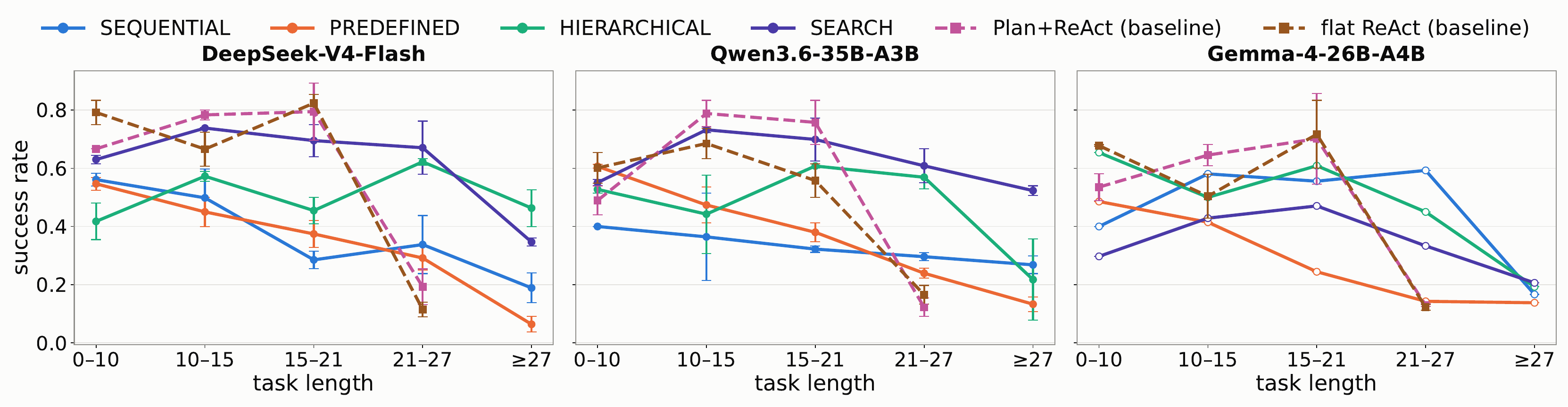}
    \hfill
    \includegraphics[width=0.9\textwidth]
    {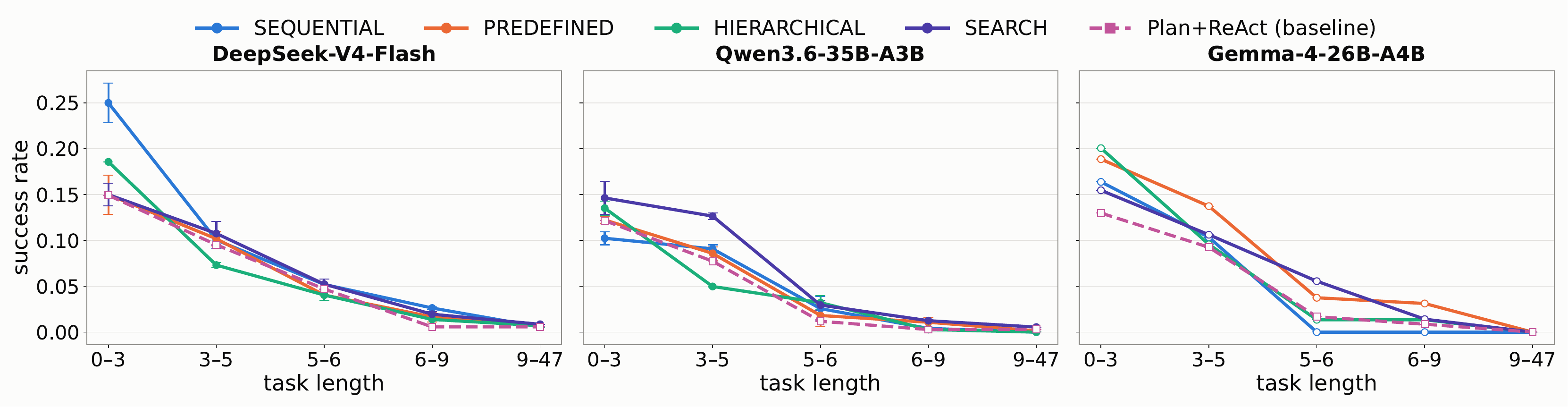}
    \hfill
    \includegraphics[width=0.9\textwidth]
    {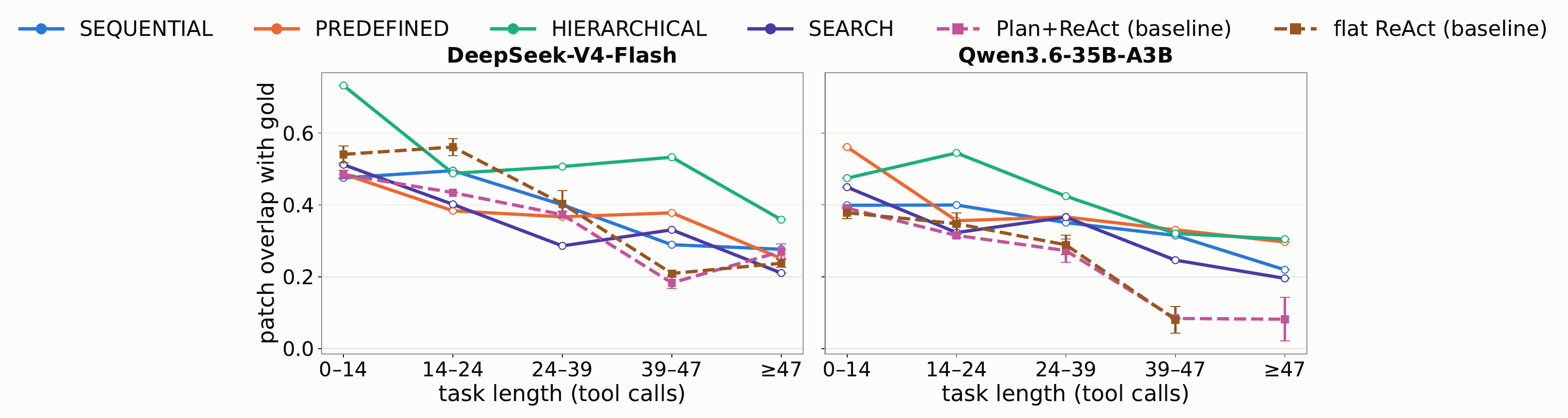}
    \caption{
    Task success as a function of task length under generic and
pattern-specific execution. Pattern-specific execution provides the largest advantage on longer tasks, where Plan+ReAct and Flat ReAct degrade more rapidly. Error bars denote variation across seeds.
    }
    \label{fig:success-by-length}
\end{figure*}

\subsection*{\textbf{[RQ3]:} Pattern-specific execution substantially improves over plan-as-context}
\label{sec:rq3}

\noindent\textbf{Pattern-specific execution provides substantially stronger task-solving capability than generic ReAct execution.}
We next test whether executing the selected planning mode with its corresponding pattern-specific executor improves end-to-end performance. Table~\ref{tab:end-to-end-routing} compares Flat ReAct, generic Plan+ReAct, the strongest fixed planning pattern, and Planning-as-Routing. Across benchmarks, generic Plan+ReAct provides limited gains over Flat ReAct, whereas pattern-specific execution yields substantially higher success, showing that selecting a plan alone is insufficient when the executor does not preserve its structure.

On ALFWorld, DeepSeek-V4 improves from \(0.480\) with Plan+ReAct to \(0.918\) with the strongest fixed pattern, while Qwen3.6-35B improves from \(0.540\) to \(0.796\). On SWE-bench Verified, DeepSeek-V4 increases from \(0.360\) to \(0.442\). Similar gains appear across the remaining benchmarks, indicating that providing a plan as context to a generic ReAct
executor does not capture the gains obtained when the planning structure is implemented directly in the execution mechanism.


\noindent\textbf{Current declarations do not consistently select the strongest planning pattern.}
Table~\ref{tab:end-to-end-routing} shows that Routing@1 often remains below the strongest fixed pattern. For example, on WebArena, Search reaches \(0.580\) and \(0.577\) for DeepSeek-V4 and Qwen3.6-35B, whereas Routing@1 reaches \(0.460\) and \(0.404\). Thus, strong pattern-specific executors are available, but current declarations do not consistently select the pattern that realizes their full performance; we examine this selection problem directly in \emph{Sec. [RQ4]}.

\noindent\textbf{The advantage of pattern-specific execution becomes more pronounced on longer tasks.}
Figure~\ref{fig:success-by-length} shows that the benefit of matching the executor to the declared planning pattern is largest as task length increases. On ALFWorld, Flat ReAct and Plan+ReAct degrade sharply on longer tasks, whereas Hierarchical and Search execution remain substantially more robust, especially for DeepSeek-V4 and Qwen3.6. This separation is modest on short tasks, where several execution strategies perform similarly, but widens as more interaction steps are required. The pattern is similar on WebArena benchmark across 3 LLMs. Mind2Web shows the same general difficulty with increasing task length: success decreases for all methods, but pattern-specific executors remain competitive with or above Plan+ReAct across most length bins. These results suggest that simply providing a plan to a generic ReAct loop is often sufficient for short tasks, but preserving the corresponding planning pattern becomes increasingly important as execution horizons grow.


\noindent\textbf{Rollout-matched control for Search.}
Because Search executes multiple candidate plans, we compare it against \(3\times\) independent Flat ReAct runs followed by the same trajectory judge. Repeated ReAct improves performance but does not fully close the Search advantage: Search remains higher by \(0.117\)--\(0.381\) on ALFWorld and \(0.153\)--\(0.233\) on WebArena, while differences are smaller and model-dependent on Mind2Web. Full multi-rollout and pass@3 controls are provided in Appendix~\ref{app:repeated-rollout-react}, Table~\ref{tab:multirollout-react}.

\begin{table*}[t]
\centering
\scriptsize
\setlength{\tabcolsep}{3.2pt}
\renewcommand{\arraystretch}{1.10}
\caption{
Planning-pattern selection and improvement. @1--@3 report ranked fallback success when declared modes are attempted in order up to rank \(k\). \emph{Few-shot@1} uses task--pattern demonstrations, and \(\Delta\) is the absolute gain over the original top-1 declaration. Oracle is the per-task forced-pattern ceiling. Mind2Web reports both TSR and SSR; results are averaged across seeds.
}
\label{tab:selection-improvement}
\resizebox{\textwidth}{!}{%
\begin{tabular}{|l|l|c|c|c|c|c|c|c|c|}
\hline
\textbf{Benchmark} &
\textbf{Model} &
\textbf{Metric} &
\textbf{Think} &
\textbf{@1} &
\textbf{@2} &
\textbf{@3} &
\textbf{Few-shot@1} &
\textbf{$\Delta_{\mathrm{FS}}$} &
\textbf{Oracle} \\
\hline

\multirow{6}{*}{ALFWorld}
& \multirow{2}{*}{DeepSeek-V4}
& TSR & Off
& $0.721{\pm}0.028$
& $0.893{\pm}0.001$
& $0.948{\pm}0.011$
& $\mathbf{0.793{\pm}0.083}$
& $\mathbf{+0.073}$
& \multirow{2}{*}{$0.953{\pm}0.013$} \\

&
& TSR & On
& $0.656{\pm}0.019$
& $0.863{\pm}0.007$
& $0.908{\pm}0.012$
& $\mathbf{0.812{\pm}0.047}$
& $\mathbf{+0.156}$
& \\

\cline{2-10}

& \multirow{2}{*}{Qwen3.6}
& TSR & Off
& $0.667{\pm}0.049$
& $0.804{\pm}0.028$
& $0.873{\pm}0.016$
& $\mathbf{0.706{\pm}0.034}$
& $\mathbf{+0.040}$
& \multirow{2}{*}{$0.925{\pm}0.016$} \\

&
& TSR & On
& $0.674{\pm}0.021$
& $0.769{\pm}0.010$
& $0.853{\pm}0.031$
& $\mathbf{0.704{\pm}0.069}$
& $\mathbf{+0.030}$
& \\

\cline{2-10}

& \multirow{2}{*}{Gemma-4-26B}
& TSR & Off
& $0.560{\pm}0.016$
& $0.744{\pm}0.015$
& $0.873{\pm}0.016$
& $\mathbf{0.635{\pm}0.015}$
& $\mathbf{+0.075}$
& \multirow{2}{*}{$0.841{\pm}0.007$} \\

&
& TSR & On
& $0.597{\pm}0.012$
& $0.744{\pm}0.015$
& $0.813{\pm}0.012$
& $\mathbf{0.647{\pm}0.060}$
& $\mathbf{+0.050}$
& \\
\hline

\multirow{12}{*}{Mind2Web}

& \multirow{4}{*}{DeepSeek-V4}
& TSR & Off
& $0.053{\pm}0.003$
& $0.075{\pm}0.004$
& $0.092{\pm}0.004$
& $0.050{\pm}0.005$ & $-0.003$
& \multirow{2}{*}{$0.097{\pm}0.003$} \\

&
& TSR & On
& $0.051{\pm}0.004$
& $0.072{\pm}0.006$
& $0.089{\pm}0.007$
& $0.049{\pm}0.003$ & $-0.002$
& \\

&
& SSR & Off
& $0.434{\pm}0.003$
& $0.495{\pm}0.004$ & $0.532{\pm}0.002$
& $\mathbf{0.438{\pm}0.007}$
& $\mathbf{+0.004}$
& \multirow{2}{*}{$0.550{\pm}0.001$} \\

&
& SSR & On
& $0.427{\pm}0.002$
& $0.492{\pm}0.006$ & $0.529{\pm}0.005$
& $\mathbf{0.432{\pm}0.004}$
& $\mathbf{+0.005}$
& \\

\cline{2-10}

& \multirow{4}{*}{Qwen3.6}
& TSR & Off
& $0.036{\pm}0.001$
& $0.053{\pm}0.002$
& $0.072{\pm}0.002$
& $0.036{\pm}0.001$ & $+0.000$
& \multirow{2}{*}{$0.084{\pm}0.004$} \\

&
& TSR & On
& $0.034{\pm}0.004$
& $0.051{\pm}0.003$
& $0.068{\pm}0.002$
& $0.036 {\pm} 0.004$ & $+0.001$
& \\

&
& SSR & Off
& $0.339{\pm}0.002$
& $0.426{\pm}0.001$ & $0.479{\pm}0.002$
& $\mathbf{0.357{\pm}0.002}$
& $\mathbf{+0.018}$
& \multirow{2}{*}{$0.505{\pm}0.001$} \\

&
& SSR & On
& $0.341{\pm}0.006$
& $0.417{\pm}0.002$ & $0.467{\pm}0.003$
& $\mathbf{0.352{\pm}0.002}$
& $\mathbf{+0.011}$
&  \\

\cline{2-10}

& \multirow{4}{*}{Gemma-4-26B}
& TSR & Off
& $0.111{\pm}0.006$
& $0.162{\pm}0.012$
& $0.190{\pm}0.026$
& $0.111 {\pm} 0.006$ & $+0.000$
& \multirow{2}{*}{$0.282{\pm}0.002$} \\

&
& TSR & On
& $0.114{\pm}0.012$
& $0.170{\pm}0.006$
& $0.204{\pm}0.025$
& $0.115 {\pm} 0.011$ & $+0.001$
& \\

&
& SSR & Off
& $0.320{\pm}0.002$
& $0.433{\pm}0.032$ & $0.514{\pm}0.029$
& $\mathbf{0.330 {\pm} 0.006}$ & $\mathbf{+0.011}$
& \multirow{2}{*}{$0.585{\pm}0.002$} \\

&
& SSR & On
& $0.321{\pm}0.002$
& $0.441{\pm}0.024$ & $0.513{\pm}0.024$
& $\mathbf{0.328 {\pm} 0.012}$ & $+0.008$
&  \\
\hline

\multirow{4}{*}{SWE-bench}

& \multirow{2}{*}{DeepSeek-V4}
& PSR & Off
& $0.415{\pm}0.024$
& $0.498{\pm}0.021$
& $0.541{\pm}0.037$
& $0.405{\pm}0.011$ & $-0.011$
& \multirow{2}{*}{$0.552{\pm}0.023$} \\

&
& PSR & On
& $0.411{\pm}0.036$
& $0.488{\pm}0.046$
& $0.534{\pm}0.043$
& $\mathbf{0.415{\pm}0.007}$ & $\mathbf{+0.004}$
& \\

\cline{2-10}

& \multirow{2}{*}{Qwen3.6-35B}
& PSR & Off
& $0.313{\pm}0.009$
& $0.400{\pm}0.013$
& $0.438{\pm}0.025$
& $\mathbf{0.334 {\pm} 0.003}$ & $\mathbf{+0.021}$
& \multirow{2}{*}{$0.445{\pm}0.038$} \\

&
& PSR & On
& $0.310{\pm}0.004$
& $0.390{\pm}0.016$
& $0.438{\pm}0.014$
& $\mathbf{0.337{\pm}0.002}$ & $\mathbf{+0.027}$
& \\
\hline

\multirow{4}{*}{WebArena}

& \multirow{2}{*}{DeepSeek-V4}
& TSR & Off
& $0.460{\pm}0.011$
& $0.639{\pm}0.009$
& $0.696{\pm}0.003$
& $\mathbf{0.480 {\pm} 0.037}$
& $\mathbf{+0.020}$
& \multirow{2}{*}{$0.736{\pm}0.020$} \\

&
& TSR & On
& $0.384{\pm}0.009$
& $0.554{\pm}0.014$
& $0.671{\pm}0.006$
& $\mathbf{0.438 {\pm} 0.057}$
& $\mathbf{+0.054}$
& \\

\cline{2-10}

& \multirow{2}{*}{Qwen3.6}
& TSR & Off
& $0.404{\pm}0.019$
& $0.539{\pm}0.000$
& $0.639{\pm}0.005$
& $\mathbf{0.410{\pm}0.009}$
& $\mathbf{+0.006}$
& \multirow{2}{*}{$0.693{\pm}0.017$} \\

&
& TSR & On
& $0.394{\pm}0.010$
& $0.505{\pm}0.005$
& $0.591{\pm}0.034$
& $\mathbf{0.419{\pm}0.023}$
& $+0.025$
& \\


\hline
\end{tabular}}
\end{table*}

\noindent\textbf{Comparison with prior work.}
Our forced-pattern and routing results fall within the performance range of representative prior agent systems across four benchmarks. Because these systems use different models, demonstrations, training, and inference procedures, we treat them only as benchmark-level context; detailed comparisons are provided in Appendix~\ref{app:comparison_with_piror_work}.


\subsection*{\textbf{[RQ4]:} Planning-mode declarations adapt only weakly to individual tasks}
\label{sec:rq4}

We finally ask whether current LLMs select effective planning modes for individual tasks. Because a model may simply favor particular modes overall, we first test whether its declarations contain task-specific information beyond these global preferences.

\noindent\textbf{Declarations provide no reliable task-specific advantage over a task-blind policy.}
We compare top-1 declaration success with a task-blind null  preserves each model's overall planning-mode frequencies while removing the task--mode association. Across all benchmark--model pairs, \(\Delta(\pi)\) lies within the 95\% interval of a 5{,}000-permutation null (\(-0.025\) to \(+0.009\); minimum one-sided \(p=0.094\)), providing no reliable evidence that current declarations match planning modes to individual tasks better than expected from their overall declaration preferences (Full results are in Appendix~\ref{app:additiona_analysis}).

\noindent\textbf{Few-shot task--pattern demonstrations.}
Few-shot demonstrations improve top-1 declaration performance in most
evaluated configurations (Table~\ref{tab:selection-improvement}), with the
largest gain on ALFWorld for DeepSeek-V4 (\(0.656 \rightarrow 0.812\),
\(+0.156\)). Gains are smaller on Mind2Web (up to \(+0.018\)),
SWE-bench (up to \(+0.027\)), and WebArena (up to \(+0.054\)),
indicating that demonstrations improve planning-mode selection unevenly
across benchmarks. Full few-shot construction details and additional
declaration analyses are provided in
Appendix~\ref{app:declaration-prompts}.

\section{Discussion and Conclusion}

In this work, we study whether LLM agents preserve the planning structure they declare, whether planning-pattern effectiveness varies across environments and models, and whether matching a planning pattern to its corresponding executor improves task success. Our experiments reveal four main findings.
First, generic Plan+ReAct does not reliably preserve the declared planning structure: structural fidelity decreases as plans become longer. Second, planning-pattern effectiveness varies across environments and models, while retry-matched controls show that the raw forced-pattern oracle largely reflects additional execution attempts rather than task-specific pattern complementarity. Third, matching planning patterns to corresponding executors substantially improves task success over generic Plan+ReAct, particularly on longer tasks. Finally, planning-mode selection remains challenging: current declarations provide little reliable task-specific advantage over task-blind preferences, while few-shot demonstrations improve selection more consistently than enabling thinking.

Together, these findings suggest that reliable agent planning requires both selecting an appropriate planning mode and executing it through a compatible control structure. \textsc{Planning-as-Routing} separates these capabilities, helping distinguish failures of mode selection from failures of execution.

\paragraph{Limitations and future directions.}
Our study considers four planning patterns and task-level routing under fixed execution budgets. Future work could extend this framework to additional or dynamically composed planning strategies, allow agents to switch modes during execution, and evaluate structural fidelity jointly with action correctness, execution cost, and recovery behavior. Detailed limitations discussed in Appendix~\ref{app:extended_discussion}.

\subsection*{AI use statement}

In this work, generative AI tools were used under human guidance to assist with prompt drafting, grammar correction, language polishing of the manuscript. All prompts were reviewed and revised by the authors before use.

\subsection*{Ethics statement}

Our work evaluates LLM agents using publicly available, established benchmarks for embodied interaction, web navigation, and software engineering. We do not conduct new experiments involving human participants, collect new personal data, or deploy agents to interact with real users. Experiments are performed within the interfaces and environments provided by the corresponding benchmarks, including
ALFWorld, Mind2Web, WebArena, and SWE-bench Verified.

Our work studies how LLM agents select and execute planning strategies. Although improved planning and routing mechanisms can make autonomous agents more capable, the same techniques could potentially be applied to higher-risk forms of automated web interaction or software modification. Our experiments are restricted to benchmark tasks and controlled tool interfaces and are not intended to support unauthorized interaction with
external systems. We report process-level metrics in addition to task success to make agent behavior and planning failures more transparent. We, the authors, are responsible for ensuring that the benchmark data, models, and software used in this work are handled in accordance with their applicable licenses and terms of use.

\subsection*{Reproducibility statement}

We provide the planning-pattern definitions and execution architectures in Appendix~\ref{app:planning_patterns}, the complete planning-mode prompts in Appendix~\ref{app:planning_prompts}, benchmark and model details in Appendix~\ref{app:dataset-details}, and inference and sampling configurations in Appendix~\ref{app:inference-config}. We report results across repeated random seeds where applicable and provide additional analyses of plan adherence (Appendix~\ref{app:plan-adherence}), repeated fixed-pattern execution (Appendix~\ref{app:ceiling-analysis}), the task-blind declaration null (Appendix~\ref{app:ceiling-analysis}), and inference cost (Appendix~\ref{app:inference-config}). 

We plan to release the implementation of the planning-pattern executors, declaration and execution prompts, trajectory verifier, evaluation and analysis scripts, and experiment configuration files used upon publication of this work. We also plan to release the derived planning declarations, execution traces, pattern-level outcomes, and analysis metadata needed to reproduce the reported results.



\bibliography{iclr2027}
\bibliographystyle{iclr2027_conference}

\appendix
\noindent{\Large\textbf{Overview of Appendix Sections}}
\begin{itemize}
    \item Appendix~\ref{app:additional-related-work}: Additional Related Work
    \item Appendix~\ref{app:planning_patterns}: Planning Patterns
    \item Appendix~\ref{app:dataset-details}: Dataset Details
    \item Appendix~\ref{app:inference-config}: Inference and Sampling Configurations
    \item Appendix~\ref{app:plan_structure_verifier} Plan-Structure Verifier and Human Validation
    \begin{itemize}
        \item Appendix~\ref{app:verifier}: Plan-Structure Verifier
        \item Appendix~\ref{app:planning-examples}: Representative Plan+ReAct Traces
        \item Appendix~\ref{app:human_validation_verifier}: Human Validation of the Structure Verifier
    \end{itemize}
    \item Appendix~\ref{app:plan_quality_judge} Plan Quality Judge
    \item Appendix~\ref{app:structure-bymode}: Structure Preservation by Declared Planning Mode

    \item Appendix~\ref{app:plan-adherence}: Additional Plan Adherence Results
    \item Appendix~\ref{app:ceiling-analysis}: Pattern-Ceiling Analysis
    \begin{itemize}
    \item Appendix~\ref{app:repeated-rollout-react}: Repeated-Rollout ReAct Control for Search
        \item Appendix~\ref{app:repeated_fix_pattern}: Repeated-Fixed-Pattern Control
        \item Appendix~\ref{app:declaration-null}: Task-Blind Declaration Null
    \end{itemize}
    \item Appendix~\ref{app:inference-cost}: Inference Cost Analysis
    \item Appendix~\ref{app:comparison_with_piror_work}: Comparison with prior work
    \item Appendix~\ref{app:extended_discussion}: Extended Discussion
    \item Appendix~\ref{app:planning_prompts}: Zero-shot and Few-shot Prompts
    \begin{itemize}
        \item Appendix~\ref{app:declaration-prompts}: Planning-Mode Declaration Prompts
    \end{itemize}   
\end{itemize}

\section{Additional Related Work}
\label{app:additional-related-work}

\noindent\textbf{LLMs as Planning Agents.}
Prior work has proposed many planning mechanisms for LLM agents, but these methods often instantiate planning behavior through a specific agent architecture. Earlier approaches used ReAct-style agents interleave reasoning and acting in a flat step-wise loop~\cite{yao2023react}; Tree-of-Thought-style methods search over multiple reasoning paths~\cite{yao2023tree}; symbolic-planning approaches integrate LLMs with external planners~\cite{liu2023llm+}; planner--executor methods, such as Plan-and-Act, first generate a high-level plan and then execute the resulting steps through an executor~\cite{erdoganplan}; and workflow-based multi-agent systems decompose tasks through predefined roles or procedures~\cite{shen2023hugginggpt,hong2024metagpt}. More recent work introduces adaptive replanning, planner--executor collaboration, trajectory refinement, and plan-compliance analysis~\cite{liu2026adaplanbench,dong2026pear,liu2026plan,otani2026agents,wu2026planner}. While these approaches can generate task-specific plans and, in some cases, adapt or revise them during execution, the underlying planning mechanism is generally fixed by the agent framework rather than chosen per task. In contrast, our work separates the declared planning mode from the execution architecture and asks whether the declared mode is actually preserved in the trajectory.

\noindent\textbf{Agent Benchmarks and Process-Level Evaluation.}
Agent benchmarks evaluate LLM agents in increasingly realistic multi-step environments, including web navigation~\cite{deng2023mind2web,zhou2024webarena,pan2024webcanvas}, computer and operating-system control~\cite{xie2024osworld,merrill2026terminal}, software engineering~\cite{jimenez2024swe}, embodied tasks~\cite{shridharalfworld}, and general assistant tasks~\cite{mialon2024gaia}. These benchmarks are valuable for measuring whether agents complete tasks, but they usually report final success, completion rate, or answer correctness rather than the planning process that produced the outcome. AgentBoard moves toward process-level evaluation by introducing progress-rate metrics and analytical visualizations for multi-turn agents~\cite{ma2024agentboard}. However, progress metrics still do not determine whether an agent's declared planning mode was preserved across planner--executor handoffs. Our work complements benchmark-level evaluation by asking not only whether an agent succeeds, but whether success or failure can be attributed to the planning mode that was declared and the execution path that actually ran.

\noindent\textbf{Trajectory Diagnosis and Failure Attribution.}
Recent work has begun to analyze agent trajectories beyond final success. AgentDiagnose provides a toolkit for diagnosing LLM-agent trajectories using competency-oriented metrics such as task decomposition, backtracking and exploration, observation reading, self-verification, and objective quality~\cite{ou2025agentdiagnose}. Other work studies agent-environment interaction failures and proposes taxonomies for where agents fail in realistic environments~\cite{kong2025aegis}, while recent multi-agent analyses identify failure modes such as specification and system-design failures, inter-agent misalignment, and task verification or termination errors~\cite{cemrimulti}. These studies make agent failures more interpretable, but they generally analyze trajectories after execution without explicitly modeling the handoff between declared planning mode and execution architecture. In contrast, our work introduces \emph{handoff-aware failure attribution}: we map each trajectory through the planner, router, executor, and verifier, allowing failures to be localized to plan declaration, plan-to-executor handoff, execution, or verification. This lets us distinguish failures caused by an inappropriate planning mode from failures caused by an executor that did not preserve the declared mode.

\FloatBarrier

\section{Planning Patterns}
\label{app:planning_patterns}

\begin{figure*}[!ht]
    \centering
    \begin{minipage}{\textwidth}
        \centering
        \includegraphics[width=\linewidth]{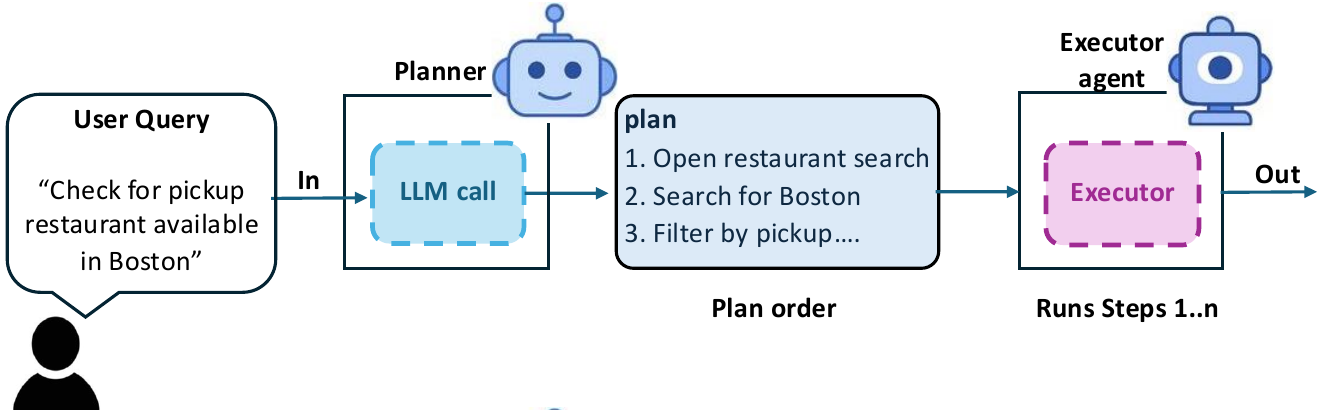}
        (a) Predefined Planning
    \end{minipage}
    \vspace{0.5cm}
    \begin{minipage}{\textwidth}
        \centering
        \includegraphics[width=\linewidth]{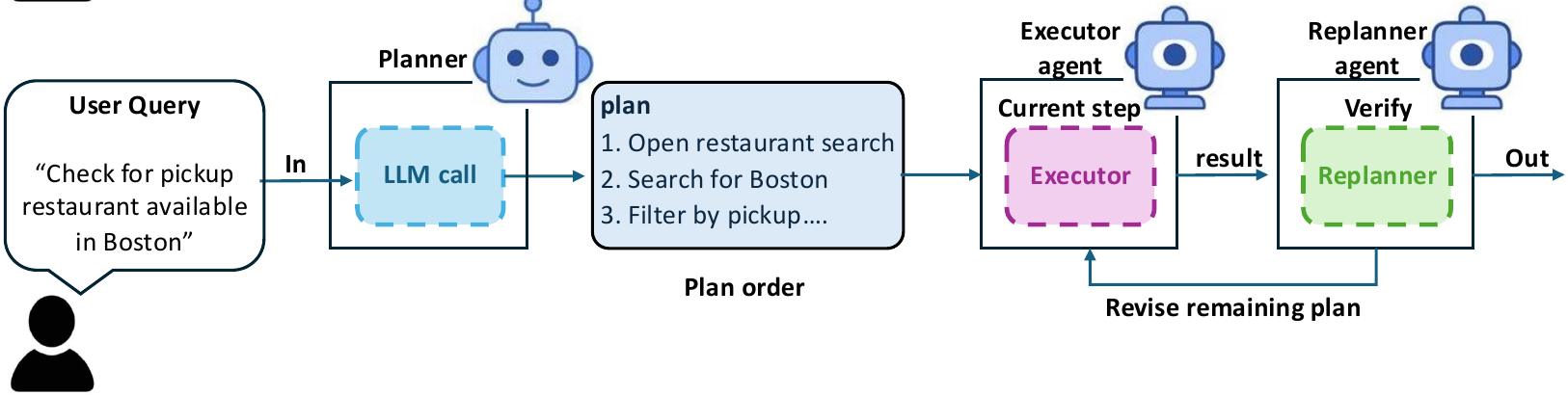}
        (b) Sequential Planning
    \end{minipage}
    \vspace{0.5cm}
    \begin{minipage}{\textwidth}
        \centering
        \includegraphics[width=\linewidth]{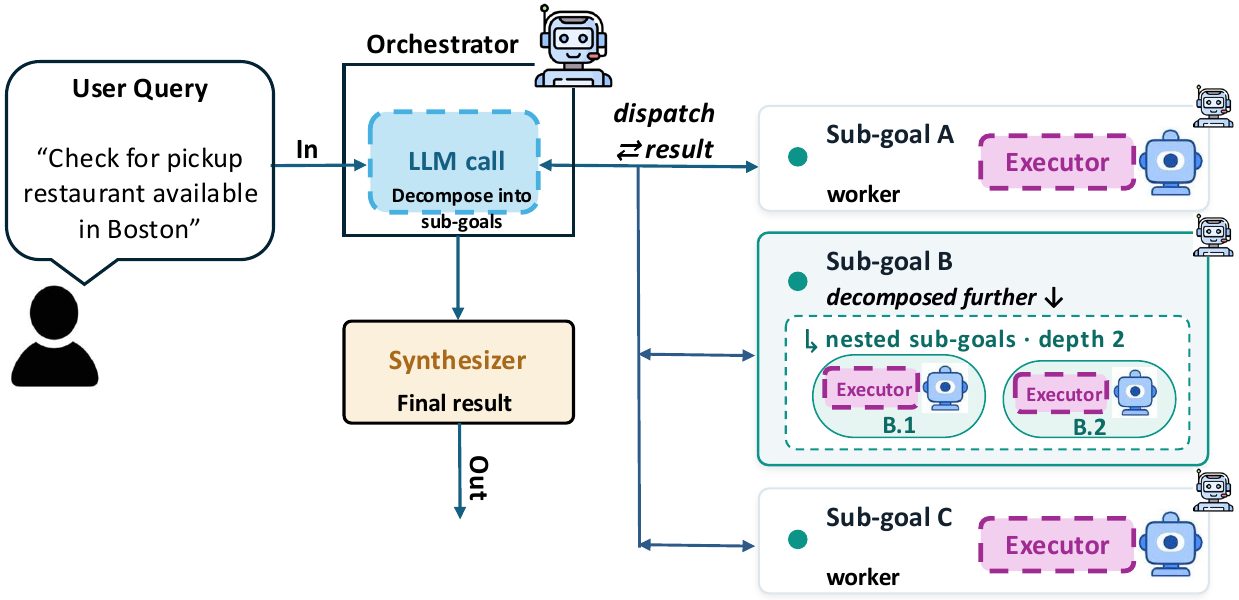}
        (c) Hierarchical Planning
    \end{minipage}
    \caption{Illustration of three planning executor patterns. 
(a) \textbf{Predefined planning}: the planner generates a fixed ordered plan and the executor follows the steps without replanning. 
(b) \textbf{Sequential planning}: the executor runs the current step, observes the result, and invokes a replanner to revise the remaining plan when needed. 
(c) \textbf{Hierarchical planning}: an orchestrator decomposes the task into subgoals, dispatches them to workers, and synthesizes the final result.
}
    \label{fig:planning_patterns}
\end{figure*}

We consider four planning patterns: \textit{predefined}, \textit{sequential}, \textit{hierarchical}, and \textit{search}. These patterns capture different ways an agent can organize task execution. A \textit{predefined} plan follows a fixed plan generated before execution, without replanning (Fig.~\ref{fig:planning_patterns} (a)). A \textit{sequential} plan executes one step at a time and updates the remaining plan based on intermediate observations (Fig.~\ref{fig:planning_patterns} (b)). A \textit{hierarchical} plan decomposes the task into subgoals and coordinates their execution through an orchestrator--worker structure (Fig.~\ref{fig:planning_patterns} (c)). A \textit{search} plan generates multiple candidate plans, executes each candidate independently, and selects the most promising resulting trajectory using a \emph{rubric-based judge} (Fig.~\ref{fig:planning_patterns} (d)).

\begin{figure*}[!ht]
    \centering
    \begin{minipage}{\textwidth}
        \centering
        \includegraphics[width=\linewidth]{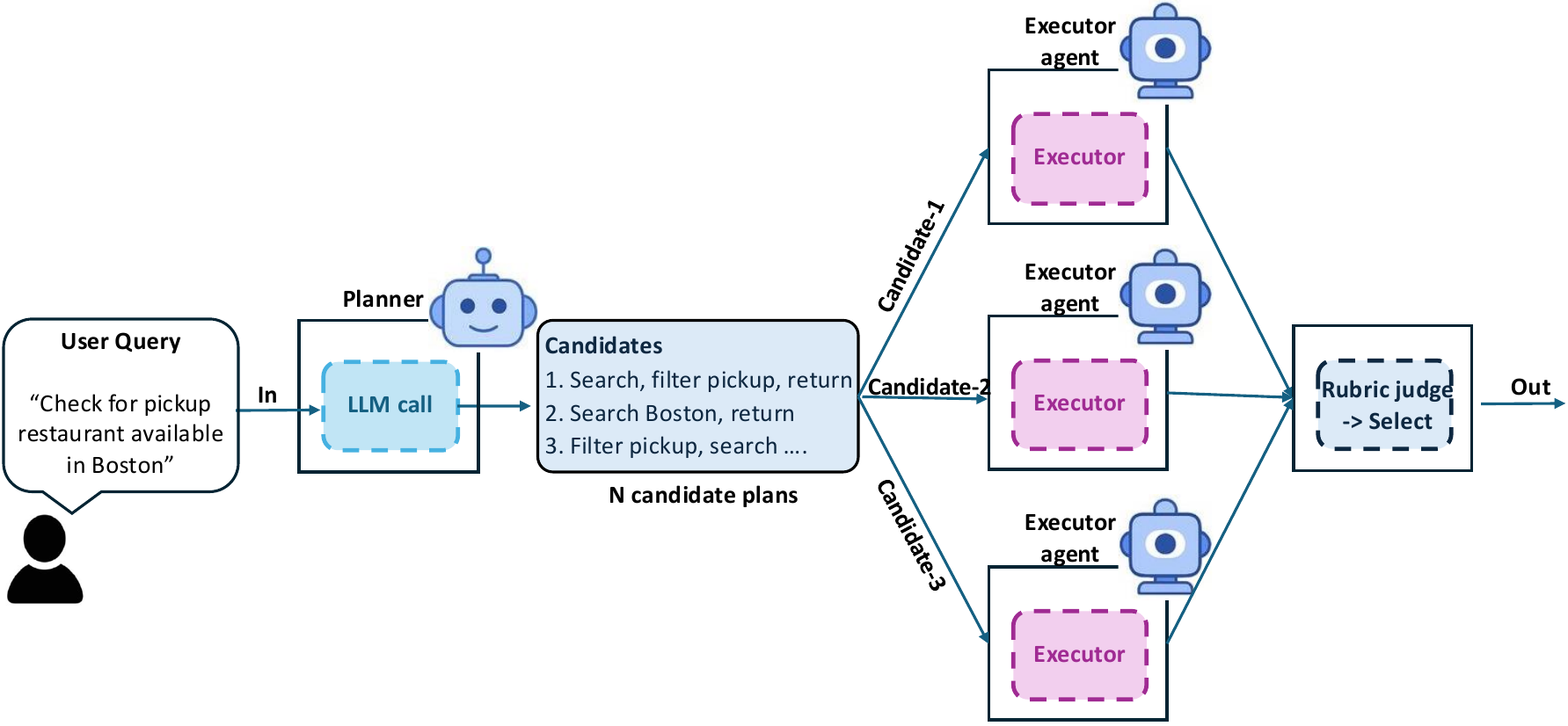}
        (a) Search Planning
    \end{minipage}
    \vspace{0.5cm}
    \begin{minipage}{\textwidth}
        \centering
        \includegraphics[width=\linewidth]{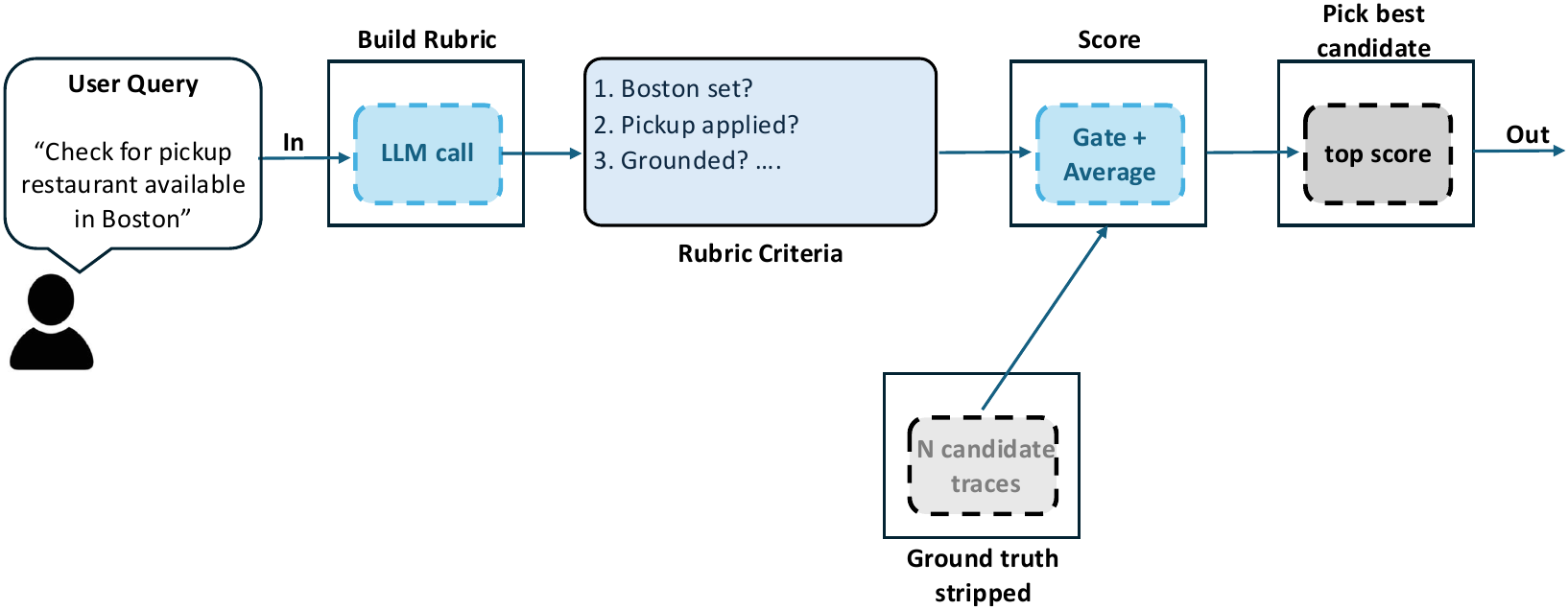}
        (b) Pick best candidate using Rubric judge
    \end{minipage}
    \caption{Illustration of search planning pattern. 
(a) \textbf{Search-based planning}: The planner generates multiple competing candidate plans, and each
candidate is executed independently from a reset environment.
(b) \textbf{Rubric judge}:  A task-specific rubric judge evaluates the resulting trajectory summaries against a rubric generated from the task specification and selects the highest-scoring candidate. The judge has no access to benchmark ground-truth success labels or hidden evaluation scores.
}
\label{fig:search_planning_patterns}
\end{figure*}

\section{Dataset and Model Details} 
\label{app:dataset-details} 

\paragraph{Benchmarks.} We evaluate our framework across four agent benchmarks spanning web interaction, software engineering, and embodied task execution. Table~\ref{tab:dataset-details} summarizes the number of evaluated tasks, task characteristics, and evaluation metric used for each benchmark.

\begin{table}[!ht] 
\centering 
\small 
\caption{Agent benchmarks used in our experiments. $N$ denotes the number of tasks evaluated in this work.} \label{tab:dataset-details} 
\resizebox{\textwidth}{!}{
\begin{tabular}{|l| l| r| l| l|} 
\hline 
\textbf{Benchmark} & \textbf{Environment} & \textbf{$N$} & \textbf{Task type} & \textbf{Evaluation} \\ 
\hline 
Mind2Web & Web interaction & 1,341 & Multi-step website interaction & Benchmark task success \\ 
WebArena & Web interaction & 204 & Navigation, retrieval, and website interaction & Benchmark task success \\ 
SWE-bench Verified & Software engineering & 500 & GitHub issue resolution and code modification & Patch-based evaluation \\ 
ALFWorld & Embodied environment & 134 & Multi-step household tasks & Goal completion \\ 
\hline 
\end{tabular}} 
\end{table}

\paragraph{Backbone models.} We evaluate three backbone LLMs: Qwen3.6-35B-A3B, DeepSeek-V4-Flash, and Gemma-4-26B-A4B-it. The same backbone is used for planning-mode declaration and planning patterns within a given experimental run. Unless otherwise specified, model identity is held fixed when comparing execution conditions. 

\begin{table}[!ht] 
\centering 
\small 
\caption{Backbone LLMs used in our experiments.} \label{tab:model-details} 
\begin{tabular}{|l|l|c|} 
\hline 
\textbf{Model} & \textbf{Role} & \textbf{Thinking} \\ \hline 
Qwen3.6-35B-A3B & Declaration and execution & Enabled \\ DeepSeek-V4-Flash & Declaration and execution & Enabled \\ Gemma-4-26B-A4B-it & Declaration and execution & Enabled \\ 
\hline 
\end{tabular} 
\end{table}

\begin{table*}[!ht]
\centering
\scriptsize
\setlength{\tabcolsep}{6pt}
\renewcommand{\arraystretch}{1.12}
\caption{
Composition of the evaluation sets used in our experiments. ALFWorld tasks are grouped by task type, with the mean number of expert demonstration steps shown for each category. Mind2Web tasks are grouped by the benchmark's three non-overlapping generalization splits; together they account for all 1,341 evaluation tasks. SWE-bench Verified instances are grouped by the dataset's time-to-fix difficulty label.
}
\label{tab:benchmark-composition}
\begin{tabular}{|l|l|r|l|l|}
\hline
\textbf{Benchmark} &
\textbf{Group} &
\textbf{Tasks} &
\textbf{Auxiliary statistic} &
\textbf{Additional information} \\
\hline
\multirow{6}{*}{ALFWorld}
& Pick \& Place       & 24 & Mean expert steps: 4.58 & \\
& Examine in Light    & 18 & Mean expert steps: 3.78 & \\
& Clean \& Place      & 31 & Mean expert steps: 6.32 & \\
& Heat \& Place       & 23 & Mean expert steps: 6.04 & \\
& Cool \& Place       & 21 & Mean expert steps: 6.10 & \\
& Pick Two \& Place   & 17 & Mean expert steps: 8.65 & \\
\cmidrule(lr){2-5}
& \textit{Total}      & \textit{134} & & \\
\hline

\multirow{3}{*}{Mind2Web}
& Cross-Task
& 252
& 69 websites
& Travel 119; Shopping 68; Entertainment 65 \\

& Cross-Website
& 177
& 10 websites
& Shopping 63; Travel 60; Entertainment 54 \\

& Cross-Domain
& 912
& 54 websites
& Information 480; Service 432 \\
\cmidrule(lr){2-5}
& \textit{Total}
& \textit{1,341}
& & \\
\hline

\multirow{4}{*}{SWE-bench Verified}
& $<15$ min
& 194
& Time-to-fix difficulty
& \\

& 15 min--1 hour
& 261
& Time-to-fix difficulty
& \\

& 1--4 hours
& 42
& Time-to-fix difficulty
& \\

& $>4$ hours
& 3
& Time-to-fix difficulty
& \\
\cmidrule(lr){2-5}
& \textit{Total}
& \textit{500}
& & \\
\hline
\end{tabular}
\end{table*}

\section{Inference and Sampling Configuration}
\label{app:inference-config}

For all experiments, we use a shared inference harness across execution conditions. For a given backbone model, we keep the decoding parameters, thinking configuration, tool-calling policy, and generation budget fixed when comparing Flat ReAct, Plan+ReAct, and Planning-as-Routing. This controls for inference-level differences when comparing planning and execution mechanisms. 

\paragraph{Generation profiles.}
We define one common generation-budget profile that uses a maximum budget of 32,768 new tokens per turn. For thinking models, we also enforce a minimum per-turn budget of 16,384 new tokens. The effective budget for a model is computed as:

\[
\max(\texttt{min\_floor}, \min(\texttt{model\_budget}, \texttt{profile\_budget})).
\]
We use a single total generation budget per model and do not impose a separate reasoning-token cap. If a model consumes the full budget and terminates with a length cutoff, the run is counted as a truncation failure rather than silently capped.

\paragraph{Thinking and context handling.}
Thinking mode is enabled for all models. By default, prior thinking traces are not passed back into the model context, which reduces context length and improves comparability across models. 

\paragraph{Tool-calling policy.}
For most models, tool use is forced during the first executor turns by setting the tool choice to required. This encourages the executor to interact with the environment rather than only producing text. 
All such backend-specific exceptions are fixed before evaluation and applied consistently across all conditions for the affected model.

\begin{table}[!ht]
\centering
\scriptsize
\caption{Sampling parameters used for each backbone LLM. Disabled settings are not sent to the backend.}
\label{tab:sampling-config}
\begin{tabular}{|l|c|c|c|c|c|c|}
\hline
Model & Temp. & Top-$p$ & Top-$k$ & Pres. Pen. & Rep. Pen. & Thinking \\
\hline
Qwen3.6-35B-A3B & 1.0 & 0.95 & 20 & 1.5 & 1.0 & Yes \\
Gemma4-26B-A4B-it & 1.0 & 0.95 & 64 & 0.0 & 1.0 & Yes \\
DeepSeek-V4-Flash & 1.0 & 0.95 & Disabled & 0.0 & 1.0 & Yes \\
\hline
\end{tabular}
\end{table}

\begin{table}[!ht]
\centering
\scriptsize
\begin{tabular}{|l|c|c|c|c|}
\hline
Model & Model Budget & Default Profile & Hard Profile & Tool Choice \\
\hline
Qwen3.6-35B-A3B & 81,920 & 32,768 & 81,920 & Required \\
Gemma-4-26B-A4B-it & 131,072 & 32,768 & 81,920 & Required \\
DeepSeek-V4-Flash & 81,920 & 32,768 & 81,920 & Required \\
\hline
\end{tabular}
\caption{Per-turn generation budgets and tool-calling configuration. The effective budget is the minimum of the model budget and the selected profile budget, subject to the minimum thinking floor.}
\label{tab:budget-config}
\end{table}

\paragraph{Model-specific decoding settings.}
For each backbone model, we adopt the recommended inference and decoding configuration provided in the model developer's official release or documentation, rather than tuning these parameters on our evaluation benchmarks. Accordingly, the Qwen models use temperature $1.0$, top-$p=0.95$, and top-$k=20$; Gemma-4-26B-A4B-it uses temperature $1.0$, top-$p=0.95$, and top-$k=64$; and DeepSeek-V4-Flash uses temperature $1.0$ and top-$p=0.95$, with top-$k$ disabled. These model-specific configurations are fixed across all experimental
conditions.

\paragraph{Search candidate budget.}
The Search planning pattern where the planner dynamically chooses how many candidate plans to generate for each task, subject to an upper bound of four candidates (\texttt{max\_candidates}=4). Thus, the realized candidate count can vary from 0 to 4 depending on the planner's decision for the task. Across all Search runs, the realized candidate count is $3.35\pm0.49$ over 15,598 episodes. At the benchmark level, the mean is $3.18\pm0.41$ for ALFWorld,
$3.31\pm0.46$ for Mind2Web, $3.38\pm0.52$ for SWE-bench, and
$3.88\pm0.33$ for WebArena. Table~\ref{tab:search-candidate-counts} provides the model- and seed-level breakdown.

\begin{table*}[!ht]
\centering
\scriptsize
\setlength{\tabcolsep}{5pt}
\renewcommand{\arraystretch}{1.10}
\caption{
Realized number of Search candidates per task. ``Episode mean'' reports mean$\pm$SD across individual Search episodes, while ``Across seeds'' reports mean$\pm$SD of the seed-level means. Search permits at most four candidate plans per task.
}
\label{tab:search-candidate-counts}
\begin{tabular}{|l|l|c|c|c|c|}
\hline
\textbf{Benchmark} &
\textbf{Model} &
\textbf{Episode mean} &
\textbf{Across seeds} &
\textbf{Range} \\
\hline

\multirow{3}{*}{ALFWorld}
& DeepSeek-V4   &  $3.33\pm0.47$ & $3.34\pm0.05$ & 3--4 \\
& Qwen3.6-35B   &  $3.22\pm0.45$ & $3.22\pm0.06$ & 0--4 \\
& Gemma-4-26B   &  $2.99\pm0.11$ & $2.99\pm0.01$ & 2--4 \\
\hline

\multirow{3}{*}{Mind2Web}
& DeepSeek-V4   &  $3.47\pm0.50$ & $3.47\pm0.01$ & 2--4 \\
& Qwen3.6-35B   &  $3.35\pm0.48$ & $3.35\pm0.02$ & 0--4 \\
& Gemma-4-26B   &  $3.00\pm0.03$ & $3.00\pm0.00$ & 2--3 \\
\hline

\multirow{3}{*}{SWE-bench}
& DeepSeek-V4   &  $3.26\pm0.44$ & $3.26\pm0.00$ & 3--4 \\
& Qwen3.6-35B   &  $3.76\pm0.43$ & $3.76\pm0.02$ & 3--4 \\
& Gemma-4-26B   &  $2.92\pm0.27$ & --            & 2--3 \\
\hline

\multirow{2}{*}{WebArena}
& DeepSeek-V4   &  $4.00\pm0.06$ & $4.00\pm0.01$ & 3--4 \\
& Qwen3.6-35B   &  $3.76\pm0.43$ & $3.76\pm0.01$ & 3--4 \\
\hline
\end{tabular}
\end{table*}

\begin{table}[!ht]
\centering
\small
\caption{Execution and planning-structure budgets used in each benchmark.
All planning-structure limits are fixed across models and random seeds.}
\label{tab:execution-budgets}
\begin{tabular}{|l|c|c|c|c|}
\hline
Benchmark & Max env. steps & \texttt{max\_steps} &
\texttt{max\_branch} & \texttt{max\_depth} \\
\hline
ALFWorld  & $24$ & 8 & 4 & 3 \\
Mind2Web  & $24$ & 8 & 4 & 3 \\
WebArena  & $24$ & 8 & 4 & 3 \\
SWE-bench & $24$ & 8 & 4 & 3 \\
\hline
\end{tabular}
\end{table}

\section{Plan-Structure Verifier and Human Validation}
\label{app:plan_structure_verifier}

\subsection{Plan-Structure Verifier}
\label{app:verifier}

We measure whether Plan+ReAct preserves the declared plan structure using a
deterministic rule-based verifier. Declared plan units are first filtered to
remove non-actionable text and are then matched to executed actions using
benchmark-specific intent and target information. Structure is considered
maintained only if all scorable plan units can be matched in their declared
order; additional environment actions between matched units are allowed.
Matching is computed using a monotonic dynamic-programming assignment rather
than greedy matching. Units without executable actions are treated as unscorable, and trajectories
with no scorable or matchable units are excluded and reported separately.

The verifier measures structural preservation rather than action correctness: whether an action successfully changes the environment is evaluated separately. Pattern-specific routed executors are not scored with this matcher because their dispatch records establish the executed control flow by construction. Representative examples of this matching procedure for the four declared planning modes are provided below.



\subsection{Representative Plan+ReAct Traces}
\label{app:planning-examples}

The following examples illustrate how the structure verifier compares the declared plan with the sequence of environment actions produced by the generic Plan+ReAct executor. Extra environment actions are allowed between declared units; structure is preserved when all scorable plan units are matched in the declared order.

\begin{planexample}[Sequential]{sequentialcolor}
\textbf{Declared plan}
\begin{enumerate}[leftmargin=*,nosep]
    \item Look around and identify the location of a pillow.
    \item Go to that location and take the pillow.
    \item Go to the sofa and put the pillow on the sofa.
    \item Look around and identify the location of the second pillow.
    \item Go to that location and take the second pillow.
    \item Go to the sofa and put the second pillow on the sofa.
\end{enumerate}

\textbf{Executed actions}
\begin{enumerate}[leftmargin=*,nosep]
    \item \texttt{go to sofa 1}
    \item \texttt{go to armchair 1}
    \item \texttt{take pillow 1 from armchair 1}
    \item \texttt{take pillow 2 from armchair 1}
    \item \texttt{inventory}
    \item \texttt{go to sofa 1}
    \item \texttt{put pillow 1 on sofa 1}
    \item \texttt{move pillow 1 to sofa 1}
    \item \texttt{go to armchair 1}
    \item \texttt{take pillow 2 from armchair 1}
    \item \texttt{go to sofa 1}
    \item \texttt{move pillow 2 to sofa 1}
\end{enumerate}
\end{planexample}

\begin{planexample}[Predefined]{predefinedcolor}
\textbf{Declared plan}
\begin{enumerate}[leftmargin=*,nosep]
    \item Go to countertop.
    \item Take mug from countertop.
    \item Go to sink.
    \item Clean mug.
    \item Go to coffeemachine.
    \item Put mug in coffeemachine.
\end{enumerate}

\textbf{Executed actions}
\begin{enumerate}[leftmargin=*,nosep]
    \item \texttt{go to countertop 1}
    \item \texttt{go to countertop 2}
    \item \texttt{take mug 1 from countertop 2}
    \item \texttt{go to sinkbasin 1}
    \item \texttt{clean mug 1 with sinkbasin 1}
    \item \texttt{go to coffeemachine 1}
    \item \texttt{move mug 1 to coffeemachine 1}
\end{enumerate}
\end{planexample}

\begin{planexample}[Hierarchical]{hierarchicalcolor}
\textbf{Declared plan}
\begin{enumerate}[leftmargin=*,nosep]
    \item \textbf{Acquire peppershaker:} go to countertop $\rightarrow$ take peppershaker.
    \item \textbf{Place in drawer:} go to drawer $\rightarrow$ open drawer $\rightarrow$ put peppershaker in drawer.
\end{enumerate}

\textbf{Executed actions}
\begin{enumerate}[leftmargin=*,nosep]
    \item \texttt{go to countertop 1}
    \item \texttt{take peppershaker 1 from countertop 1}
    \item \texttt{go to drawer 1}
    \item \texttt{open drawer 1}
    \item \texttt{put peppershaker 1 in drawer 1}
    \item \texttt{move peppershaker 1 to drawer 1}
\end{enumerate}
\end{planexample}

\begin{planexample}[Search]{searchcolor}
\textbf{Declared plan}
\begin{enumerate}[leftmargin=*,nosep]
    \item \textbf{Path 1: Tomato in fridge.}
    Go to fridge $\rightarrow$ open fridge $\rightarrow$ take tomato
    $\rightarrow$ go to microwave $\rightarrow$ heat tomato
    $\rightarrow$ go to garbage can $\rightarrow$ put tomato in garbage can.
    
    \item \textbf{Path 2: Tomato on countertop.}
    Go to countertop $\rightarrow$ take tomato
    $\rightarrow$ go to microwave $\rightarrow$ heat tomato
    $\rightarrow$ go to garbage can $\rightarrow$ put tomato in garbage can.
\end{enumerate}

\textbf{Executed actions}
\begin{enumerate}[leftmargin=*,nosep]
    \item \texttt{go to fridge 1}
    \item \texttt{open fridge 1}
    \item \texttt{go to countertop 1}
    \item \texttt{take tomato 1 from countertop 1}
    \item \texttt{go to microwave 1}
    \item \texttt{open microwave 1}
    \item \texttt{put tomato 1 in microwave 1}
    \item \texttt{move tomato 1 to microwave 1}
    \item \texttt{heat tomato 1 with microwave 1}
    \item \texttt{close microwave 1}
    \item \texttt{open microwave 1}
    \item \texttt{examine microwave 1}
    \item \texttt{use microwave 1}
    \item \texttt{close microwave 1}
    \item \texttt{use microwave 1}
    \item \texttt{inventory}
    \item \texttt{open microwave 1}
    \item \texttt{take tomato 1 from microwave 1}
    \item \texttt{heat tomato 1 with microwave 1}
    \item \texttt{examine tomato 1}
\end{enumerate}
\end{planexample}

\subsection{Human Validation of the Structure Verifier}
\label{app:human_validation_verifier}

We validate the rule-based verifier against two independent human annotators, denoted A1 and A2, on (100) sampled Plan+ReAct trajectories from ALFWorld. The sample is stratified across the three evaluated backbone models: DeepSeek-V4, Qwen3.6-35B, and Gemma-4-26B, and across the verifier outcome types. Both annotators independently evaluated all (100) trajectories and judged whether the executed action sequence preserved the declared plan structure.

\paragraph{Human--human agreement.}
Across the (100) trajectories, A1 and A2 achieved (71\%) raw agreement and Cohen's ($\kappa$=0.35) on the binary structure-maintained judgment. The agreement varied between models: Gemma-4-26B was achieved ($\kappa$ = 0.42), DeepSeek-V4 ($\kappa$ = 0.37) and Qwen3.6-35B ($\kappa$ = 0.19) ((n=18) for the Qwen3.6 subset), indicating that deciding whether a free-form ReAct trajectory preserves a declared plan can itself be ambiguous for human annotators.

\paragraph{Verifier--human agreement.}
The rule-based verifier shows similar agreement with both annotators: ($\kappa$=0.31) against A1 and ($\kappa$=0.34) against A2. These values are comparable to the human--human agreement of ($\kappa$=0.35), suggesting that disagreement with the verifier is of similar magnitude to disagreement between independent human judgments rather than being dominated by one annotator or a systematic verifier bias.

\section{Plan Quality Judge}
\label{app:plan_quality_judge}

\paragraph{Judge model.} 
Plan quality and plan adherence are scored with a fixed LLM judge. For the DeepSeek-V4-Flash and Qwen3.6-35B ALFWorld sweeps, we use \texttt{Gemma-4-26B-A4B-it} for every cell, ensuring that trajectories are not evaluated by the model that produced them and that mode- and model-level comparisons use a common judging scale. The judge identity is stored explicitly in each output. Every task in each cell is judged without subsampling, and each metric is evaluated in a separate call.

For Gemma-4-26B agent runs, self-judging is avoided by using
DeepSeek-V4-Flash as the judge. Scores produced by different judge
models are therefore not pooled; comparisons of plan-quality scores
are made within a common judging configuration.

\begin{promptbox}{Plan Quality: goal + initial plan only; execution withheld}
\small
You are a meticulous and analytical PLAN QUALITY evaluator.

Your task is to evaluate the intrinsic quality of the initial written plan
using only:
(1) the user's goal,
(2) the tools available when the plan was created, and
(3) the initial plan itself.

CRITICAL: You must not use or reference execution information. Do not use
agent actions, tool outputs, observations, errors, replans, or the final
answer to judge the quality of the initial plan. You are evaluating whether
the plan was a good strategy when it was written, not whether it eventually
succeeded.

Plan Extraction Procedure:
1. Scan for the first section explicitly labeled with a PLAN keyword.
   This is the initial plan.
2. If no explicitly labeled PLAN section exists, infer the plan from the
   initial Thinking or planning section.
3. If no plan can be identified, output: "I cannot find a plan."
4. Do not infer missing plan steps from the execution trace.

TOOLS

Evaluate the initial plan according to the following criteria:

1. Goal Coverage:
Does the plan address all important parts of the user's goal?
Are any necessary sub-tasks missing?

2. Tool Selection:
Does the plan select appropriate tools from those available?
Does it ignore a clearly more suitable or efficient tool?
Does it propose using a tool that does not exist?

3. Tool Feasibility:
Are the planned tool calls consistent with the tools' descriptions,
capabilities, and required inputs?

4. Step Structure:
Are the planned steps clear, actionable, and ordered logically?
Are there unnecessary, redundant, or unsupported steps?

5. Use of Available Information:
Does the plan avoid unnecessary work when the required information is
already provided in the goal or context?

6. Efficiency:
Is the plan a reasonable and efficient strategy given the available
resources, rather than merely a theoretically possible sequence of actions?

List only inherent flaws in the written plan. Do not infer flaws from later
execution outcomes.

You must assign a single numerical score from 0 to 3:

3: The plan is well-structured, feasible, efficient, and directly addresses
the goal. Necessary steps are present, logically ordered, and compatible
with the available tools. No important unnecessary or unsupported steps are
present.

2: The plan generally addresses the goal and is feasible, but contains minor
issues such as unclear steps, small omissions, unnecessary actions, weak
ordering, or insufficient detail.

1: The plan partially addresses the goal but contains substantial omissions,
inefficiencies, unsupported assumptions, or tool-use problems that could
prevent successful completion.

0: The plan does not meaningfully address the goal or is infeasible. Critical
steps are missing, irrelevant, unsupported, or rely on unavailable or
incorrectly used tools.

Be critical. For every identified issue, refer to the relevant plan step
where possible and explain the problem specifically.

Please respond using exactly the following template:

Initial Plan Identification
[Paste the initial plan or state: "I cannot find a plan."]

Plan Quality Analysis
[Evaluate the plan using only the goal, available tools, and written plan.]

Verdict on Plan Flaws
[List only intrinsic flaws in the written plan.]

Criteria:
<Briefly restate the evaluation criteria applied.>

Supporting Evidence:
<Explain the score, tied to specific plan steps and criteria.>

Score:
$<$0, 1, 2, or 3$>$
\end{promptbox}

\begin{promptbox}{Plan Adherence: declared plan + execution trajectory}
\small
You are a meticulous and analytical PLAN ADHERENCE evaluator.

Your task is to evaluate how faithfully the execution trajectory completes
the executable units of the declared plan.

You are given:
(1) the user's goal,
(2) the declared plan, and
(3) the execution trajectory, including actions and observations.

Evaluate adherence to the declared plan, not the intrinsic quality of the
plan and not final task success. A plan may be poor but followed faithfully,
or good but only partially executed.

Plan Extraction Procedure:
1. Identify the initial declared plan.
2. Decompose it into executable plan units.
3. Ignore purely explanatory, motivational, or non-actionable text.
4. For hierarchical plans, evaluate executable leaf-level units.
5. For each executable unit, determine whether the trajectory provides
   sufficient evidence that the unit was completed.
6. Extra actions, retries, or intermediate environment interactions do not
   count against adherence unless they replace or prevent completion of a
   declared plan unit.

Evaluate the execution according to the following criteria:

1. Plan-unit completion:
How many executable units of the declared plan are actually completed?

2. Coverage:
Does the trajectory complete all important executable parts of the plan,
or are substantial planned units omitted?

3. Execution evidence:
Are completed units supported by concrete actions or observations in the
trajectory rather than merely mentioned in reasoning text?

4. Replanning:
If the executor explicitly revises the remaining plan, judge adherence
relative to the active plan produced by that replanning step. Do not penalize
a valid revision merely because later actions differ from superseded units.

Do not score task correctness, efficiency, or plan quality. Do not infer that
a plan unit was completed solely because the overall task succeeded.

You must assign a single numerical score from 0 to 3:

3: The executable plan is followed almost completely. All or nearly all
important plan units are completed and are supported by trajectory evidence.

2: The trajectory follows most of the executable plan, but one or more
meaningful units are omitted, only partially completed, or weakly supported.

1: The trajectory follows only a small portion of the declared plan.
Several important executable units are skipped, abandoned, or replaced by
actions not corresponding to the plan.

0: The execution does not meaningfully follow the declared plan, or no
executable plan unit can be identified as completed.

Please respond using exactly the following template:

Declared Plan
[Paste or summarize the executable units of the declared plan.]

Plan Adherence Analysis
[Identify which plan units were completed, partially completed, or omitted,
using evidence from the trajectory.]

Criteria:
<Briefly restate the adherence criteria applied.>

Supporting Evidence:
<Tie the judgment to specific declared plan units and trajectory actions or
observations.>

Score:
$<$0, 1, 2, or 3$>$
\end{promptbox}

\section{Structure Preservation by Declared Planning Mode}
\label{app:structure-bymode}

\paragraph{Hierarchical/Search plans are particularly difficult for generic ReAct execution to preserve.}
We next analyze structural maintenance by planning mode, the planning mode-level breakdown in Table~\ref{tab:structure-bymode}, show that the declaration--execution gap is most pronounced for richer planning structures. Hierarchical declarations are preserved poorly across benchmarks: on ALFWorld, structure maintenance is only
$11.4\%$, $4.6\%$, and $0.8\%$ for DeepSeek-V4, Qwen3.6-35B, and
Gemma-4-26B, respectively; on Mind2Web, the corresponding values are
$14.5\%$, $17.3\%$, and $22.6\%$. SWE-bench shows the same pattern, with Hierarchical maintenance of $17.3\%$ and $14.5\%$ for DeepSeek-V4 and Qwen3.6-35B. Search is similarly difficult to preserve where enough declarations are observed, although several Search cells are sparsely populated and its competing-candidate structure is not directly comparable to a single ordered plan. Together with the plan-length analysis, these results indicate that a shared generic Plan+ReAct executor is particularly unreliable at maintaining richer multi-level or multi-candidate planning structures.

Table~\ref{tab:structure-bymode} provides the mode-level breakdown of Plan+ReAct structure preservation. Hierarchical declarations exhibit consistently low structure maintenance across benchmarks, while Search is also difficult to preserve where sufficient declarations are available. These results complement the main-text plan-length analysis and show that generic ReAct execution is particularly poorly suited to richer multi-level or multi-candidate planning structures.

\begin{table*}[!ht]
\centering
\scriptsize
\setlength{\tabcolsep}{4pt}
\caption{
Plan+ReAct structure preservation by declared planning mode.
Hierarchical declarations exhibit consistently low structure maintenance across benchmarks, while Search is also difficult to preserve where enough declarations are available. $n$ denotes the smallest per-seed count when multiple seeds are available; values are mean $\pm$ standard deviation across seeds. $^{\ddagger}$ indicates cells with fewer than 15 declarations in at least one seed and should be interpreted cautiously. Search is included for
completeness, but its competing-candidate structure is not directly comparable to a single ordered Sequential, Predefined, or Hierarchical plan.
}
\label{tab:structure-bymode}
\renewcommand{\arraystretch}{1.08}
\begin{tabular}{|l|l|l|c|c|l|}
\hline
\textbf{Benchmark} &
\textbf{Model} &
\textbf{Declared mode} &
\textbf{$n$} &
\textbf{Structure maintained} &
\textbf{Steps declared} \\
\hline

\multirow{8}{*}{ALFWorld}
& DeepSeek-V4
& Sequential
& $107$/seed
& $30.5{\pm}5.0\%$
& $5.3{\pm}0.1$ \\

& DeepSeek-V4
& Predefined$^{\ddagger}$
& $2$
& $0.0\%$
& $5.5$ \\

& DeepSeek-V4
& Hierarchical$^{\ddagger}$
& $5$/seed
& $11.4{\pm}8.4\%$
& $9.5{\pm}0.3$ \\

& DeepSeek-V4
& Search$^{\ddagger}$
& $5$/seed
& $0.0{\pm}0.0\%$
& $16.8{\pm}0.9$ \\

& Qwen3.6-35B
& Sequential
& $89$/seed
& $30.5{\pm}2.2\%$
& $5.0{\pm}0.1$ \\

& Qwen3.6-35B
& Hierarchical
& $28$/seed
& $4.6{\pm}0.8\%$
& $9.1{\pm}0.4$ \\

& Gemma-4-26B
& Sequential
& $36$/seed
& $49.4{\pm}0.6\%$
& $4.3{\pm}0.1$ \\

& Gemma-4-26B
& Hierarchical
& $41$/seed
& $0.8{\pm}0.8\%$
& $10.0{\pm}0.2$ \\
\hline

\multirow{12}{*}{Mind2Web}
& DeepSeek-V4
& Sequential
& $808$/seed
& $56.6{\pm}0.9\%$
& $4.1{\pm}0.1$ \\

& DeepSeek-V4
& Predefined
& $363$/seed
& $29.0{\pm}1.5\%$
& $4.7{\pm}0.0$ \\

& DeepSeek-V4
& Hierarchical
& $104$/seed
& $14.5{\pm}0.3\%$
& $9.8{\pm}0.2$ \\

& DeepSeek-V4
& Search$^{\ddagger}$
& $3$/seed
& $0.0{\pm}0.0\%$
& $10.4{\pm}0.9$ \\

& Qwen3.6-35B
& Sequential
& $678$/seed
& $51.0{\pm}0.4\%$
& $3.6{\pm}0.0$ \\

& Qwen3.6-35B
& Predefined
& $64$/seed
& $25.8{\pm}4.8\%$
& $4.1{\pm}0.1$ \\

& Qwen3.6-35B
& Hierarchical
& $481$/seed
& $17.3{\pm}1.8\%$
& $6.6{\pm}0.1$ \\

& Qwen3.6-35B
& Search$^{\ddagger}$
& $3$/seed
& $6.7{\pm}9.4\%$
& $5.8{\pm}0.6$ \\

& Gemma-4-26B
& Sequential
& $182$/seed
& $69.8{\pm}3.3\%$
& $2.8{\pm}0.0$ \\

& Gemma-4-26B
& Predefined$^{\ddagger}$
& $8$/seed
& $15.1{\pm}2.6\%$
& $4.0{\pm}0.1$ \\

& Gemma-4-26B
& Hierarchical
& $1047$/seed
& $22.6{\pm}0.4\%$
& $6.1{\pm}0.1$ \\

& Gemma-4-26B
& Search
& $55$/seed
& $0.8{\pm}0.8\%$
& $8.3{\pm}0.1$ \\
\hline

\multirow{7}{*}{SWE-bench}
& DeepSeek-V4
& Sequential
& $198$/seed
& $23.0{\pm}1.3\%$
& $4.6{\pm}0.1$ \\

& DeepSeek-V4
& Predefined
& $229$/seed
& $29.6{\pm}2.8\%$
& $4.1{\pm}0.1$ \\

& DeepSeek-V4
& Hierarchical$^{\ddagger}$
& $13$/seed
& $17.3{\pm}5.8\%$
& $11.0{\pm}0.2$ \\

& DeepSeek-V4
& Search$^{\ddagger}$
& $1$/seed
& $0.0{\pm}0.0\%$
& $13.0{\pm}3.0$ \\

& Qwen3.6-35B
& Sequential
& $201$/seed
& $21.4{\pm}1.5\%$
& $4.7{\pm}0.3$ \\

& Qwen3.6-35B
& Predefined
& $186$/seed
& $28.2{\pm}0.3\%$
& $4.3{\pm}0.0$ \\

& Qwen3.6-35B
& Hierarchical
& $30$/seed
& $14.5{\pm}1.1\%$
& $8.6{\pm}0.1$ \\
\hline
\end{tabular}
\end{table*}

\section{Additional Plan Adherence Results}
\label{app:plan-adherence}

\paragraph{Plan completion is conditioned on execution budget and plan size.}

Plan adherence measures how much of the still-needed plan is completed within the available interaction budget. Full adherence results for Mind2Web and SWE-bench are reported in Tables~\ref{tab:plan-adherence-mind2web} and~\ref{tab:plan-adherence-swebench}, respectively.

Mind2Web additionally provides a more dependency-sensitive setting: browser tasks often require a sequence of prerequisite interactions before the final goal becomes reachable. Consequently, skipping a necessary intermediate plan unit can prevent subsequent actions from being executed successfully. This makes plan adherence particularly informative for web navigation, while still
distinguishing necessary execution units from redundant or obsolete steps in the generated plan.

\begin{table}[!ht]
\centering
\scriptsize
\caption{Mind2Web: 
Plan adherence under pattern-specific execution. \emph{Planned/task} and \emph{Completed/task} report the mean number of needed plan units generated and completed per task, respectively. Adherence measures the fraction of needed plan units completed within the execution budget, while \emph{Full plan} reports the fraction of trajectories completing all needed units. Values are mean $\pm$ standard deviation across seeds 7, 13, and 42. Search operates over competing candidate strategies and its completion values
are therefore not directly comparable to the ordered planning patterns.
}
\label{tab:plan-adherence-mind2web}
\resizebox{\columnwidth}{!}{%
\begin{tabular}{|l|l|c|c|c|c|}
\hline
\textbf{Model} &
\textbf{Planning mode} &
\textbf{Planned/task} &
\textbf{Completed/task} &
\textbf{Adherence} &
\textbf{Full plan} \\
\hline

\multirow{4}{*}{DeepSeek-V4}
& Predefined & $4.597{\pm}0.023$ & $2.590{\pm}0.000$
& $0.558{\pm}0.002$ & $9.2{\pm}0.1\%$ \\

& Sequential & $4.443{\pm}0.012$ & $3.543{\pm}0.012$
& $0.800{\pm}0.003$ & $40.0{\pm}0.7\%$ \\

& Hierarchical & $4.800{\pm}0.010$ & $2.637{\pm}0.006$ & $0.542{\pm}0.002$ & $1.2{\pm}0.4\%$ \\

& Search & $3.467{\pm}0.015$ & $3.433{\pm}0.015$ & $0.989{\pm}0.001$
& $96.5{\pm}0.2\%$ \\
\hline

\multirow{4}{*}{Qwen3.6-35B}
& Predefined & $4.987{\pm}0.006$ & $2.597{\pm}0.015$ & $0.521{\pm}0.003$ & $6.8{\pm}0.9\%$ \\

& Sequential & $4.540{\pm}0.050$ & $3.603{\pm}0.021$ & $0.797{\pm}0.006$ & $42.5{\pm}2.0\%$ \\

& Hierarchical & $5.233{\pm}0.006$ & $2.653{\pm}0.006$ & $0.504{\pm}0.002$ & $1.9{\pm}0.3\%$ \\

& Search & $3.353{\pm}0.021$ & $3.070{\pm}0.046$ & $0.915{\pm}0.011$
& $75.0{\pm}2.7\%$ \\
\hline
\multirow{4}{*}{Gemma-4-26B}
& Predefined & $3.813{\pm}0.005$ & $1.223{\pm}0.034$ & $0.334{\pm}0.008$ & $3.1{\pm}0.4\%$ \\
& Sequential$^{\S}$ & $3.725{\pm}0.015$ & $3.565{\pm}0.015$ & $0.956{\pm}0.001$ & $87.1{\pm}0.0\%$ \\
& Hierarchical & $7.395{\pm}0.085$ & $1.950{\pm}0.010$ & $0.311{\pm}0.006$ & $0.2{\pm}0.2\%$ \\
& Search & $3.000{\pm}0.000$ & $2.380{\pm}0.110$ & $0.793{\pm}0.036$ & $60.6{\pm}4.1\%$ \\
\hline
\end{tabular}}
\end{table}

\begin{table}[!ht]
\centering
\scriptsize
\begin{tabular}{|l|l|c|c|c|c|}
\hline
\textbf{Model} &
\textbf{Planning mode} &
\textbf{Planned/task} &
\textbf{Completed/task} &
\textbf{Adherence} &
\textbf{Full plan} \\
\hline
\multirow{4}{*}{DeepSeek-V4}
& Predefined & $4.790$ & $1.410$ & $0.297$ & $1.0\%$ \\
& Sequential$^{\S}$ & $3.770$ & $3.770$ & $1.000$ & $100.0\%$ \\
& Hierarchical & $13.650$ & $2.040$ & $0.163$ & $0.5\%$ \\
& Search & $3.260$ & $3.020$ & $0.927$ & $83.0\%$ \\
\hline
\multirow{4}{*}{Qwen3.6-35B}
& Predefined & $5.490$ & $1.700$ & $0.311$ & $1.0\%$ \\
& Sequential$^{\S}$ & $3.750$ & $3.750$ & $1.000$ & $100.0\%$ \\
& Hierarchical & $9.370$ & $2.390$ & $0.271$ & $0.2\%$ \\
& Search & $3.770$ & $2.920$ & $0.777$ & $49.8\%$ \\
\hline
\end{tabular}
\caption{SWE-bench Verified: plan adherence under pattern-specific execution. Columns as in the Mind2Web table. Both models ran at seed 13 only, so no row carries a spread. \textsc{Hierarchical} declares by far the most units ($13.7$ and $9.4$ per task against $3.3$--$5.5$ for the other patterns) and completes the smallest fraction of them. $^{\S}$ marks the same \textsc{Sequential} caveat as above.}
\label{tab:plan-adherence-swebench}
\end{table}

\section{Pattern-Ceiling Analysis}
\label{app:ceiling-analysis}
 
Comparing execution conditions on task success alone conflates two distinct questions: whether a pattern-specific executor can solve a task at all, and whether the declaration module selects the pattern that solves it. We therefore measure the two separately. For every benchmark--model pair we first build a \emph{planning pattern-ceiling matrix} by running each of the four planning patterns on every task under forced dispatch, i.e. bypassing the declaration module and dispatching pattern $p$ regardless of what the model would have declared.
 
\paragraph{Definitions.}
Let $\mathcal{T} = \{1,\dots,N\}$ be the tasks of a benchmark and
$\mathcal{P} = \{\textsc{predefined}, \textsc{sequential}, \textsc{hierarchical}, \textsc{search}\}$ the planning patterns. The ceiling matrix $M \in \{0,1\}^{N \times |\mathcal{P}|}$ has entries \[ m_{i,p} = \mathbb{1}\left[\text{forced dispatch of pattern } p
\text{ solves task } i\right],\]
scored by the benchmark's own evaluation protocol. From $M$ we derive four quantities:
 
\begin{itemize}[leftmargin=*]
    \item \textbf{Per-planning-pattern success rate} (column mean), $S(p) = \frac{1}{N}\sum_i m_{i,p}$. This is the task success rate of a fixed policy that always executes $p$.
 
    \item \textbf{Single-best-pattern policy},
    $S^\star = \max_{p} S(p)$, attained by $p^\star$. This is the strongest policy available \emph{without} per-task selection, and therefore the baseline any router must outperform to demonstrate that selection is doing work.
 
    \item \textbf{Oracle ceiling},
    $\mathrm{DSR}_{\max} = \frac{1}{N}\sum_i \max_{p} m_{i,p}$: the success rate of a policy that selects the correct pattern for every task. Tasks with $\max_p m_{i,p} = 0$ are solved by no pattern; we call these \textsc{none} tasks, and no declaration policy can affect them.
 
    \item \textbf{Routing headroom},
    $H = \mathrm{DSR}_{\max} - S^\star \geq 0$. This is the entire budget available to per-task selection. $H$ is strictly positive only when pattern wins are complementary, when some task solved by a weaker pattern is \emph{not} solved by $p^\star$. If patterns succeed on nested subsets of tasks, $H = 0$ and no selection policy, however good, can improve on always executing $p^\star$.
\end{itemize}
 
\paragraph{Evaluating a declaration policy.}
A declaration policy $\pi: \mathcal{T} \to \mathcal{P}$ achieves
$\mathrm{TSR}(\pi) = \frac{1}{N}\sum_i m_{i,\pi(i)}$. Because
$\mathrm{TSR}(\pi)$ is a lookup into $M$, it can be evaluated without re-running any executor: the declaration is the only model call per task. This \emph{free-lookup} protocol lets us compare declaration variants (zero-shot, few-shot, thinking on/off) under a fixed execution substrate, so differences are attributable to selection rather than to execution stochasticity.
 
$\mathrm{TSR}(\pi)$ alone, however, does not establish that $\pi$ selects \emph{per task}. A policy that emits a healthy spread of patterns but chooses them independently of the task achieves, in expectation,
\[
\mathrm{TSR}_{\text{null}}(\pi) \;=\; \frac{1}{N}\sum_{p} n_p\, S(p),
\qquad n_p = |\{i : \pi(i) = p\}|,
\]
the \emph{distribution-matched null}: the same marginal pattern distribution as $\pi$, but zero task-conditional information. We therefore report the excess $\Delta(\pi) = \mathrm{TSR}(\pi) - \mathrm{TSR}_{\text{null}}(\pi)$, which isolates per-task discrimination from any global prior over patterns that the prompt may have induced. $\Delta(\pi) \approx 0$ with a broad
distribution and $\Delta(\pi) \approx 0$ with a degenerate one are both selection failures, and $\mathrm{TSR}$ distinguishes neither. We additionally report performance restricted to the solvable subset $\{i : \max_p m_{i,p} = 1\}$, since \textsc{none} tasks contribute zero under every policy and dilute all of these quantities toward zero.
 
\paragraph{Separating selection failure from dispatch failure.}
The ceiling matrix and the trace verifier isolate different failure modes. A \emph{dispatch} failure means the executed pattern differs from the declared one; the verifier detects this directly, and it is the failure mode Planning-as-Routing is designed to eliminate. A \emph{selection} failure means dispatch was faithful but the declared pattern was the wrong one for the task; it appears as $\mathrm{TSR}(\pi) < S^\star$ with the verifier reporting full conformance. This distinction is what allows us to attribute the residual gap in Section~\ref{sec:rq4} to the declaration module rather than to the router or the executors.
 
\paragraph{Stochasticity and interpretation of the oracle ceiling.}
Each forced-pattern execution is stochastic. We therefore compute $S(p)$, $S^\star$, $\mathrm{DSR}_{\max}$, and $H$ independently for each random seed and report the mean and standard deviation across seeds. Thus, $\mathrm{DSR}_{\max}$ represents the expected empirical per-task ceiling when each planning pattern receives one independently sampled execution.

Because $\mathrm{DSR}_{\max}$ takes the maximum across multiple stochastic pattern executions, part of its advantage over the strongest fixed pattern could arise from repeated opportunities for success rather than from
planning-pattern complementarity alone. We therefore separately compare the oracle and ranked-fallback results against pass@$k$ controls obtained by repeating the strongest fixed planning pattern under matched numbers of
attempts. This separates gains due to trying different planning patterns from gains obtainable by simply retrying the same strong pattern.

\begin{figure*}[!ht]
    \centering
    \includegraphics[width=0.5\textwidth]
    {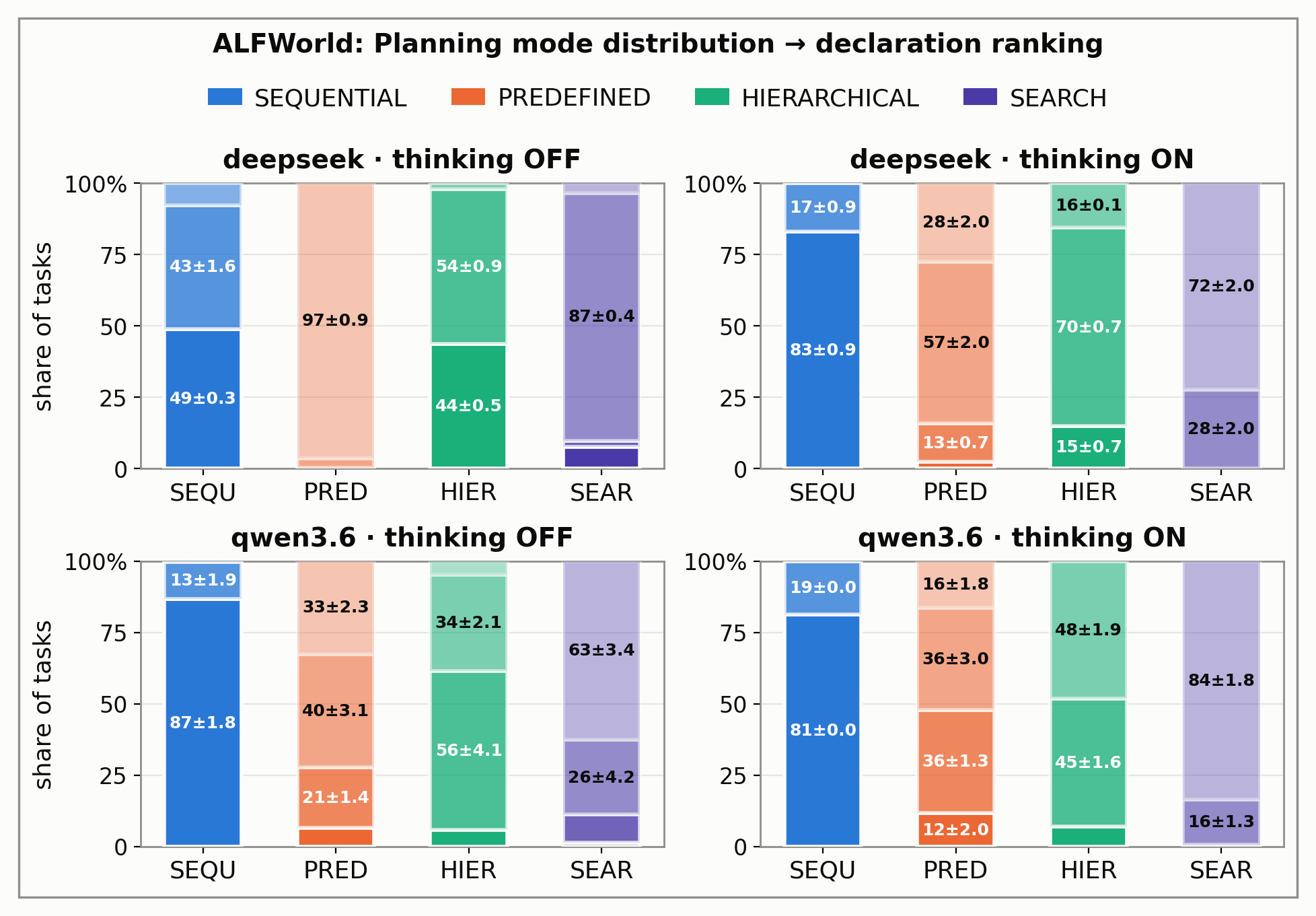}
    \hfill
    \includegraphics[width=0.5\textwidth]
    {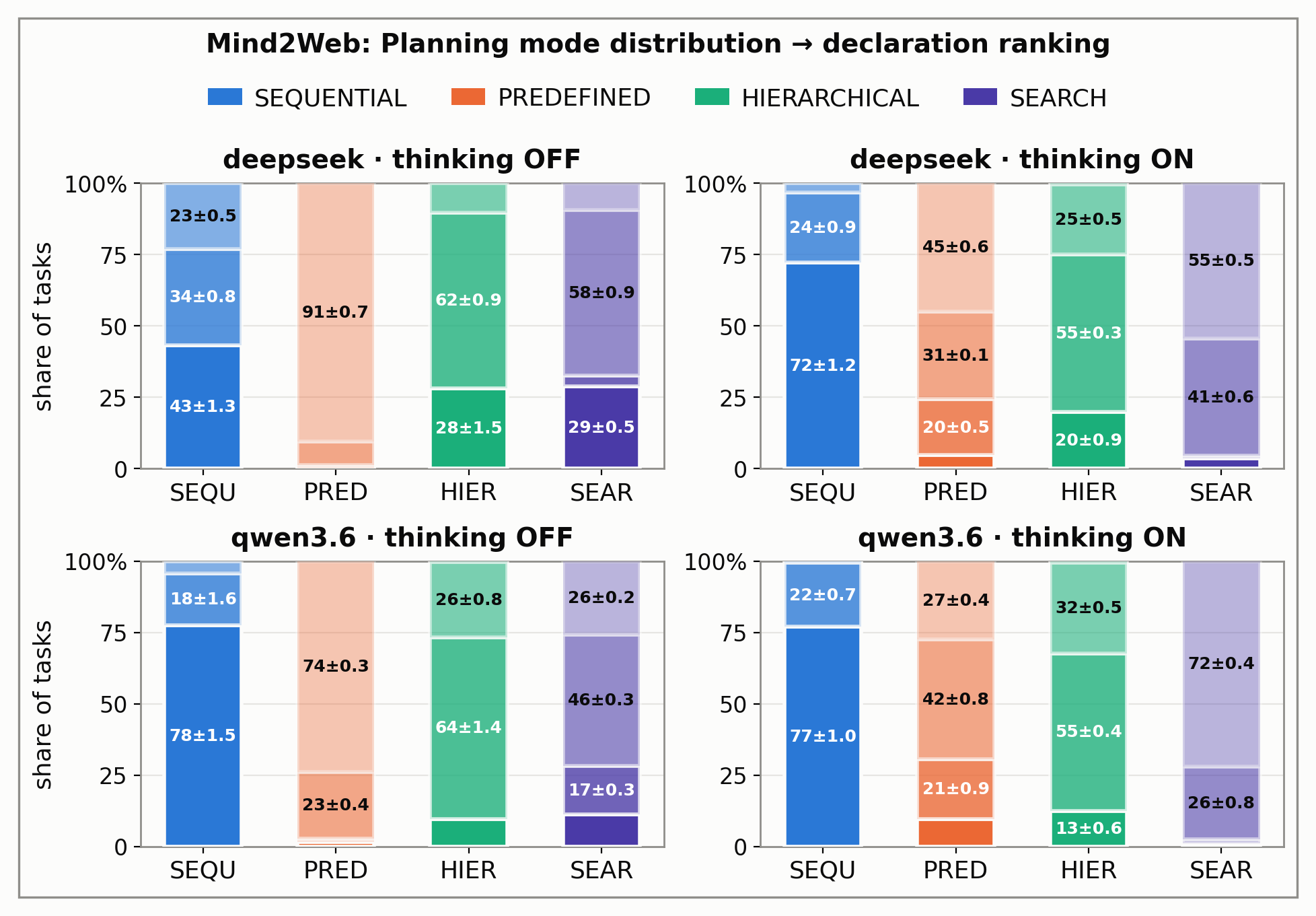}
    \hfill
    \includegraphics[width=0.5\textwidth]
    {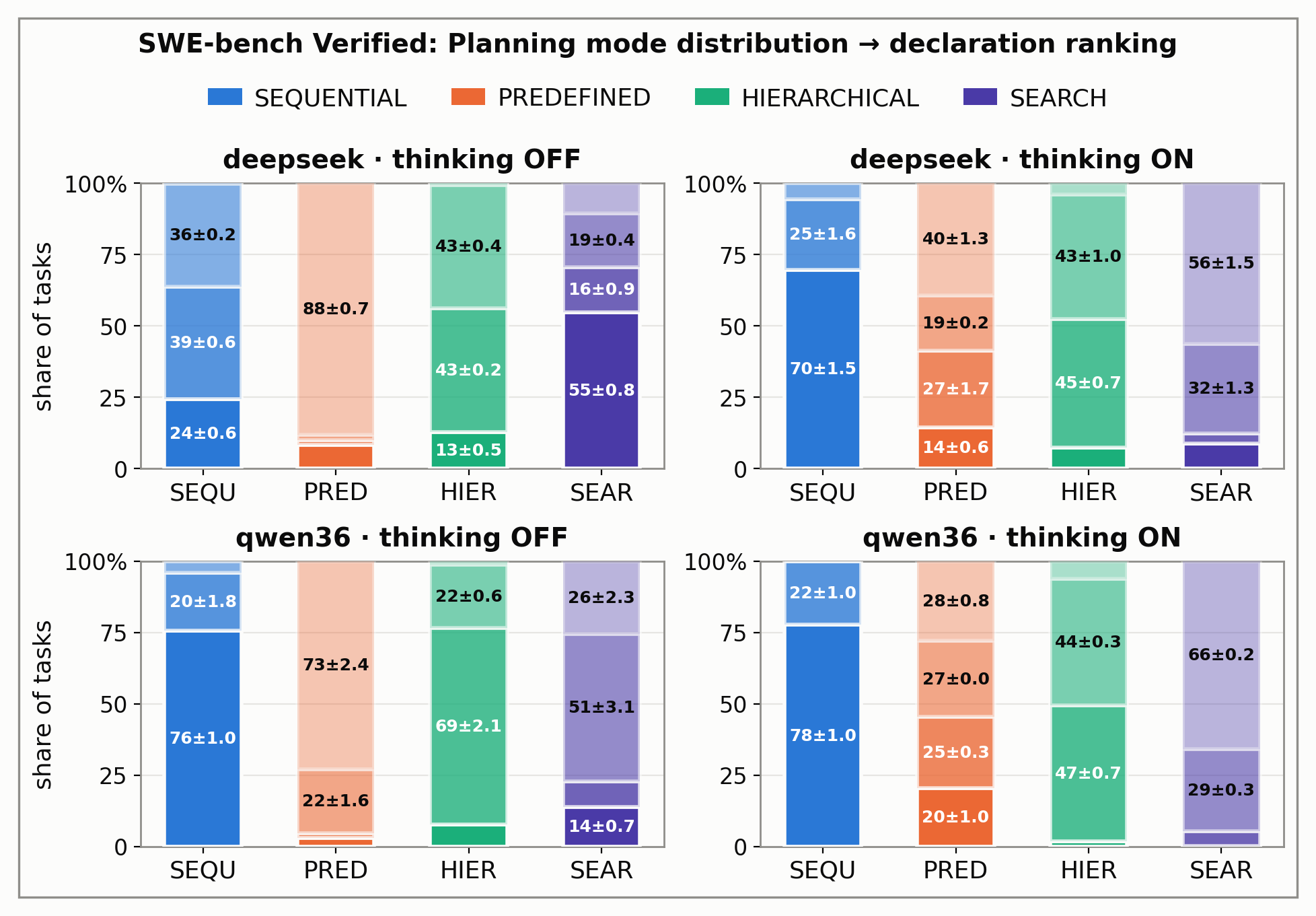}
    \hfill
    \includegraphics[width=0.5\textwidth]
    {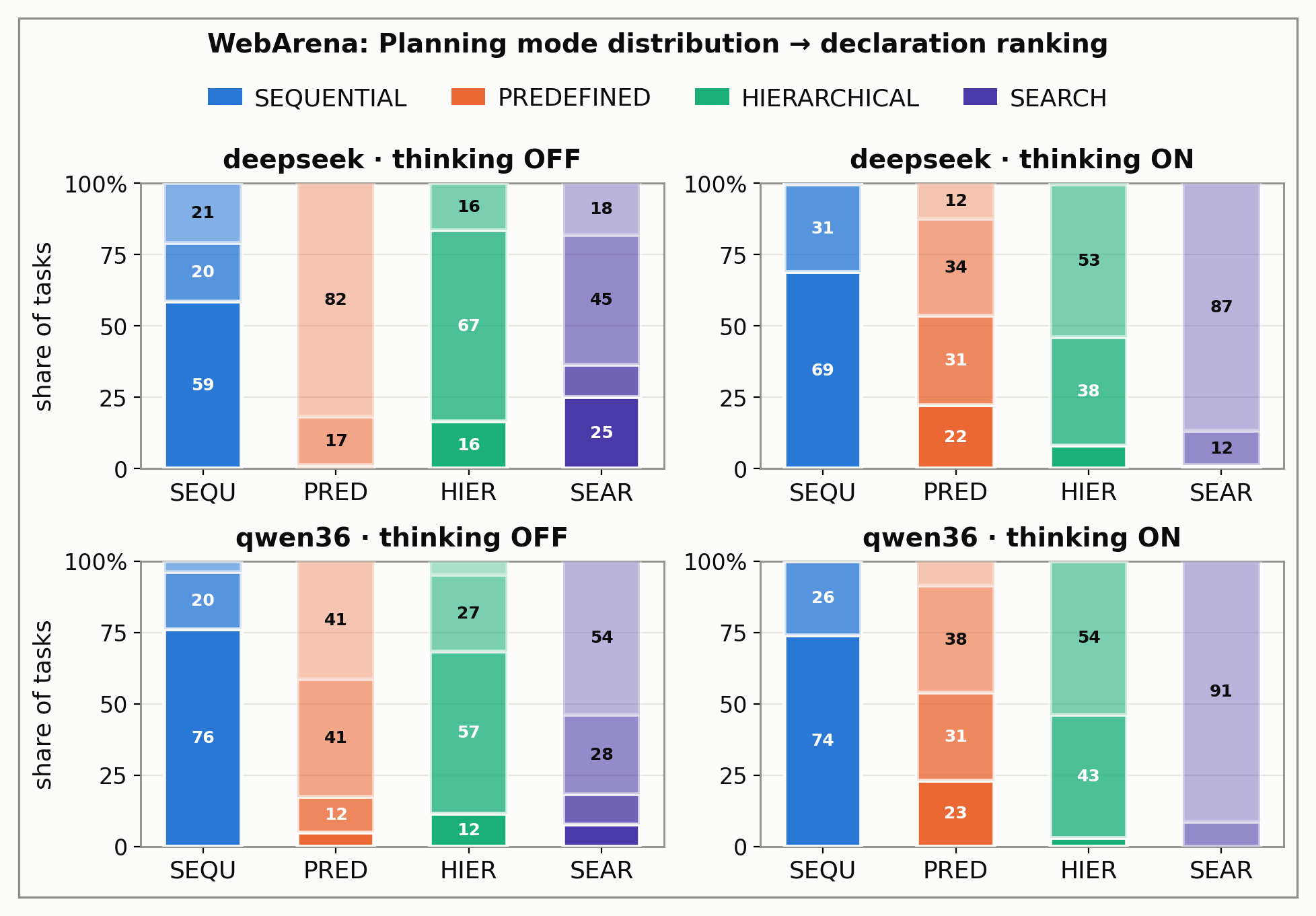}
    \caption{
    Distribution of declaration ranks assigned to each planning mode on ALFWolrd, Mind2Web, SWE-Bench Verified and WebArena (right), averaged across three seeds. Each stacked bar shows the fraction of tasks for which a planning mode is assigned Rank 1--4. Results are shown for DeepSeek-V4 and Qwen3.6 with thinking enabled and disabled. The strongly non-uniform distributions reveal systematic mode preferences in the declaration policy.
    }
    \label{fig:rank-distribution-swebench-webarena}
\end{figure*}

\begin{table}[!ht]
\centering
\scriptsize
\setlength{\tabcolsep}{3.5pt}
\caption{
Repeated-rollout ReAct control for Search. $3\times$Flat ReAct executes three independent trajectories from a reset environment and applies the same rubric-based judge used by Search to select one trajectory. Search dynamically chooses between 0 and 4 candidate plans per task, with four as the configured maximum. The realized candidate count averages $3.35\pm0.49$ across Search episodes (Appendix~\ref{app:inference-config}, Table~\ref{tab:search-candidate-counts}). Results in this table report mean $\pm$ standard deviation across seeds 7, 13, and 42. This control tests how much of Search's performance can be reproduced by repeated generic ReAct execution and trajectory selection.
}
\label{tab:multirollout-react}
\begin{tabular}{|l|l|c|c|c|c|c|}
\hline
\textbf{Benchmark} &
\textbf{Model} &
\textbf{Metric} &
\textbf{Flat ReAct} &
\textbf{$3\times$Flat ReAct + Judge} &
\textbf{Flat ReAct Pass@3} &
\textbf{Search} \\
\hline

\multirow{2}{*}{ALFWorld}
& DeepSeek-V4 & TSR
& $0.440{\pm}0.012$
& $0.526 {\pm} 0.011$
& $0.600 {\pm} 0.015$
& $\mathbf{0.918{\pm}0.026}$ \\

& Qwen3.6-35B & TSR
& $0.535{\pm}0.007$
& $0.664 {\pm} 0.015$ & $0.758 {\pm} 0.004$
& $\mathbf{0.796{\pm}0.004}$ \\
\hline

\multirow{4}{*}{Mind2Web}
& \multirow{2}{*}{DeepSeek-V4}
& TSR & $\mathbf{0.058{\pm}0.003}$& $\mathbf{0.059 {\pm} 0.004}$ & $0.077 {\pm} 0.002$&$\mathbf{0.057{\pm}0.006}$ \\
&
& SSR & $\mathbf{0.434{\pm}0.001}$ & $0.448 {\pm} 0.005$ & $0.504 {\pm}0.001$ & $\mathbf{0.449{\pm}0.000}$ \\
\cline{2-7}

& \multirow{2}{*}{Qwen3.6-35B}
& TSR & $0.042{\pm}0.002$ & $0.044 {\pm} 0.000$ & $0.069 {\pm} 0.001$ & $\mathbf{0.052{\pm}0.001}$ \\
&
& SSR & $0.366{\pm}0.003$ & $0.385 {\pm} 0.002$ &$0.466 {\pm} 0.002$ & $\mathbf{0.411{\pm}0.003}$ \\
\hline

\multirow{2}{*}{SWE-bench}
& DeepSeek-V4
& PSR & $0.384{\pm}0.010$ & $\mathbf{0.410 {\pm} 0.007}$ & $0.513 {\pm} 0.003$ &$0.400{\pm}0.018$ \\

& Qwen3.6-35B
& PSR & $0.179{\pm}0.034$ & $0.304 {\pm} 0.023$ & $0.381 {\pm} 0.019$& $0.296{\pm}0.040$ \\
\hline

\multirow{2}{*}{WebArena}
& DeepSeek-V4
& TSR & $0.455 {\pm} 0.000$ & $0.452 {\pm} 0.031$ &$0.594 {\pm} 0.004$ & $\mathbf{0.580 {\pm} 0.006}$ \\

& Qwen3.6-35B
& TSR & $0.332 {\pm} 0.009$ & $0.386 {\pm} 0.017$ & $0.548 {\pm} 0.037$& $\mathbf{0.577 {\pm} 0.026}$  \\
\hline
\end{tabular}
\end{table}

\subsection{Repeated-rollout ReAct control for Search.}
\label{app:repeated-rollout-react}
Search differs from the other planning patterns because it can execute multiple candidate trajectories before selecting one with a rubric-based judge. The Search planner dynamically selects between 0 and 4 candidate plans per task, rather than always executing the maximum of four. Across all Search runs, it realizes $3.35\pm0.49$ candidates per task on average (See Table~\ref{tab:search-candidate-counts}). Thus, the Flat ReAct control of three rolls provides a rollout count of approximately matched overall. We therefore test how much of its advantage can be reproduced by repeated generic ReAct execution. For each task, we run Flat ReAct three times independently from a reset environment and apply the same rubric-based judge used by Search to select among the resulting trajectories. We repeat this control across three different seeds.

Table~\ref{tab:multirollout-react} shows that repeated execution with the same rubric-based judge generally improves Flat ReAct, but does not fully account for the gains of Search. Across ALFWorld and WebArena, Search remains
substantially stronger than $3\times$Flat ReAct + Judge, with improvements
ranging from $0.132$ to $0.392$ on ALFWorld and from $0.128$ to $0.191$ on WebArena. On Mind2Web, the differences are smaller and more model dependent: Search is slightly below repeated ReAct for DeepSeek-V4 in TSR ($0.057$ versus $0.059$), but remains higher in SSR and for both metrics with Qwen3.6-35B. On SWE-bench, repeated ReAct performs slightly better than Search for both models ($0.410$ versus $0.400$ for DeepSeek-V4 and $0.304$ versus $0.296$ for Qwen3.6-35B). This is consistent with the forced-pattern results, where Hierarchical rather than Search is the strongest planning pattern on SWE-bench. Overall, repeated execution explains part of the gain from
additional rollout budget, but Search retains an advantage in most benchmark--model settings.

To distinguish candidate quality from trajectory-selection quality, we measured Flat ReAct pass@3, as shown in Table~\ref{tab:multirollout-react}. For each task, pass@3 counts the task as successful if any of the three independent Flat ReAct trajectories succeeds, thereby providing hindsight selection over exactly the same rollout set used by $3\times$Flat ReAct + Judge. The difference between pass@3 and the judge-selected result therefore measures how much attainable success is lost by trajectory selection, while comparing Search against Flat ReAct pass@3 tests whether Search generates stronger candidate trajectories than repeated generic ReAct alone.

\subsection{Repeated-fixed-pattern control.}
\label{app:repeated_fix_pattern}

For a fixed planning pattern $p$, we define
\begin{equation}
\mathrm{pass@}k(p)
=
\frac{1}{N}
\sum_{i=1}^{N}
\mathbf{1}
\left[
\max_{r\in\{1,\ldots,k\}}
m_{i,p}^{(r)}=1
\right],
\end{equation}
where $m_{i,p}^{(r)}$ denotes the outcome of the $r$th independent
execution of pattern $p$ on task $i$. We use $p^\star$, the strongest
fixed planning pattern, as the primary retry baseline. We compare
TSR@2 and TSR@3 against pass@2($p^\star$) and pass@3($p^\star$),
respectively, and compare the four-pattern forced oracle against
pass@4($p^\star$).

\begin{table*}[!ht]
\centering
\scriptsize
\setlength{\tabcolsep}{3.5pt}
\renewcommand{\arraystretch}{1.10}
\caption{
Retry control for the forced-pattern ceiling.
For each benchmark--model pair, $p^\star$ denotes the strongest fixed
planning pattern under the primary task-level metric, and
$S(p^\star)$ denotes its corresponding score.
$\mathrm{DSR}_{\max}$ denotes the per-task forced-pattern oracle over
the evaluated planning patterns, and
$H=\mathrm{DSR}_{\max}-S(p^\star)$ is the original oracle gap.
pass@$k$ reports performance when the same fixed pattern $p^\star$ is
executed independently $k$ times, with a task counted as successful if
any of the $k$ executions succeeds.
$H_{\mathrm{retry}}@3 =
\mathrm{DSR}_{\max}-\mathrm{pass@3}(p^\star)$ denotes the residual
oracle gap after three executions of the same fixed pattern.
Negative values indicate that repeated execution of the fixed pattern
exceeds the forced-pattern oracle.
For Mind2Web, we report both task success rate (TSR) and step success
rate (SSR). For Gemma-4-26B, $p^\star$ is Predefined because it is the
strongest fixed pattern under TSR; the corresponding SSR values therefore use the same Predefined pattern rather than the SSR-maximizing Search pattern.
}
\label{tab:retry-null}
\resizebox{\textwidth}{!}{%
\begin{tabular}{|l|l|c|c|c|c|c|c|c|c|}
\hline
\textbf{Benchmark} &
\textbf{Model} &
\textbf{Metric} &
\textbf{$n$} &
\textbf{$S(p^\star)$ (Pattern)} &
\textbf{$\mathrm{DSR}_{\max}$} &
\textbf{$H$} &
\textbf{pass@2} &
\textbf{pass@3} &
\textbf{$H_{\mathrm{retry}}@3$} \\
\hline

\multirow{3}{*}{ALFWorld}
& DeepSeek-V4
& TSR
& 134
& $0.918$ (SEARCH)
& $0.953$
& $+0.035$
& $0.963$
& $\mathbf{0.978}$
& $\mathbf{-0.025}$ \\

& Qwen3.6-35B
& TSR
& 134
& $0.796$ (SEARCH)
& $0.925$
& $+0.129$
& $0.896$
& $\mathbf{0.940}$
& $\mathbf{-0.015}$ \\

& Gemma-4-26B
& TSR
& 134
& $0.600$ (HIER)
& $0.841$
& $+0.241$
& $0.754$
& $0.813$
& $\mathbf{+0.027}$ \\
\hline

\multirow{6}{*}{Mind2Web}

& \multirow{2}{*}{DeepSeek-V4}
& TSR
& 1341
& $0.057$ (SEARCH)
& $0.097$
& $+0.040$
& $0.073$
& $0.083$
& $\mathbf{+0.014}$ \\

&
& SSR
& 1341
& $0.449$ (SEARCH)
& $0.550$
& $+0.101$
& $0.500$
& $0.526$
& $\mathbf{+0.024}$ \\
\cline{2-10}

& \multirow{2}{*}{Qwen3.6-35B}
& TSR
& 1341
& $0.053$ (SEARCH)
& $0.084$
& $+0.031$
& $0.071$
& $0.083$
& $\mathbf{+0.001}$ \\

&
& SSR
& 1341
& $0.411$ (SEARCH)
& $0.505$
& $+0.094$
& $0.470$
& $0.501$
& $\mathbf{+0.005}$ \\
\cline{2-10}

& \multirow{2}{*}{Gemma-4-26B}
& TSR
& 1341
& $0.147$ (PRED)
& $0.283$
& $+0.135$
& $0.237$
& $0.292$
& $\mathbf{-0.010}$ \\

&
& SSR
& 1341
& $0.347$ (PRED)$^\dagger$
& $0.585$
& $+0.238$
& $0.461$
& $0.524$
& $\mathbf{+0.062}$ \\
\hline

\multirow{2}{*}{SWE-bench}
& DeepSeek-V4
& PSR
& 500
& $0.442$ (HIER)
& $0.552$
& $+0.110$
& $0.503$
& $0.538$
& $\mathbf{+0.014}$ \\

& Qwen3.6-35B
& PSR
& 500
& $0.330$ (HIER)
& $0.445$
& $+0.115$
& $0.420$
& $0.449$
& $\mathbf{-0.004}$ \\
\hline

\multirow{2}{*}{WebArena}
& DeepSeek-V4
& TSR
& 204
& $0.580$ (SEARCH)
& $0.736$
& $+0.147$
& $0.695$
& $0.744$
& $-0.009$ \\

& Qwen3.6-35B
& TSR
& 204
& $0.578$ (SEARCH)
& $0.693$
& $+0.116$
& $0.665$
& $0.693$
& $-0.000$ \\
\hline

\end{tabular}%
}

\vspace{2pt}
\begin{minipage}{0.98\textwidth}
\scriptsize
$^\dagger$For Gemma-4-26B on Mind2Web, Predefined is the strongest
fixed pattern under TSR and is therefore used as the fixed-pattern retry
control for both TSR and SSR. Search has the highest single-run SSR
($0.388$), but the reported SSR pass@2 and pass@3 values correspond to
repeated execution of Predefined.
\end{minipage}

\end{table*}

\subsection{Task-Blind Declaration Null}
\label{app:declaration-null}

\begin{table*}[!ht]
\centering
\scriptsize
\setlength{\tabcolsep}{3.5pt}
\renewcommand{\arraystretch}{1.12}
\caption{
Task-blind declaration null. $\mathrm{TSR}(\pi)$ denotes task success under
the model's top-1 declared planning mode. $\mathrm{TSR}_{\mathrm{null}}$
preserves each model's overall declaration frequencies while randomly
reassigning declarations across tasks. We report
$\Delta(\pi)=\mathrm{TSR}(\pi)-\mathrm{TSR}_{\mathrm{null}}$, the
95\% interval under 5{,}000 permutations, one-sided permutation $p$-values
for $\Delta>0$, two-sided $p$-values, and Holm-corrected $p$-values.
}
\label{tab:declaration-null}
\resizebox{\textwidth}{!}{%
\begin{tabular}{|l|l|r|c|c|c|c|c|c|c|}
\hline
\textbf{Benchmark} &
\textbf{Model} &
\textbf{$n$} &
\textbf{$\mathrm{TSR}(\pi)$} &
\textbf{$\mathrm{TSR}_{\mathrm{null}}$} &
\textbf{$\Delta(\pi)$} &
\textbf{95\% null interval} &
\textbf{$p$ ($\Delta>0$)} &
\textbf{$p$ (two-sided)} &
\textbf{Holm} \\
\hline

\multirow{3}{*}{ALFWorld}
& DeepSeek-V4
& 134
& $0.721{\pm}0.029$
& $0.745{\pm}0.011$
& $\mathbf{-0.024{\pm}0.035}$
& $[-0.027,+0.026]$
& $0.972$
& $0.062$
& $1.000$ \\

& Qwen3.6-35B
& 134
& $0.667{\pm}0.049$
& $0.659{\pm}0.028$
& $+0.008{\pm}0.021$
& $[-0.017,+0.018]$
& $0.226$
& $0.406$
& $1.000$ \\

& Gemma-4-26B
& 134
& $0.560{\pm}0.016$
& $0.553{\pm}0.012$
& $+0.007{\pm}0.005$
& $[-0.013,+0.014]$
& $0.217$
& $0.364$
& $1.000$ \\
\hline

\multirow{3}{*}{Mind2Web}
& DeepSeek-V4
& 1341
& $0.053{\pm}0.003$
& $0.051{\pm}0.001$
& $+0.003{\pm}0.002$
& $[-0.004,+0.004]$
& $0.094$
& $0.180$
& $0.944$ \\

& Qwen3.6-35B
& 1341
& $0.036{\pm}0.001$
& $0.034{\pm}0.001$
& $+0.002{\pm}0.000$
& $[-0.003,+0.003]$
& $0.126$
& $0.251$
& $1.000$ \\

& Gemma-4-26B
& 1341
& $0.113{\pm}0.006$
& $0.113{\pm}0.007$
& $+0.001{\pm}0.000$
& $[-0.001,+0.002]$
& $0.707$
& $0.715$
& $1.000$ \\
\hline

\multirow{2}{*}{SWE-bench}
& DeepSeek-V4
& 500
& $0.415{\pm}0.016$
& $0.406{\pm}0.018$
& $+0.009{\pm}0.002$
& $[-0.014,+0.014]$
& $0.107$
& $0.213$
& $0.959$ \\

& Qwen3.6-35B
& 500
& $0.315{\pm}0.005$
& $0.317{\pm}0.005$
& $-0.001{\pm}0.000$
& $[-0.011,+0.011]$
& $0.598$
& $0.808$
& $1.000$ \\
\hline

\multirow{2}{*}{WebArena}
& DeepSeek-V4
& 204
& $0.460{\pm}0.011$
& $0.477{\pm}0.005$
& $-0.017{\pm}0.005$
& $[-0.029,+0.028]$
& $0.627$
& $0.710$
& $1.000$ \\

& Qwen3.6-35B
& 204
& $0.404{\pm}0.019$
& $0.429{\pm}0.005$
& $\mathbf{-0.025{\pm}0.009}$
& $[-0.030,+0.033]$
& $0.925$
& $0.215$
& $1.000$ \\
\hline

\end{tabular}%
}
\end{table*}

\section{Comparison with prior work.}
\label{app:comparison_with_piror_work}
We compare forced planning strategies, model-declared routing, and the empirical ceiling with representative prior work to assess whether they achieve comparable task success across benchmarks and models.
Because prior systems differ in backbone models, demonstrations, training, and inference procedures, we treat these results as benchmark-level context rather than controlled comparisons. On ALFWorld, AdaPlanner~\citep{sun2023adaplanner} reports $0.918$ TSR with GPT-3 and $0.806$ with GPT-3.5, using six expert demonstrations together with closed-loop plan refinement. On Mind2Web, TRAD~\citep{zhou2024trad} reports $0.021$ TSR and $0.280$ SSR, while the more recent ReasoningBank~\citep{ouyang2026reasoningbank} reaches $0.051$ TSR and $0.456$ SSR using experience memory accumulated from prior trajectories. On SWE-bench Verified, SWE-agent~\citep{yang2024swe} reports a $0.336$ resolved rate with Claude-3.5. Finally, on full WebArena, PLAN-AND-ACT~\citep{erdoganplan} reports $0.457$ TSR with Llama-70B and $0.482$ with QwQ-32B using a dedicated planner--executor architecture. These results show that our forced-pattern and routing-based evaluations operate in a performance regime comparable to representative prior systems, while our primary conclusions rely on the controlled Flat ReAct, Plan+ReAct, and Planning-as-Routing comparisons under the same models and execution protocol.

\FloatBarrier

\section{Additional Analysis}
\label{app:additiona_analysis}

\paragraph{Declarations provide no reliable task-specific advantage over a task-blind policy.}
For each model and benchmark, we compare the task success obtained from the
declared planning mode, $\mathrm{TSR}(\pi)$, with a task-blind null policy,
$\mathrm{TSR}_{\mathrm{null}}(\pi)$. The null preserves how frequently the
model declares each planning mode, but removes the association between the
declared mode and the individual task. We measure
\[
\Delta(\pi)=
\mathrm{TSR}(\pi)-\mathrm{TSR}_{\mathrm{null}}(\pi).
\]

Using the top-1 declarations underlying the selection results,
$\Delta(\pi)$ falls within the 95\% interval of a 5{,}000-permutation
null for every benchmark--model pair. No pair shows a significant advantage over the task-blind policy (minimum one-sided $p=0.094$; all Holm-corrected $p\geq0.944$). Across pairs, $\Delta(\pi)$ ranges from $-0.025$ to $+0.009$, with the largest observed deviations being negative (ALFWorld/DeepSeek-V4: $\Delta=-0.024$; WebArena/Qwen3.6-35B:
$\Delta=-0.025$). Thus, we find no reliable evidence that current
top-1 declarations match planning modes to individual tasks better than expected from each model's overall declaration frequencies.

\paragraph{Fallback through ranked planning-mode declarations improves success, but also adds execution attempts.}
We next evaluate the declaration module as a ranking over planning patterns rather than considering only its top-1 choice
(Table~\ref{tab:selection-improvement}). We define TSR@$k$ as a sequential fallback policy: the agent first executes the rank-1 declared mode and, if the task is not solved, proceeds to the rank-2 mode and then the rank-3 mode. Task success increases substantially with $k$. On ALFWorld, DeepSeek-V4 improves from $0.721$ at TSR@1 to $0.893$ at TSR@2 and $0.948$ at TSR@3, while Qwen3.6-35B increases from $0.667$ to $0.804$ and $0.873$. The same trend appears across the other benchmarks: DeepSeek-V4 increases from $0.415$ to $0.498$ and $0.541$ on SWE-bench, from $0.460$ to $0.639$ and $0.696$ on WebArena, and from $0.053$ to $0.075$ and $0.092$ on Mind2Web.

However, each additional rank also provides another opportunity to execute the task. We therefore compare TSR@$k$ with pass@$k(p^\star)$, which repeats the strongest fixed planning pattern for the same number of attempts (Table~\ref{tab:retry-null}). On ALFWorld, pass@3 reaches $0.978$ for DeepSeek-V4 and $0.940$ for Qwen3.6-35B, exceeding the corresponding TSR@3 values of $0.948$ and $0.873$. On SWE-bench, TSR@3 and pass@3 are nearly identical for DeepSeek-V4 ($0.541$ versus $0.538$), while pass@3 is slightly higher for Qwen3.6-35B ($0.449$ versus $0.438$). Thus, ranked fallback improves task success, but much of this gain can also be obtained
by repeatedly executing a strong fixed planning pattern.


\paragraph{Additional deliberation does not consistently improve planning-mode declaration.}
Figure~\ref{fig:rank-distribution-swebench-webarena} shows that planning-mode rankings are strongly non-uniform across models and benchmarks, indicating systematic preferences for particular modes. Enabling thinking changes these ranking distributions, but does not consistently improve the task success obtained from the top-1 declaration. For DeepSeek-V4, TSR@1 decreases from $0.721$ to $0.656$ on ALFWorld, $0.053$ to $0.051$ on Mind2Web, $0.415$ to $0.411$ on SWE-bench, and $0.460$ to $0.384$ on WebArena. Qwen3.6-35B shows only a small increase on ALFWorld (0.667 to 0.674), while decreasing on Mind2Web (0.036 to 0.034), SWE-bench
(0.313 to 0.310), and WebArena
(0.404 to 0.394). Thus, enabling thinking changes the model's planning-mode preferences, but does not consistently improve top-1 declaration performance.

\paragraph{Few-shot task--pattern demonstrations improve raw top-1 performance, but rarely task-specific discrimination.}
Since enabling thinking alone does not reliably improve planning-mode declaration, we next test whether explicit task--pattern demonstrations can better guide the declaration model. The few-shot demonstrations are constructed from held-out forced-pattern executions and are disjoint from the evaluation tasks. Table~\ref{tab:selection-improvement} (see main paper) shows that few-shot prompting improves top-1 declaration performance in most evaluated configurations. The largest gain occurs on ALFWorld for DeepSeek-V4
with thinking enabled, where TSR increases from $0.656$ to $0.812$
($+0.156$). Improvements are smaller on the other benchmarks: up to $+0.018$ on Mind2Web, $+0.027$ on SWE-bench, and $+0.054$ on WebArena. One configuration decreases slightly (DeepSeek-V4 on SWE-bench with thinking disabled, $-0.011$), while the two negative changes on DeepSeek-V4 Mind2Web TSR are negligible ($-0.003$ and $-0.002$), and two configurations remain unchanged. Overall, task--pattern demonstrations provide a more consistent benefit than enabling thinking alone, although the magnitude of the gain depends on the benchmark and model.

The smaller gains on Mind2Web suggest that improving planning-mode
declaration alone does not necessarily translate into large downstream
improvements. Mind2Web requires correct element grounding and operation selection across an entire multi-step trajectory, so errors during execution can still dominate even when the declared planning mode improves. Its diversity across websites and task types may also make a small set of task--pattern demonstrations less informative than in more regular environments.

\section{Inference Cost Analysis}
\label{app:inference-cost}

\begin{figure*}[!ht]
    \centering
    \includegraphics[width=\textwidth]
    {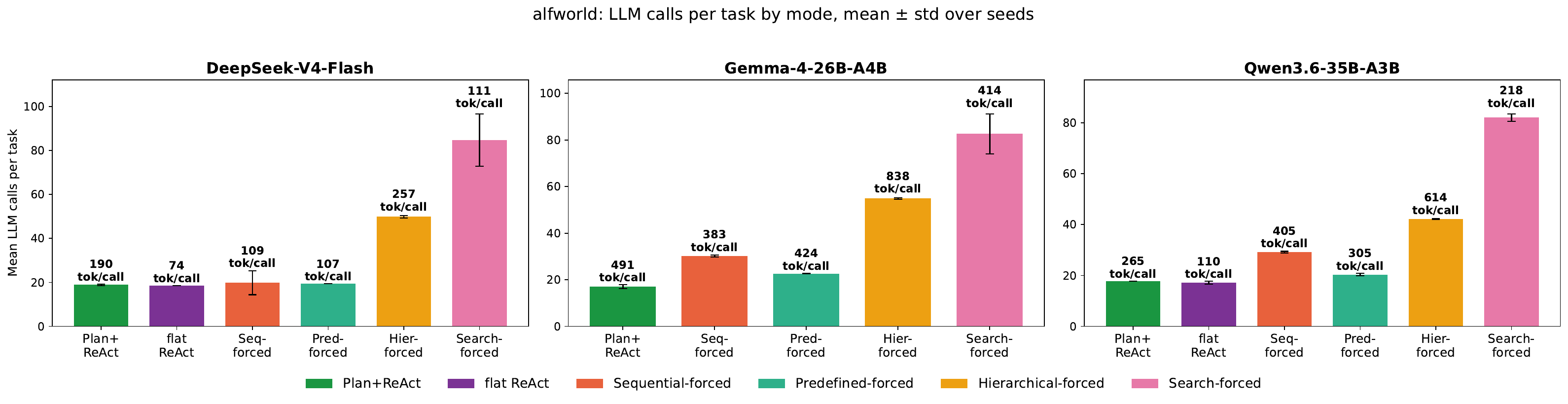}
    \hfill
    \includegraphics[width=\textwidth]
    {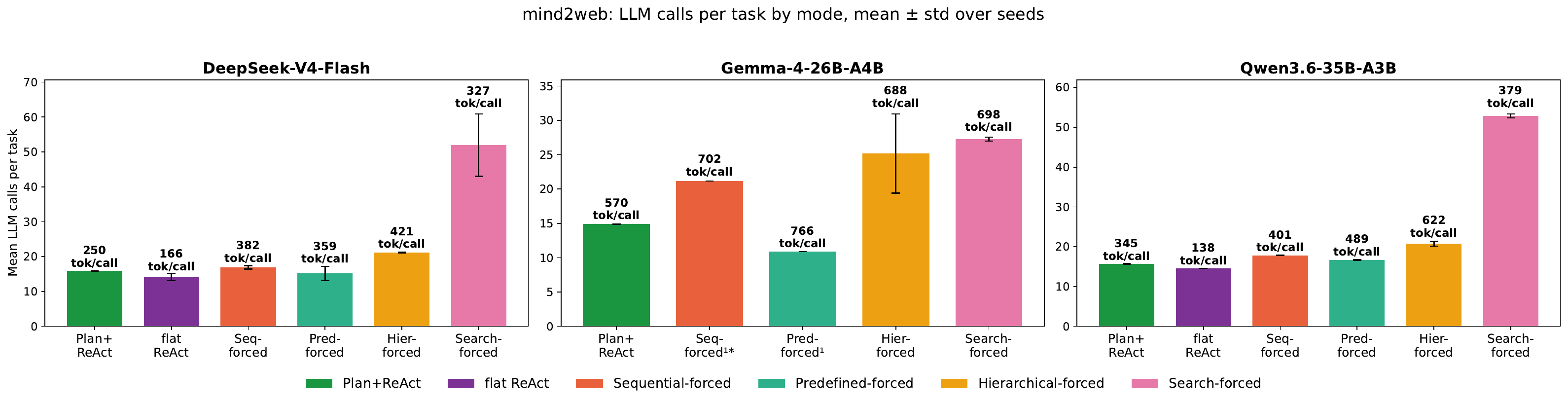}
    \hfill
    \includegraphics[width=\textwidth]
    {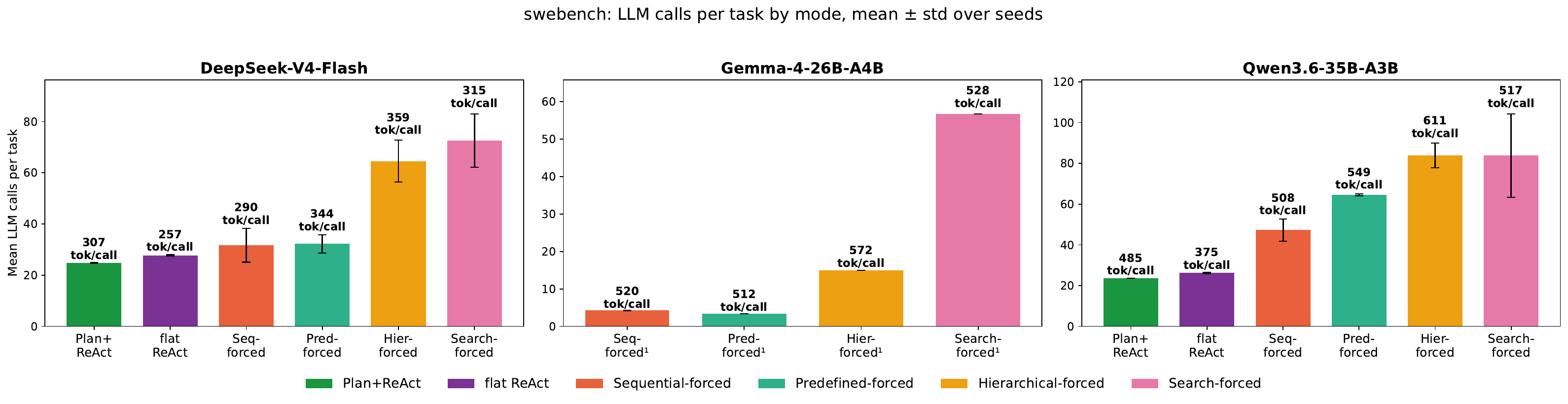}
    \hfill
    \includegraphics[width=\textwidth]
    {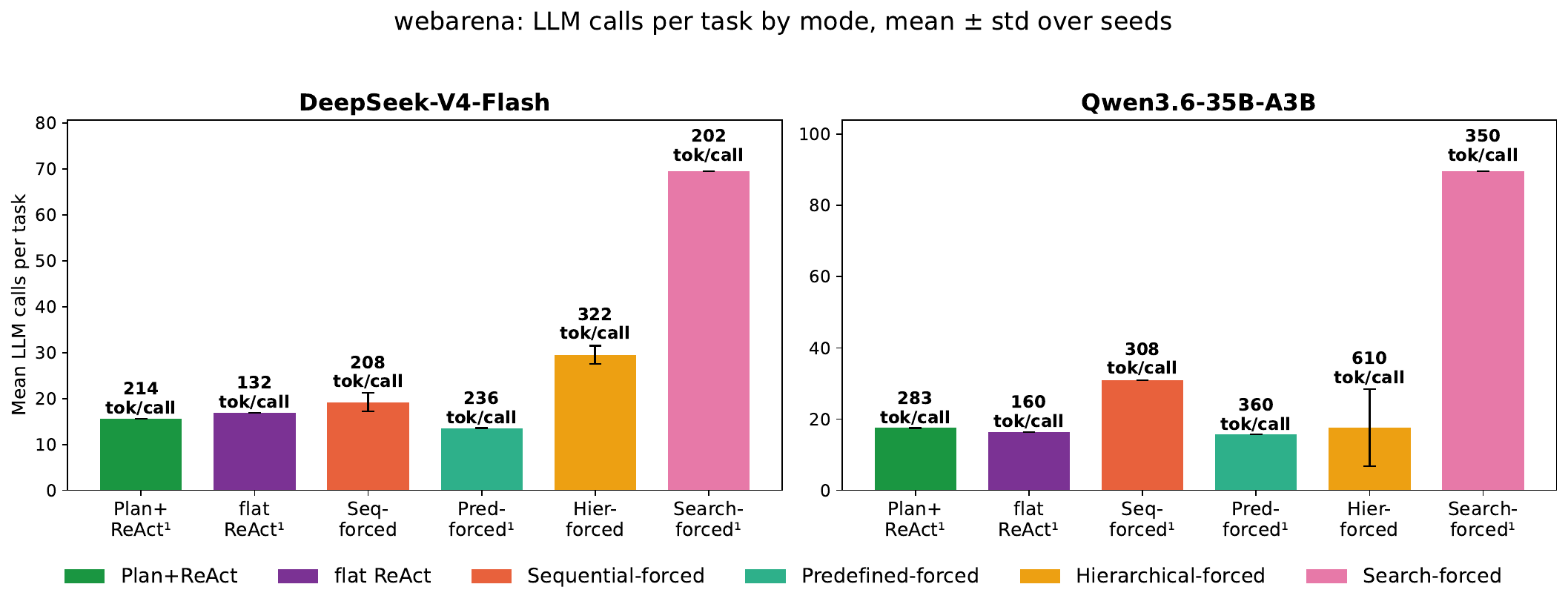}
    \caption{
    LLM calls per task across execution patterns on ALFWorld, Mind2Web, SWE-bench, and WebArena. Bars report the mean number of LLM invocations per task and error bars denote standard deviation across available seeds. Labels above the bars report the mean generated tokens per LLM call. Hierarchical and Search generally require more model invocations because of decomposition, coordination, and candidate exploration, while the average length of each call does not increase proportionally.
    }
    \label{fig:llm_calls_all_modes}
\end{figure*}

\paragraph{Inference-cost characteristics.}
We next examine whether the gains from pattern-specific execution can be explained simply by greater inference cost. We compare aggregate token usage, the fraction spent on thinking, and the number of LLM calls across execution patterns. As shown in Appendix~\ref{app:inference-cost},
Figs.~\ref{fig:thinking_content_tokens_distribution}
and~\ref{fig:llm_calls_all_modes}, both Hierarchical and Search generally use more aggregate tokens than Flat ReAct, Plan+ReAct, Sequential, and Predefined execution, largely because they invoke the LLM more frequently
for decomposition, coordination, or candidate exploration. However, these additional calls are not necessarily longer: Search often makes the largest
number of calls while using fewer tokens per call than Hierarchical execution. Moreover, neither total token usage nor the number of LLM calls tracks task
success monotonically. For instance, on ALFWorld, Search uses fewer aggregate tokens than Hierarchical for DeepSeek-V4 and Qwen3.6-35B while achieving higher task success. These results suggest that inference volume alone does not explain the observed performance differences. 
Detailed inference-cost statistics are reported in
Appendix~\ref{app:inference-cost}.

\begin{figure*}[!ht]
    \centering
    \includegraphics[width=\textwidth]
    {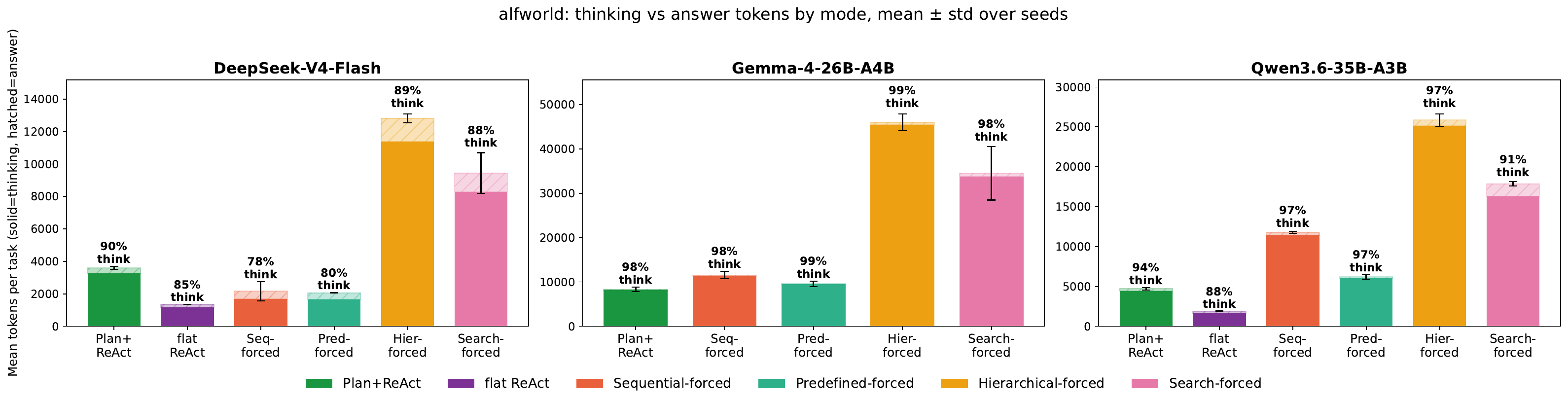}
    \hfill
    \includegraphics[width=\textwidth]
    {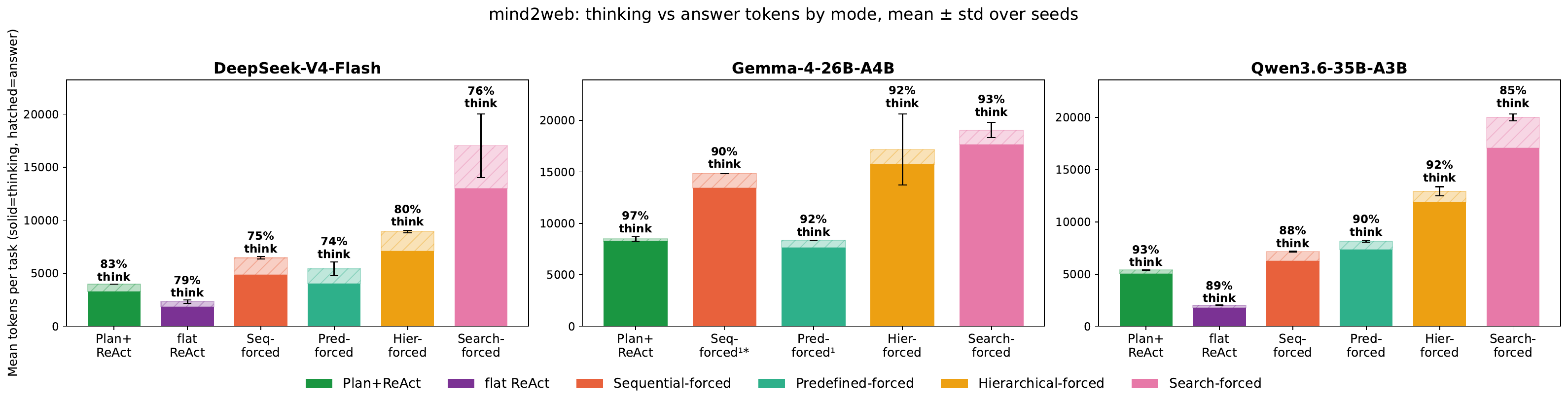}
    \hfill
    \includegraphics[width=\textwidth]
    {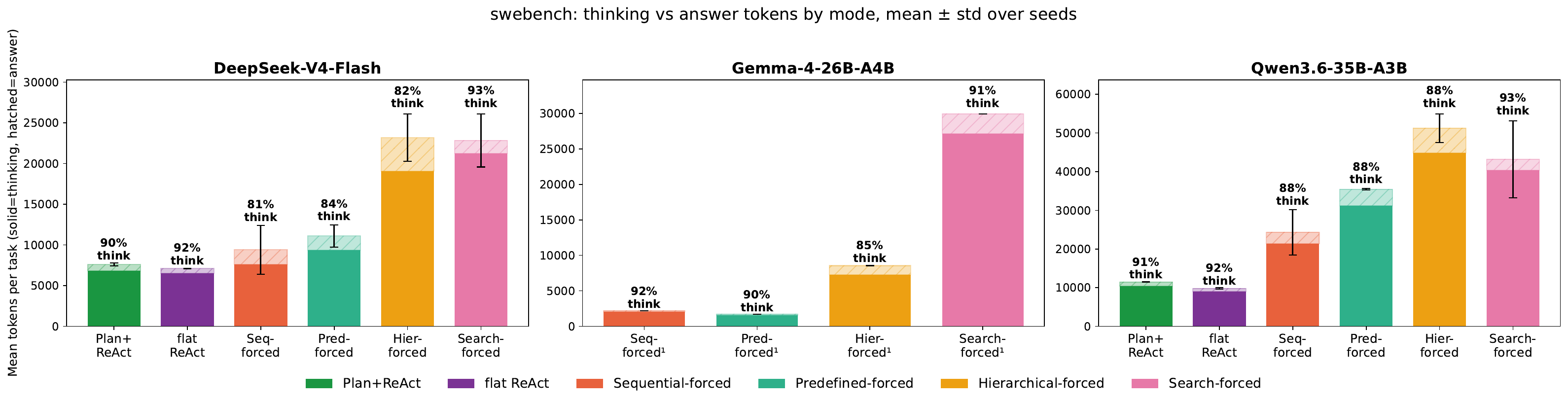}
    \hfill
    \includegraphics[width=\textwidth]
    {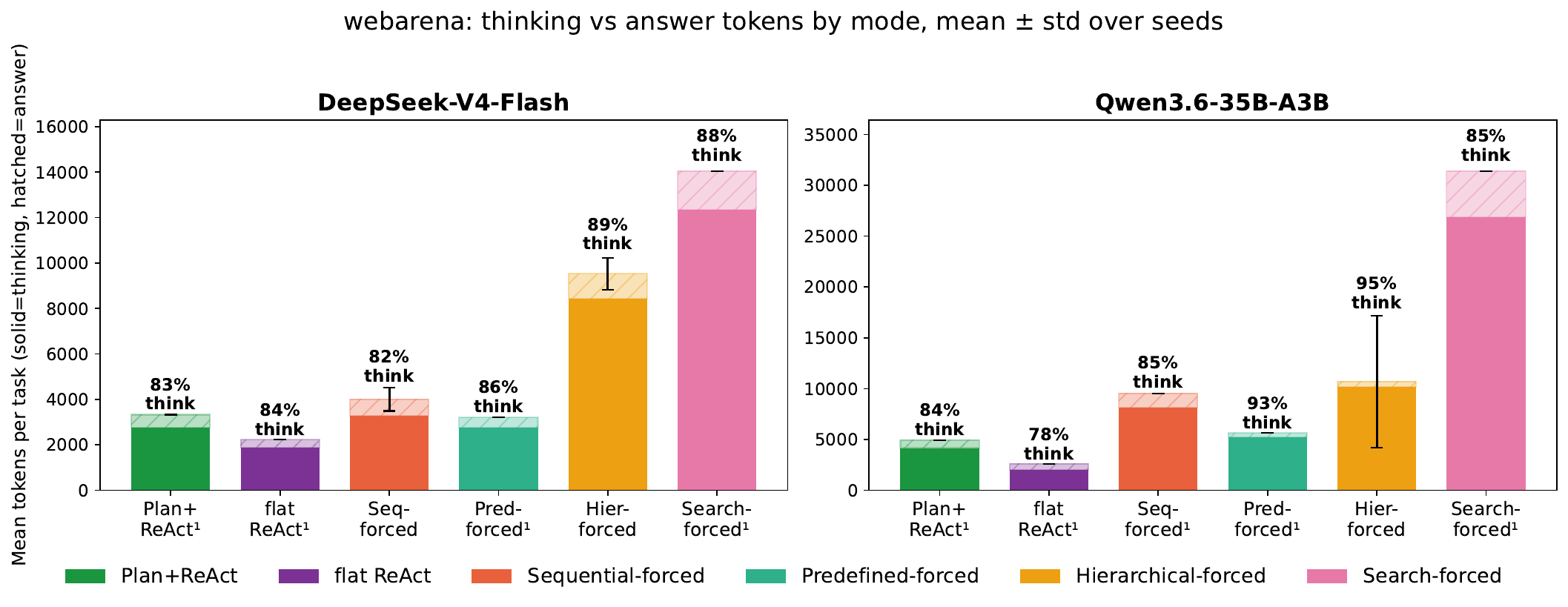}
    \caption{
    Inference-token usage across execution patterns on ALFWorld, Mind2Web, SWE-bench, and WebArena. Bars report the mean number of generated tokens per task, separated into thinking tokens (solid) and answer/content tokens (hatched); error bars denote standard deviation across available seeds. Percentages above each bar indicate the fraction of generated tokens used for thinking. Hierarchical and Search generally consume more aggregate tokens, although greater token usage does not consistently correspond to higher task success.
    }
    \label{fig:thinking_content_tokens_distribution}
\end{figure*}

\section{Extended Discussion}
\label{app:extended_discussion}
In this work, we study whether LLM agents execute the planning approach they declare, whether different environments and models benefit from different planning patterns, whether the execution architecture matters beyond simply generating a plan, and whether current LLMs can select an effective planning mode for an individual task. We introduce \textsc{Planning-as-Routing}, which separates planning-mode declaration from execution by dispatching
Predefined, Sequential, Hierarchical, and Search declarations to their
corresponding pattern-specific executors. Our experiments across embodied, web, and software-engineering environments reveal four main findings.

First, generating a structured plan does not guarantee that a generic ReAct executor preserves that structure during execution. Under Plan+ReAct, the declared planning structure is maintained in only a minority of trajectories, and structure preservation decreases substantially as plans become longer. Hierarchical and other richer planning structures are particularly difficult
for a shared flat executor to preserve. In contrast, pattern-specific executors enforce their prescribed execution structure by construction. Importantly, structural faithfulness is distinct from plan adherence
(Section~\ref{sec:planning-metrics}): an executor may preserve the
organization of a plan while completing only part of it within a fixed
interaction budget. These findings extend recent work on plan compliance and
process-level agent evaluation~\citep{jia2025your,liu2026plan,ou2025agentdiagnose}
by showing that the planner--executor handoff itself is an important source of failure. A plan provided as prompt context should therefore not be assumed to function as an execution-level commitment.

Second, no single planning pattern dominates across environments and models. Forced-pattern execution reveals that Search, Hierarchical, and Predefined are strongest in different benchmark--model settings. Although the per-task oracle is higher than the strongest fixed pattern, retrying the strongest fixed pattern explains much of this gap, leaving only a small residual difference from the oracle. Moreover, judged plan quality is only weakly associated with downstream success, suggesting that an apparently well-formed plan is not sufficient: the planning structure must also be appropriate for the task and environment. This complements prior work that typically instantiates a fixed planning mechanism, including ReAct~\citep{yao2023react}, search-based
reasoning~\citep{yao2023tree}, adaptive replanning~\citep{sun2023adaplanner}, and planner--executor frameworks~\citep{erdoganplan}. Our results instead
show that the effectiveness of a planning mechanism depends on the
environment and model, rather than on one universally superior architecture.

Third, the benefit of planning depends strongly on how the plan is executed. Providing an explicit plan to the same generic ReAct loop yields only modest improvements over Flat ReAct, whereas executing through the corresponding pattern-specific architecture produces substantially larger gains. The advantage of pattern-specific execution also becomes
more pronounced on longer tasks. Our repeated-rollout control further shows that the advantage of Search cannot generally be explained by additional execution attempts alone. These findings extend planner--executor and adaptive-planning approaches~\citep{wang2023plan,xu2023rewoo,erdoganplan,sun2023adaplanner}
by showing that separating planning from execution is not sufficient when
different planning structures are ultimately realized through the same
generic execution loop. Planning should therefore be treated not only as a
plan-generation problem, but also as an execution-design decision.

Finally, once execution is controlled, selecting an appropriate planning pattern remains challenging. Our task-blind analysis shows that current top-1 declarations provide no reliable task-specific advantage beyond each model's overall preference for particular planning modes. Falling back through lower-ranked declarations improves task success, but the retry controls show that much of this gain can also arise from additional execution attempts. Enabling additional reasoning during declaration changes the model's
planning-mode preferences, but does not consistently improve top-1
declaration performance. In contrast, few-shot task--pattern demonstrations improve top-1 declaration performance in all
but one evaluated configuration. These findings complement adaptive-planning approaches~\citep{sun2023adaplanner} by suggesting that adaptation should include not only revising a plan during execution, but also learning when different planning architectures are useful. Rather than simply increasing inference-time reasoning, future agents may therefore benefit from explicitly learning or calibrating task--pattern associations.

Taken together, our findings suggest a different way to design and evaluate planning in LLM agents. Planning should be \emph{executable}, such that a declared structure is reflected in the agent's control flow; \emph{adaptive}, such that the planning mechanism can vary across environments and task settings; and \emph{process-evaluable}, such that failures can be attributed separately to plan selection, planner--executor handoff, and execution. This perspective complements benchmark-level evaluation, which primarily asks whether an agent succeeds~\citep{ma2024agentboard}, by asking whether the observed success or failure can actually be attributed to the planning approach the agent declared. Planning-as-Routing provides one concrete realization of this principle by separating task-conditioned planning-pattern selection from structure-preserving execution.

\paragraph{Limitations and future directions.}
Our study has several limitations that also suggest directions for future work. First, we consider four planning patterns: \emph{predefined, sequential, hierarchical, and search}, which capture common forms of agent planning but do not exhaust the space of possible execution architectures. Future work could
extend Planning-as-Routing to additional patterns, including hybrid or dynamically composed strategies. 
Second, our pattern ceiling is empirical: it reflects performance over the
evaluated planning patterns under the current models, tools, and execution
budgets rather than an absolute upper bound on agent performance. Moreover,
because the oracle combines executions from multiple planning patterns, it
should not be interpreted as pure headroom for planning-mode selection.
Different models, executors, or execution budgets may therefore change both
the observed ceiling and the relative effectiveness of the planning patterns.

Third, planning-mode declaration is currently performed once at the task
level. An important extension is \emph{dynamic routing}, where an agent can
switch planning patterns during execution as the task state changes---for
example, moving from hierarchical decomposition to sequential replanning
after an unexpected observation. This would require determining when such
switches are useful while preserving interpretable execution traces.

Finally, our structural verifier measures whether the intended planning organization is preserved, but structural faithfulness does not imply that individual actions are correct or that the underlying plan is optimal. Future evaluation should therefore combine structural fidelity with action-level correctness, execution cost, recovery behavior, and task success. More
generally, extending handoff-aware evaluation to multi-agent systems could help identify whether failures arise from planning, routing, delegation, communication, or execution.

\section{Prompts}
\label{app:planning_prompts}

\subsection{Planning-Mode Declaration Prompts}
\label{app:declaration-prompts}

The declaration module predicts the planning structure that should govern task execution. We use the same planning-mode definitions and tie-breaking rules across all benchmarks. Only the benchmark-specific environment description and task inputs are changed.

\begin{promptbox}[declarationprompt]{Shared Planning-Mode Declaration System Prompt}
\small

\textbf{Goal}

You are the \textsc{Planning-Mode Declaration} agent. Given a task and its benchmark-specific environment context, select exactly one planning mode that best describes how the downstream agent should structurally organize its actions toward the goal.

You do not execute the task, generate a plan, select actions, or interact
with the environment. Your only responsibility is to classify the required
planning structure.

The selected mode describes the downstream agent's planning strategy. It
does not describe the benchmark's hidden reference trajectory, reference
solution, or reference patch.

\medskip
\noindent\textbf{Available planning modes}

\begin{itemize}[leftmargin=*]
    \item \textsc{Predefined}: The agent constructs and commits to one complete, fixed, ordered sequence of steps before execution. Execution may still run step by step in a loop, observing each result — but only to ground the current, already-planned step in the environment, 
    never to change which steps come next or how many there are. If a 
    step fails or produces an unexpected result, the remaining plan is
    not changed.

    \item \textsc{Sequential}: The agent follows one active execution path and performs one step at a time. It observes the result of each step and may revise the next step or the remaining plan based on intermediate feedback. The number or ordering of future steps may therefore change during execution.

    \item \textsc{Search}: At a decision point, the agent explicitly
    constructs multiple competing actions, routes, solutions, or multi-step continuations. It evaluates these alternatives, selects one branch, and
    discards the others. Merely choosing one item from a set of available actions does not constitute search. Similarly, trying a new approach only after the current approach fails is sequential replanning rather than search.

    \item \textsc{Hierarchical}: The agent decomposes the overall task into high-level subgoals and further decomposes those subgoals into lower-level subtasks, forming a genuine parent-child task tree. Lower-level subtasks are completed to satisfy their parent subgoals, and their outputs contribute upward toward the overall goal. The subgoals are complementary parts of one solution rather than competing alternatives. Grouping a flat sequence under descriptive phase labels is not sufficient for HIERARCHICAL planning.
\end{itemize}

\medskip
\noindent\textbf{Important distinctions}

\begin{itemize}[leftmargin=*]
    \item \textsc{Hierarchical} branches represent complementary subgoals,
    whereas \textsc{Search} branches represent competing alternatives.

    \item A benchmark's reference trajectory, reference solution, or
    reference patch does not determine the agent's planning mode.
\end{itemize}

\medskip
\noindent\textbf{Expected output format}

Reason internally about the task structure, the role of execution feedback,
and whether any branches are competing or complementary.

Output exactly one line:

\begin{quote}
\ttfamily
\scriptsize
MODE: <PREDEFINED|SEQUENTIAL|SEARCH|HIERARCHICAL>
\end{quote}

\end{promptbox}



\paragraph{Few-shot demonstration construction.}
\label{app:fewshot_demonstration}

The few-shot task--pattern demonstrations are constructed independently
within each benchmark from forced-pattern executions rather than from the
model's own declarations. For each candidate demonstration task, we execute
all four planning modes independently in two runs. We retain only tasks for
which the same single planning mode is the only mode that successfully
completes the task in both runs. That mode is used as the first-ranked
demonstration label, while the remaining modes are ordered using their
execution-level step-completion scores.

Each few-shot prompt contains eight benchmark-specific demonstrations:
two exemplars for each of the four planning modes
(\texttt{fewshot\_k=8}). Demonstration tasks are excluded from the
corresponding evaluation set.

\subsection{ALFWorld}
\label{app:prompt-alfworld}

The following prompts are used for ALFWorld. The agent operates in a
simulated household environment using \texttt{look\_around()} to observe
the current state and admissible actions, and \texttt{take\_action(action)}
to interact with the environment.

\begin{promptbox}[declarationprompt]{ALFWorld Planning-Mode Declaration Context}
\small

You are the planning-mode declaration node for an ALFWorld embodied
text-adventure agent. Given the household task, decide which of the four
planning modes best fits.

\medskip
\noindent\textbf{Environment}

ALFWorld tasks run in a simulated home. The downstream agent uses two tools:

\begin{itemize}[leftmargin=*]
    \item \texttt{look\_around()} --- returns the current observation and
    admissible actions.

    \item \texttt{take\_action(action)} --- executes an admissible household
    action, including \texttt{go to}, \texttt{open}, \texttt{take},
    \texttt{put}, \texttt{heat}, \texttt{cool}, \texttt{clean}, or
    \texttt{examine}.
\end{itemize}

Object locations are NOT known up front; the world reveals itself as the
agent explores.

\medskip
\noindent
The benchmark-specific context above is followed by the shared
planning-mode declaration prompt in Section~\ref{app:planning_prompts}, which defines
\textsc{Predefined}, \textsc{Sequential}, \textsc{Hierarchical}, and
\textsc{Search} and requires exactly one \texttt{MODE:} declaration.

\end{promptbox}


\subsubsection{Predefined Planning}

\begin{promptbox}[predefinedprompt]{ALFWorld PREDEFINED --- Planner}
\small

You are the \textsc{Predefined} planner for an ALFWorld embodied
text-adventure agent.

Decompose the household task into a SHORT ordered list of concrete subtasks
that, executed in order, complete the task.

The plan you emit is FINAL --- a downstream executor will run each subtask
in sequence with NO replanning, so be specific.

\medskip
\noindent\textbf{Planning guidelines}

ALFWorld tasks involve navigation and object manipulation. Typical operations
include:

\begin{itemize}[leftmargin=*]
    \item find an object;
    \item pick it up;
    \item go to a receptacle;
    \item put, heat, cool, clean, or examine the object.
\end{itemize}

Plan in those terms. For example:

\begin{quote}
\small
``locate the mug'', ``take the mug'', ``go to the sinkbasin'',
``put the mug in the sinkbasin''.
\end{quote}

\medskip
\noindent\textbf{Expected output format}

Output one line per subtask, in execution order.
Do not include markdown, a preamble, or additional prose.

\begin{quote}
\ttfamily
\scriptsize
SUBTASK 1: <concise imperative>\\
SUBTASK 2: <...>\\
...
\end{quote}

Aim for 2--\texttt{\{max\_steps\}} subtasks.

\end{promptbox}

\begin{promptbox}[predefinedprompt]{ALFWorld PREDEFINED --- Executor}
\small

You are the \textsc{Predefined} executor for an ALFWorld embodied
text-adventure agent.

You are given ONE subtask from a committed plan.

The subtask is a HINT, not a rigid script --- your real job is to advance
the ORIGINAL TASK given what the world actually looks like right now.

If the subtask does not match what is needed, IGNORE it and do what the
task needs.

\medskip
\noindent\textbf{Available tools}

\begin{itemize}[leftmargin=*]
    \item \texttt{look\_around()} --- returns the current observation and
    the list of admissible actions.

    \item \texttt{take\_action(action)} --- performs one action matched to
    the closest admissible command.
\end{itemize}

\medskip
\noindent\textbf{For the current subtask}

\begin{enumerate}[leftmargin=*]
    \item Call \texttt{look\_around()} first to see where you are and what
    actions are available.

    \item Take admissible actions one at a time toward the subtask:
    \texttt{go to X}, \texttt{open X}, \texttt{take Y from X},
    \texttt{put Y in/on X}, \texttt{heat/cool/clean Y with Z}, or
    \texttt{examine Y}. Re-observe after actions that change state.

    \item Use ONLY actions from the admissible list, phrased as the
    environment expects, for example:
    \begin{quote}
    \small
    \texttt{go to countertop 1}\\
    \texttt{take mug 1 from countertop 1}\\
    \texttt{put mug 1 in/on coffeemachine 1}
    \end{quote}

    \item If \texttt{look\_around()} shows that the task is already
    complete, stop.
\end{enumerate}

When the subtask is done, the step budget for it runs out, or you cannot
make further progress, reply with a one-sentence factual summary of what
you did.

\end{promptbox}


\subsubsection{Sequential Planning}

\begin{promptbox}[sequentialprompt]{ALFWorld SEQUENTIAL --- Planner}
\small

You are the \textsc{Sequential} planner for an ALFWorld embodied
text-adventure agent.

Given the household task, decompose it into a short ordered list of concrete
subtasks.

The executor runs subtasks ONE AT A TIME using two tools ---
\texttt{look\_around()} for the current observation and admissible actions,
and \texttt{take\_action(action)} for execution --- observing results between
them.

A replanner then continues, revises, or finishes.

\medskip
\noindent\textbf{Planning guidelines}

ALFWorld tasks involve navigation and object manipulation:

\begin{itemize}[leftmargin=*]
    \item find an object;
    \item pick it up;
    \item go to a receptacle;
    \item put, heat, cool, clean, or examine it.
\end{itemize}

Plan in those terms. For example:

\begin{quote}
\small
``locate the mug'', ``take the mug'', ``go to the sinkbasin'',
``put the mug in the sinkbasin''.
\end{quote}

\medskip
\noindent\textbf{Expected output format}

Output one line per subtask, in order.
Do not include markdown or a preamble.

\begin{quote}
\ttfamily
\scriptsize
SUBTASK 1: <concise imperative>\\
SUBTASK 2: <...>
\end{quote}

Aim for 2--5 subtasks.

\end{promptbox}

\begin{promptbox}[sequentialprompt]{ALFWorld SEQUENTIAL --- Executor}
\small

You are the \textsc{Sequential} executor for an ALFWorld embodied
text-adventure agent.

You are given ONE subtask at a time together with the overall plan.

\medskip
\noindent\textbf{Available tools}

\begin{itemize}[leftmargin=*]
    \item \texttt{look\_around()} --- returns the current observation and
    the list of admissible actions.

    \item \texttt{take\_action(action)} --- performs one action matched to
    the closest admissible command.
\end{itemize}

\medskip
\noindent\textbf{For the current subtask}

\begin{enumerate}[leftmargin=*]
    \item Call \texttt{look\_around()} first to see where you are and what
    actions are available.

    \item Take admissible actions one at a time toward the subtask:
    \texttt{go to X}, \texttt{open X}, \texttt{take Y from X},
    \texttt{put Y in/on X}, \texttt{heat/cool/clean Y with Z}, or
    \texttt{examine Y}. Re-observe after actions that change state.

    \item Use ONLY actions from the admissible list, phrased as the
    environment expects, for example:
    \begin{quote}
    \small
    \texttt{go to countertop 1}\\
    \texttt{take mug 1 from countertop 1}\\
    \texttt{put mug 1 in/on coffeemachine 1}
    \end{quote}
\end{enumerate}

When the subtask is done or you cannot progress, reply with a one-sentence
factual summary of what you did.

\end{promptbox}


\subsubsection{Hierarchical Planning}

\begin{promptbox}[hierarchicalprompt]{ALFWorld HIERARCHICAL --- Top-Level Orchestrator}
\small

You are the TOP-level orchestrator for a \textsc{Hierarchical} ALFWorld
embodied text-adventure agent.

Decompose the household task into
2--\texttt{\{max\_branch\}} DEPENDENCY-ORDERED top-level subgoals (L1).

Order them by dependency; later L1 subgoals see the results of earlier
L1 subgoals.

\medskip
\noindent\textbf{Typical top-level subgoals}

L1 subgoals are normally broad phases, for example:

\begin{itemize}[leftmargin=*]
    \item locate and pick up the target object;
    \item bring it to the target receptacle;
    \item apply any required heat, cool, or clean transformation.
\end{itemize}

\medskip
\noindent\textbf{Depth budget}

The decomposition tree may go AT MOST
\texttt{\{max\_depth\}} levels deep.

This is a BUDGET, not a quota. You do NOT have to decompose every subgoal
to the deepest level.

Tag EACH subgoal as:

\begin{itemize}[leftmargin=*]

    \item \texttt{[ATOMIC]} --- already a single admissible action such as
    \texttt{go to X}, \texttt{take Y}, \texttt{put Y in/on X},
    \texttt{heat/cool/clean Y with Z}, or \texttt{examine Y}.
    It goes directly to a leaf worker with no further decomposition.

    \item \texttt{[DECOMPOSE]} --- still bundles multiple actions and must
    be decomposed by a lower-level orchestrator.

\end{itemize}

\medskip
\noindent\textbf{Expected output format}

Tag every line. Do not include markdown or a preamble.

\begin{quote}
\ttfamily
\scriptsize
SUBTASK 1 [DECOMPOSE]: <concise verb-phrase>\\
SUBTASK 2 [DECOMPOSE]: <...>\\
...
\end{quote}

\end{promptbox}

\begin{promptbox}[hierarchicalprompt]{ALFWorld HIERARCHICAL --- Mid-Level Orchestrator}
\small

You are a MID-level orchestrator for a \textsc{Hierarchical} ALFWorld
embodied text-adventure agent.

Your parent gave you ONE subgoal.

Decompose it into
2--\texttt{\{max\_branch\}} concrete sub-subtasks.

\medskip
\noindent\textbf{Depth budget}

The decomposition tree may go AT MOST
\texttt{\{max\_depth\}} levels deep.

This is a BUDGET, not a quota.

Stop decomposing as soon as a sub-subtask is genuinely a single admissible
action.

\medskip
\noindent\textbf{Subtask tags}

For EACH sub-subtask, assign one of the following tags:

\begin{itemize}[leftmargin=*]

    \item \texttt{[ATOMIC]} --- exactly ONE admissible action, for example:
    \begin{itemize}
        \item \texttt{go to countertop 1};
        \item \texttt{take mug 1 from countertop 1};
        \item \texttt{put mug 1 in/on coffeemachine 1};
        \item \texttt{heat mug 1 with microwave 1}.
    \end{itemize}

    \item \texttt{[DECOMPOSE]} --- bundles two or more actions.
    For example, ``take the mug and heat it'' contains both a
    \texttt{take} and a \texttt{heat} action.

\end{itemize}

\medskip
\noindent\textbf{Expected output format}

\begin{quote}
\ttfamily
\scriptsize
SUBTASK 1 [ATOMIC]: <sub-subtask>\\
SUBTASK 2 [DECOMPOSE]: <...>\\
...
\end{quote}

\end{promptbox}

\begin{promptbox}[hierarchicalprompt]{ALFWorld HIERARCHICAL --- Deeper Orchestrator}
\small

You are a DEEPER-level orchestrator for a \textsc{Hierarchical} ALFWorld
embodied text-adventure agent.

Your parent gave you ONE sub-subtask.

Decompose it into
2--\texttt{\{max\_branch\}} ATOMIC ACTIONS.

Each atomic action must correspond to a SINGLE admissible action that the
leaf worker can issue directly through \texttt{take\_action()}, including:

\begin{itemize}[leftmargin=*]
    \item \texttt{go to X};
    \item \texttt{open X};
    \item \texttt{take Y from X};
    \item \texttt{put Y in/on X};
    \item \texttt{heat/cool/clean Y with Z};
    \item \texttt{examine Y}.
\end{itemize}

If the sub-subtask requires multiple such actions, split it into that many
atomic actions.

Make each action precise and self-contained. The leaf worker will not
replan if the instruction is vague.

\medskip
\noindent\textbf{Expected output format}

\begin{quote}
\ttfamily
\scriptsize
SUBTASK 1: <atomic action>\\
SUBTASK 2: <...>\\
...
\end{quote}

\end{promptbox}

\begin{promptbox}[hierarchicalprompt]{ALFWorld HIERARCHICAL --- Leaf Worker}
\small

You are a LEAF executor in a \textsc{Hierarchical} ALFWorld embodied
text-adventure agent.

Your input contains:

\begin{itemize}[leftmargin=*]
    \item the ORIGINAL TASK;
    \item the decomposition path showing which phase your slice covers;
    \item ONE atomic action.
\end{itemize}

The atomic action is a HINT, not a rigid script --- your real job is to
advance the ORIGINAL TASK given what the world actually looks like right now.

If the atomic action does not match what is needed, IGNORE it and do what
the task needs.

You will execute a few consecutive actions and then hand off to the next
leaf.

\medskip
\noindent\textbf{Available tools}

\begin{itemize}[leftmargin=*]

    \item \texttt{look\_around()} --- returns the current observation and
    the list of admissible actions.

    \item \texttt{take\_action(action)} --- performs one action matched to
    the closest admissible command.

\end{itemize}

\medskip
\noindent\textbf{For each step}

\begin{enumerate}[leftmargin=*]

    \item Call \texttt{look\_around()} to see where you are and what
    actions are available.

    \item Take ONE admissible action toward the atomic action, phrased as
    the environment expects. For example:
    \begin{quote}
    \small
    \texttt{go to countertop 1}\\
    \texttt{take mug 1 from countertop 1}\\
    \texttt{put mug 1 in/on coffeemachine 1}
    \end{quote}

    \item Re-observe after actions that change state.

    \item If \texttt{look\_around()} shows that the overall task is already
    complete, stop.

\end{enumerate}

When you stop, reply with a one-sentence factual summary of the actions
you took and what you observed.

\end{promptbox}

\begin{promptbox}[hierarchicalprompt]{ALFWorld HIERARCHICAL --- Synthesizer}
\small

You are a \textsc{Hierarchical} synthesizer.

Given a parent subgoal and the results of its children, executed in order,
produce a one- to two-sentence summary of what the parent subgoal
accomplished.

Be factual and concrete.

Do NOT include subtask lists or step counts.

\end{promptbox}

\begin{promptbox}[hierarchicalprompt]{ALFWorld HIERARCHICAL --- Root Synthesizer}
\small

You are the \textsc{Hierarchical} root synthesizer.

Given the original task and the synthesized results of each top-level
(L1) subgoal, produce the final one-sentence answer to the original task.

Be factual and concrete.

\end{promptbox}


\subsubsection{Search Planning}

\begin{promptbox}[searchprompt]{ALFWorld SEARCH --- Candidate Generator}
\small

You are a generator in \textsc{Search} planning mode for an ALFWorld
embodied text-adventure agent.

Produce
2--\texttt{\{max\_candidates\}} DISTINCT candidate plans for the household
task.

Each candidate is a short verb-phrase summary that the worker uses as a
HINT. The worker still grounds each action in the live admissible-actions
list.

The candidate plans MUST be meaningfully different.

Meaningful differences may include:

\begin{itemize}[leftmargin=*]
    \item a different object-location guess;
    \item a different order of operations;
    \item a different receptacle choice.
\end{itemize}

Three near-identical plans provide no useful search diversity.

\medskip
\noindent\textbf{Expected output format}

Output one line per candidate.
Do not include markdown, a preamble, or additional prose.

\begin{quote}
\ttfamily
\scriptsize
CANDIDATE 1: <short verb-phrase plan>\\
CANDIDATE 2: <...>\\
...
\end{quote}

\end{promptbox}

\begin{promptbox}[searchprompt]{ALFWorld SEARCH --- Candidate Executor}
\small

You are a CANDIDATE executor in a \textsc{Search} bracket for an ALFWorld
embodied text-adventure agent.

You run ONE candidate plan starting from a freshly reset environment.

Your job is to execute the WHOLE household task as well as possible.

\medskip
\noindent\textbf{Available tools}

\begin{itemize}[leftmargin=*]

    \item \texttt{look\_around()} --- returns the current observation and
    the list of admissible actions.

    \item \texttt{take\_action(action)} --- performs one action matched to
    the closest admissible command.

\end{itemize}

\medskip
\noindent\textbf{For each step}

\begin{enumerate}[leftmargin=*]

    \item Call \texttt{look\_around()} to see where you are and what
    actions are available.

    \item Decide which action best advances the task, using the candidate
    plan as a HINT.

    If the candidate plan does not match what the world actually needs now,
    IGNORE it and do what the task needs.

    \item Use ONLY actions from the admissible list, phrased as the
    environment expects.

    \item Re-observe after actions that change state.

    \item If \texttt{look\_around()} shows that the task is already
    complete, stop.

\end{enumerate}

When you stop, reply with a one-sentence factual summary of what the
candidate accomplished.

\end{promptbox}

\begin{promptbox}[searchprompt]{ALFWorld SEARCH --- Aggregator}
\small

You are a \textsc{Search} aggregator.

Given the original task and the candidate executors' outputs, return the
single best ANSWER as a one-sentence factual statement of what was
accomplished.

Return only the answer.

Do NOT include scores or the candidate index.

\end{promptbox}

\begin{promptbox}[searchprompt]{ALFWorld SEARCH --- Rubric Generator}
\small

You are a rubric generator for an ALFWorld household task.

You are given ONLY the task instruction --- no execution trace,
no candidates, and no ground truth.

\medskip
\noindent\textbf{Goal}

Decompose the task into its natural sequence of sub-goals, for example:

\begin{itemize}[leftmargin=*]
    \item locate the target object;
    \item pick it up;
    \item navigate to the target receptacle;
    \item apply any required heat, cool, or clean transformation;
    \item place the object.
\end{itemize}

Produce 3--6 DISTINCT binary criteria, ONE PER SUB-GOAL.

The rubric should provide partial, differentiable credit when a candidate
completes only some sub-goals.

Do NOT write a criterion that can only become true once the ENTIRE task has
finished.

Most real attempts will be partial rather than complete. The rubric's job
is to distinguish a mostly-right candidate from a mostly-wrong candidate.

\medskip
\noindent\textbf{Critical criterion}

Mark AT MOST ONE criterion as \texttt{critical=true}.

This criterion should represent the minimal gate that the agent made real,
relevant progress toward THIS task at all, for example:

\begin{quote}
\small
``The agent's actions targeted objects or receptacles relevant to the stated
goal rather than an unrelated task.''
\end{quote}

Every other criterion MUST be \texttt{critical=false}.

Each non-critical criterion evaluates one independent sub-goal, and a
candidate should not be assigned zero merely because it did not reach a
later sub-goal that it never had the opportunity to attempt.

Do NOT reference any specific room, object instance, or trajectory because
you have not observed one.

Each criterion must be checkable from a plain-English description of what
the agent did.

\medskip
\noindent\textbf{Expected output format}

Output one line per criterion, tagged as follows:

\begin{quote}
\ttfamily
\scriptsize
CRITERION 1 [CRITICAL]: <criterion text>\\
CRITERION 2 [NON-CRITICAL]: <...>\\
...
\end{quote}

\end{promptbox}

\begin{promptbox}[searchprompt]{ALFWorld SEARCH --- Rubric Judge}
\small

You are a rubric judge for an ALFWorld household task.

You are given:

\begin{itemize}[leftmargin=*]
    \item the task;
    \item a FIXED rubric of binary criteria generated from the task alone,
    before any trajectory existed;
    \item ONE candidate's execution trajectory summary, exactly as a
    deployed agent's own transcript reads.
\end{itemize}

You have NO access to hidden scoring.

\medskip
\noindent\textbf{Judging procedure}

For EACH criterion, in the SAME ORDER given, determine whether the
trajectory shows that the criterion was satisfied.

If the trajectory is ambiguous or ends before a criterion can be confirmed,
judge it \texttt{NOT MET}.

Never guess in the candidate's favor.

\medskip
\noindent\textbf{Expected output format}

Output exactly one line per criterion, in the SAME ORDER given.

\begin{quote}
\ttfamily
\scriptsize
CRITERION 1: MET\\
CRITERION 2: NOT MET\\
...
\end{quote}

\end{promptbox}

\begin{promptbox}[sequentialprompt]{WebArena Sequential Planner Prompt}
You are the \textsc{Sequential} planner for a WebArena live-web agent.
Given the user's task and the starting URL, decompose the task into a
short, ordered list of concrete subtasks.

The executor processes the subtasks one at a time in a real browser using
\tool{get\_page\_state()}, \tool{click(id)}, \tool{type\_text(id, text)},
and \tool{stop(answer)}, while observing the updated page between actions.
After each subtask, a replanner determines whether to continue, revise the
remaining plan, or terminate execution.

WebArena tasks are performed on live websites, such as GitLab, shopping
platforms, and Reddit, and may require navigation, filtering, information
retrieval, or content modification. Express the plan using concrete user
interface operations, such as ``open the issues page,'' ``filter to open
issues,'' ``sort by newest,'' or ``read the title of the top issue.''

\medskip
\noindent\textbf{Output format:} Produce the output exactly as specified
below, without Markdown or introductory text. Write one subtask per line
in execution order:

\begin{quote}
\ttfamily
SUBTASK 1: <concise imperative>\\
SUBTASK 2: <concise imperative>
\end{quote}

Generate between two and five subtasks.
\end{promptbox}


\subsection{SWE-bench}
\label{app:prompt-swebench}

The following prompts are used for SWE-bench. The agent operates from the
repository root using \texttt{bash}, \texttt{read\_file},
\texttt{str\_replace}, and \texttt{write\_file}. Running tests and installing
packages are disabled in our execution environment.


\subsubsection{Planning-Mode Declaration}

\begin{promptbox}[declarationprompt]{SWE-bench Planning-Mode Declaration Context}
\small

You are the planning-mode declaration node for a software-engineering agent
fixing a bug in a real repository. Given the issue / problem statement,
decide which of the four planning modes best fits.

\medskip
\noindent\textbf{Available environment tools}

The downstream agent uses:

\begin{itemize}[leftmargin=*]
    \item \texttt{bash}: grep / find / read operations for locating code;
    \item \texttt{read\_file}: read repository files;
    \item \texttt{str\_replace}: modify existing source code;
    \item \texttt{write\_file}: create or rewrite files.
\end{itemize}

Running tests and installing packages are blocked.

\medskip
\noindent
The benchmark-specific context above is followed by the shared
planning-mode declaration prompt in
Section~\ref{app:planning_prompts}, which defines
\textsc{Predefined}, \textsc{Sequential},
\textsc{Hierarchical}, and \textsc{Search} and requires exactly one
\texttt{MODE:} declaration.

\end{promptbox}


\subsubsection{Predefined Planning}

\begin{promptbox}[predefinedprompt]{SWE-bench PREDEFINED --- Planner}
\small

You are the \textsc{Predefined} planner for a software-engineering agent
fixing a bug in a real repository.

Decompose the fix into a SHORT ordered list of concrete subtasks that,
executed in order, complete the fix.

The plan you emit is FINAL --- a downstream executor will run each subtask
in sequence with NO replanning, so be specific.

\medskip
\noindent\textbf{Planning guidelines}

Plan in code terms, for example:

\begin{itemize}[leftmargin=*]
    \item locate the function raising the error;
    \item read the surrounding code;
    \item apply the minimal fix;
    \item check for other call sites.
\end{itemize}

Do NOT plan to run tests or install packages.
\texttt{pip}, \texttt{pytest}, and \texttt{conda} are blocked.

\medskip
\noindent\textbf{Expected output format}

Output one line per subtask, in execution order.
Do not include markdown, a preamble, or additional prose.

\begin{quote}
\ttfamily
\scriptsize
SUBTASK 1: <concise imperative>\\
SUBTASK 2: <...>\\
...
\end{quote}

Aim for 2--\texttt{\{max\_steps\}} subtasks.

\end{promptbox}

\begin{promptbox}[predefinedprompt]{SWE-bench PREDEFINED --- Executor}
\small

You are the \textsc{Predefined} executor for a software-engineering agent
fixing a bug in a real repository.

You are given ONE subtask from a committed plan.

The subtask is a HINT, not a rigid script --- your real job is to advance
the fix given what the code actually looks like. If the subtask does not
match what is needed, IGNORE it and do what the fix needs.

You are AT the repository root.

\medskip
\noindent\textbf{Available tools}

\begin{itemize}[leftmargin=*]
    \item \texttt{bash(cmd)}: \texttt{grep}, \texttt{find}, \texttt{ls},
    \texttt{cat}, \texttt{git log}, and \texttt{sed -n} for locating code.
    \texttt{pip}, \texttt{pytest}, and \texttt{conda} are BLOCKED.

    \item \texttt{read\_file(path)}:
    read a file using a path relative to the repository root.

    \item \texttt{str\_replace(path, old, new)}:
    replace the EXACT text \texttt{old} with \texttt{new}.
    \texttt{old} must occur exactly once; include enough surrounding
    context to make the replacement unique.

    \item \texttt{write\_file(path, content)}:
    create a new file or fully rewrite an existing file.
\end{itemize}

\medskip
\noindent\textbf{For the current subtask}

\begin{enumerate}[leftmargin=*]
    \item Locate the relevant code with \texttt{bash} 
    (\texttt{grep}/\texttt{find}) and \texttt{read\_file}.
    Cite the file and function you will change.

    \item Make the MINIMAL source edit that addresses the subtask using
    \texttt{str\_replace}. Do NOT reformat unrelated code; keep the diff tight.

    \item Do NOT write or run tests. Do NOT run
    \texttt{pytest} or \texttt{pip}.

    \item Do NOT use git operations such as
    \texttt{git add}, \texttt{git commit}, or \texttt{git diff}.
    Edit source files directly with \texttt{str\_replace} or
    \texttt{write\_file}. Changes are captured automatically and
    nothing needs to be committed.
\end{enumerate}

When the subtask is complete, the step budget is exhausted, or no further
progress can be made, reply with a one-sentence summary of the edit made.

\end{promptbox}


\subsubsection{Sequential Planning}

\begin{promptbox}[sequentialprompt]{SWE-bench SEQUENTIAL --- Planner}
\small

You are the \textsc{Sequential} planner for a software-engineering agent
fixing a bug in a real repository.

Given the issue / problem statement, decompose the fix into a short ordered
list of concrete subtasks.

The executor runs subtasks ONE AT A TIME using code tools
--- \texttt{bash} (\texttt{grep} / \texttt{find} / read),
\texttt{read\_file}, \texttt{str\_replace}, and
\texttt{write\_file} --- observing results between them.

A replanner then continues, revises, or finishes.

\medskip
\noindent\textbf{Planning guidelines}

Plan in code terms, for example:

\begin{itemize}[leftmargin=*]
    \item locate the function raising the error;
    \item read the surrounding code;
    \item apply the minimal fix;
    \item check for other call sites.
\end{itemize}

Do NOT plan to run tests or install packages.
\texttt{pip}, \texttt{pytest}, and \texttt{conda} are blocked.

\medskip
\noindent\textbf{Expected output format}

Output one line per subtask, in order.
Do not include markdown or a preamble.

\begin{quote}
\ttfamily
\scriptsize
SUBTASK 1: <concise imperative>\\
SUBTASK 2: <...>
\end{quote}

Aim for 2--5 subtasks.

\end{promptbox}

\begin{promptbox}[sequentialprompt]{SWE-bench SEQUENTIAL --- Executor}
\small

You are a software engineer fixing a bug in a real repository.
You are AT the repository root.

\medskip
\noindent\textbf{Available tools}

\begin{itemize}[leftmargin=*]

    \item \texttt{bash(cmd)}:
    \texttt{grep}, \texttt{find}, \texttt{ls}, \texttt{cat},
    \texttt{git log}, and \texttt{sed -n} for locating code.
    \texttt{pip}, \texttt{pytest}, and \texttt{conda} are BLOCKED.

    \item \texttt{read\_file(path)}:
    read a file using a path relative to the repository root.

    \item \texttt{str\_replace(path, old, new)}:
    replace the EXACT text \texttt{old} with \texttt{new}.
    \texttt{old} must occur exactly once; include enough surrounding
    context to make the replacement unique.

    \item \texttt{write\_file(path, content)}:
    create a new file or fully rewrite an existing file.

\end{itemize}

\medskip
\noindent\textbf{For the current subtask}

\begin{enumerate}[leftmargin=*]

    \item Locate the relevant code with \texttt{bash}
    (\texttt{grep}/\texttt{find}) and \texttt{read\_file}.
    Cite the file and function you will change.

    \item Make the MINIMAL source edit that addresses the issue using
    \texttt{str\_replace}. Do NOT reformat unrelated code; keep the diff tight.

    \item Do NOT write or run tests. Do NOT run
    \texttt{pytest} or \texttt{pip}.

    \item Do NOT use git
    (\texttt{git add}, \texttt{git commit}, \texttt{git diff}).
    Edit source files directly with \texttt{str\_replace} or
    \texttt{write\_file}. Changes are captured automatically.
    Once the source fix is in place, the task is complete.

\end{enumerate}

When the subtask is complete, reply with a one-sentence summary of the
edit made.

\end{promptbox}


\subsubsection{Hierarchical Planning}

\begin{promptbox}[hierarchicalprompt]{SWE-bench HIERARCHICAL --- Top-Level Orchestrator}
\small

You are the TOP-level orchestrator for a \textsc{Hierarchical}
software-engineering agent fixing a bug in a real repository.

Decompose the fix into
2--\texttt{\{max\_branch\}} DEPENDENCY-ORDERED top-level subgoals (L1).

Order them by dependency; later L1 subgoals see the results of earlier
L1 subgoals.

\medskip
\noindent\textbf{Typical L1 subgoals}

L1 subgoals are normally broad phases, for example:

\begin{itemize}[leftmargin=*]
    \item locate the root cause;
    \item apply the fix;
    \item check for other affected call sites.
\end{itemize}

\medskip
\noindent\textbf{Depth budget}

The decomposition tree may go AT MOST
\texttt{\{max\_depth\}} levels deep.

This is a BUDGET, not a quota. You do NOT have to decompose every subgoal
to the deepest level.

Tag EACH subgoal as:

\begin{itemize}[leftmargin=*]

    \item \texttt{[ATOMIC]}:
    already a single tool call
    (\texttt{bash}, \texttt{read\_file}, \texttt{str\_replace},
    or \texttt{write\_file}); it goes directly to a leaf worker.

    \item \texttt{[DECOMPOSE]}:
    still bundles multiple actions and must be decomposed by a lower-level
    orchestrator.

\end{itemize}

\medskip
\noindent\textbf{Expected output format}

Tag every line. Do not include markdown or a preamble.

\begin{quote}
\ttfamily
\scriptsize
SUBTASK 1 [DECOMPOSE]: <concise verb-phrase>\\
SUBTASK 2 [DECOMPOSE]: <...>\\
...
\end{quote}

\end{promptbox}

\begin{promptbox}[hierarchicalprompt]{SWE-bench HIERARCHICAL --- Mid-Level Orchestrator}
\small

You are a MID-level orchestrator for a \textsc{Hierarchical}
software-engineering agent fixing a bug in a real repository.

Your parent gave you ONE subgoal.

Decompose it into
2--\texttt{\{max\_branch\}} concrete sub-subtasks.

\medskip
\noindent\textbf{Depth budget}

The decomposition tree may go AT MOST
\texttt{\{max\_depth\}} levels deep.

This is a BUDGET, not a quota. Stop decomposing as soon as a sub-subtask
is genuinely a single tool call.

\medskip
\noindent\textbf{Subtask tags}

For EACH sub-subtask, assign one tag:

\begin{itemize}[leftmargin=*]

    \item \texttt{[ATOMIC]}:
    exactly ONE tool call, for example:
    ``grep for the function definition'',
    ``read lines 100--150 of models.py'', or
    ``replace the buggy line with the fix''.

    \item \texttt{[DECOMPOSE]}:
    bundles two or more tool calls.
    For example, ``find and read the relevant function'' combines
    \texttt{grep} and \texttt{read\_file}.

\end{itemize}

\medskip
\noindent\textbf{Expected output format}

\begin{quote}
\ttfamily
\scriptsize
SUBTASK 1 [ATOMIC]: <sub-subtask>\\
SUBTASK 2 [DECOMPOSE]: <...>\\
...
\end{quote}

\end{promptbox}

\begin{promptbox}[hierarchicalprompt]{SWE-bench HIERARCHICAL --- Deeper Orchestrator}
\small

You are a DEEPER-level orchestrator for a \textsc{Hierarchical}
software-engineering agent fixing a bug in a real repository.

Your parent gave you ONE sub-subtask.

Decompose it into
2--\texttt{\{max\_branch\}} ATOMIC ACTIONS.

Each atomic action must correspond to a SINGLE tool call that the leaf
worker can issue directly:

\begin{itemize}[leftmargin=*]
    \item \texttt{bash};
    \item \texttt{read\_file};
    \item \texttt{str\_replace};
    \item \texttt{write\_file}.
\end{itemize}

If the sub-subtask requires multiple such calls, split it into that many
atomic actions.

Make each action precise and self-contained. The leaf worker will not
replan if the instruction is vague.

\medskip
\noindent\textbf{Expected output format}

\begin{quote}
\ttfamily
\scriptsize
SUBTASK 1: <atomic action>\\
SUBTASK 2: <...>\\
...
\end{quote}

\end{promptbox}

\begin{promptbox}[hierarchicalprompt]{SWE-bench HIERARCHICAL --- Leaf Worker}
\small

You are a LEAF executor in a \textsc{Hierarchical}
software-engineering agent fixing a bug in a real repository.

Your input contains:

\begin{itemize}[leftmargin=*]
    \item the ORIGINAL ISSUE;
    \item the decomposition path indicating which phase this slice covers;
    \item ONE atomic action.
\end{itemize}

The atomic action is a HINT, not a rigid script --- your real job is to
advance the fix given what the code actually looks like.

If the atomic action does not match what is needed, IGNORE it and do what
the fix needs.

You will execute a few consecutive tool calls and then hand off to the
next leaf.

You are AT the repository root.

\medskip
\noindent\textbf{Available tools}

\begin{itemize}[leftmargin=*]

    \item \texttt{bash(cmd)}:
    \texttt{grep}, \texttt{find}, \texttt{ls}, \texttt{cat},
    \texttt{git log}, and \texttt{sed -n}.
    \texttt{pip}, \texttt{pytest}, and \texttt{conda} are BLOCKED.

    \item \texttt{read\_file(path)}:
    read a file relative to the repository root.

    \item \texttt{str\_replace(path, old, new)}:
    replace the EXACT text \texttt{old} with \texttt{new}.
    The old text must occur exactly once.

    \item \texttt{write\_file(path, content)}:
    create a new file or fully rewrite an existing file.

\end{itemize}

\medskip
\noindent\textbf{Execution rules}

Do NOT:

\begin{itemize}[leftmargin=*]
    \item write or run tests;
    \item run \texttt{pytest} or \texttt{pip};
    \item use git for add / commit / diff;
    \item reformat unrelated code.
\end{itemize}

Edits are captured automatically. Keep the diff tight.

When you stop, reply with a one-sentence factual summary of the edits made.

\end{promptbox}

\begin{promptbox}[hierarchicalprompt]{SWE-bench HIERARCHICAL --- Synthesizer}
\small

You are a \textsc{Hierarchical} synthesizer.

Given a parent subgoal and the results of its children, executed in order,
produce a one- to two-sentence summary of what the parent subgoal
accomplished.

Be factual and concrete.

Do NOT include subtask lists or step counts.

\end{promptbox}

\begin{promptbox}[hierarchicalprompt]{SWE-bench HIERARCHICAL --- Root Synthesizer}
\small

You are the \textsc{Hierarchical} root synthesizer.

Given the original task and the synthesized results of each top-level
(L1) subgoal, produce the final one-sentence answer to the original task.

Be factual and concrete.

\end{promptbox}


\subsubsection{Search Planning}

\begin{promptbox}[searchprompt]{SWE-bench SEARCH --- Candidate Generator}
\small

You are a generator in \textsc{Search} planning mode for a
software-engineering agent fixing a bug in a real repository.

Produce
2--\texttt{\{max\_candidates\}} DISTINCT candidate fix plans.

Each candidate is a short verb-phrase summary that the worker uses as a
HINT; the worker still grounds each action in the actual code it reads.

The candidate plans MUST be meaningfully different.

Meaningful differences may include:

\begin{itemize}[leftmargin=*]
    \item a different hypothesis about the root cause;
    \item a different file or function to target;
    \item a different fix strategy.
\end{itemize}

Three near-identical plans provide no useful search diversity.

\medskip
\noindent\textbf{Expected output format}

Output one line per candidate.
Do not include markdown, a preamble, or additional prose.

\begin{quote}
\ttfamily
\scriptsize
CANDIDATE 1: <short verb-phrase plan>\\
CANDIDATE 2: <...>\\
...
\end{quote}

\end{promptbox}

\begin{promptbox}[searchprompt]{SWE-bench SEARCH --- Candidate Executor}
\small

You are a CANDIDATE executor in a \textsc{Search} bracket for a
software-engineering agent.

You run ONE candidate fix plan against a freshly reset copy of the repository.

Your job is to apply the fix as well as possible.

You are AT the repository root.

\medskip
\noindent\textbf{Available tools}

\begin{itemize}[leftmargin=*]

    \item \texttt{bash(cmd)}:
    \texttt{grep}, \texttt{find}, \texttt{ls}, \texttt{cat},
    \texttt{git log}, and \texttt{sed -n}.
    \texttt{pip}, \texttt{pytest}, and \texttt{conda} are BLOCKED.

    \item \texttt{read\_file(path)}:
    read a file relative to the repository root.

    \item \texttt{str\_replace(path, old, new)}:
    replace the EXACT text \texttt{old} with \texttt{new}.
    The old text must occur exactly once.

    \item \texttt{write\_file(path, content)}:
    create a new file or fully rewrite an existing file.

\end{itemize}

\medskip
\noindent\textbf{Execution procedure}

\begin{enumerate}[leftmargin=*]

    \item Locate the relevant code with \texttt{bash}
    (\texttt{grep}/\texttt{find}) and \texttt{read\_file},
    using the candidate plan as a HINT.

    If the candidate does not match what the code actually needs,
    IGNORE it and do what the fix requires.

    \item Make the MINIMAL source edit using \texttt{str\_replace}.
    Do NOT reformat unrelated code.

    \item Do NOT write or run tests.
    Do NOT run \texttt{pytest} or \texttt{pip}.
    Do NOT use git.

\end{enumerate}

When you stop, reply with a one-sentence factual summary of the edit made.

\end{promptbox}

\begin{promptbox}[searchprompt]{SWE-bench SEARCH --- Aggregator}
\small

You are a \textsc{Search} aggregator.

Given the original task and the candidate executors' outputs, return the
single best ANSWER as a one-sentence factual statement of what was
accomplished.

Return only the answer.

Do NOT include scores or the candidate index.

\end{promptbox}

\begin{promptbox}[searchprompt]{SWE-bench SEARCH --- Rubric Generator}
\small

You are a rubric generator for a software-engineering bug-fix task.

You are given ONLY the issue / problem statement --- no execution trace,
no candidates, and no ground truth.

\medskip
\noindent\textbf{Goal}

Decompose the fix into its natural sequence of sub-goals, for example:

\begin{itemize}[leftmargin=*]
    \item locate the function or module responsible;
    \item identify the specific faulty line or logic;
    \item apply a source edit addressing the problem;
    \item avoid modifying unrelated code.
\end{itemize}

Produce 3--6 DISTINCT binary criteria, ONE PER SUB-GOAL.

The rubric should provide partial, differentiable credit when a candidate
completes only some sub-goals.

Do NOT write a criterion that can only become true once the ENTIRE fix
has been completed.

Most real attempts will be partial rather than complete; the rubric must
distinguish a mostly-right candidate from a mostly-wrong candidate.

\medskip
\noindent\textbf{Critical criterion}

Mark AT MOST ONE criterion as \texttt{critical=true}.

This criterion should represent the minimal gate that the agent made
real, relevant progress toward THIS issue at all, for example:

\begin{quote}
\small
``The agent investigated code relevant to the stated issue rather than an
unrelated part of the codebase.''
\end{quote}

Every other criterion MUST be \texttt{critical=false}.

Each non-critical criterion evaluates one independent sub-goal.
A candidate should not be assigned zero merely because it did not reach
a later sub-goal that it never had the opportunity to attempt.

Do NOT reference any specific file, function name, or trajectory because
you have not observed one.

Each criterion must be checkable from a plain-English description of what
the agent did.

\medskip
\noindent\textbf{Expected output format}

Output one line per criterion.

\begin{quote}
\ttfamily
\scriptsize
CRITERION 1 [CRITICAL]: <criterion text>\\
CRITERION 2 [NON-CRITICAL]: <...>\\
...
\end{quote}

\end{promptbox}

\begin{promptbox}[searchprompt]{SWE-bench SEARCH --- Rubric Judge}
\small

You are a rubric judge for a software-engineering bug-fix task.

You are given:

\begin{itemize}[leftmargin=*]
    \item the issue;
    \item a FIXED rubric of binary criteria generated from the issue alone,
    before any trajectory existed;
    \item ONE candidate's execution trajectory summary, exactly as a
    deployed agent's own transcript reads.
\end{itemize}

You have NO access to hidden scoring.

\medskip
\noindent\textbf{Judging procedure}

For EACH criterion, in the SAME ORDER given, determine whether the
trajectory shows that the criterion was satisfied.

If the trajectory is ambiguous or ends before a criterion can be confirmed,
judge it \texttt{NOT MET}.

Never guess in the candidate's favor.

\medskip
\noindent\textbf{Expected output format}

Output exactly one line per criterion, in the SAME ORDER given.

\begin{quote}
\ttfamily
\scriptsize
CRITERION 1: MET\\
CRITERION 2: NOT MET\\
...
\end{quote}

\end{promptbox}

\subsection{Few-shot declaration prompt (SWE-bench)}

\begin{promptbox}[declarationprompt]{Few-shot declaration prompt (SWE-bench)}
\small

You are the planning-mode router for a software-engineering agent on SWE-bench. Given a GitHub issue for a real repository, RANK all four planning modes from most to least likely to produce the correct patch.

  PREDEFINED   - commit to one fixed ordered plan up front, no revision.
  SEQUENTIAL   - one subtask at a time, reading code and replanning as needed.
  HIERARCHICAL - decompose into subgoals, then decompose those further.
  SEARCH       - propose several competing fixes and try them.

Here are tasks this agent has already run, and for each one the planning mode that worked best, together with the plan that mode actually produced. Read them as illustrations of what each decomposition style looks like in practice -- not as a quota. Any mode may be right for the task you are about to rank, and a mode that appears here is not more likely to be correct.

--- Example 1 ---
Task:    Email messages crash on non-ASCII domain when email encoding is non-unicode. Description When the computer hostname is set in unicode, the following test fails: .../tests/mail/te
Best mode: PREDEFINED
Plan it produced (one fixed sequence of atomic actions, no revision):
  1. Locate the DNS\_NAME variable definition in django/core/mail/utils.py (or django/core/mail/message.py) and the call to make\_msgid(domain=DNS\_NAME) in d
  2. In the location where DNS\_NAME is used to generate the Message-ID, convert the domain to punycode (e.g., using `domain.encode('idna').decode('ascii')`
  3. Apply the punycode conversion in a way that handles both ASCII and non-ASCII hostnames gracefully (e.g., using `idna.encode(domain).decode('ascii')`).
  4. Review the code path to ensure the fix does not break when DNS\_NAME is already ASCII or when encoding is unicode.
  5. Check for any other places in django/core/mail that use DNS\_NAME or generate headers with the hostname and apply the same punycode conversion if neede

--- Example 4 ---
Task:    Allow calling reversed() on an OrderedSet Description Currently, OrderedSet isn't reversible (i.e. allowed to be passed as an argument to Python's reversed()). This would be natural to support given that OrderedSet is ordered. This shou
Best mode: SEQUENTIAL
Plan it produced (one step at a time, each chosen after seeing the last result):
  1. Locate the OrderedSet class definition (likely in a file like ordered\_set.py or similar) using grep or find.
  2. Read the surrounding code of the OrderedSet class to identify the internal storage (e.g., self.items list, self.map dict, or similar).
  3. Add a \_\_reversed\_\_ method that returns an iterator over the stored items in reverse order (e.g., iter(self.items[::-1]) or reversed(self.items)).
  4. Verify the method is syntactically correct by reading the modified file (e.g., check indentation and no stray characters).

--- Example 7 ---
Task:    Using multiple FilteredRelation with different filters but for same relation is ignored. Description (last modified by lind-marcus) I have a relation that ALWAYS have at least 1 entry with is\_all=True and then I have an optional ent
Best mode: HIERARCHICAL
Plan it produced (top-level phases, each decomposed into concrete actions only once that phase is reached -- the nesting IS the mode):
  1. Reproduce the bug with a minimal test case
       1.1 Locate the buggy code and existing tests by searching for relevant keywords and reading key files.
       1.2 Create a minimal test file that triggers the bug.
       1.3 Execute the test to verify failure.
  2. Locate the root cause in Django ORM join building
       2.1 Search for and read the Django ORM source files responsible for join construction, specifically `django/db/models/sql/query.py` and `django/db/models/sql/compiler.py`.
       2.2 Trace the join-building code path to understand how join conditions are generated, focusing on the `ON` clause construction.
       2.3 Pinpoint the specific line or condition where the join logic deviates from expected behavior.
  3. Implement the fix to allow multiple FilteredRelation for same relation
       3.1 Search for the code that handles FilteredRelation alias generation in the Django ORM query module.
       3.2 Read the relevant code section and the existing test file to understand the current alias logic.
       3.3 Modify the alias assignment logic to ensure unique aliases when multiple FilteredRelation objects reference the same relation, then run the related test suite.
  4. Verify the fix with tests and check for regressions
       4.1 Run the specific unit test(s) that cover the fixed bug to confirm they pass.
       4.2 Run the full test suite to check for regressions.

--- Example 8 ---
Task:    Management command subparsers don't retain error formatting Description Django management commands use a subclass of argparse.ArgumentParser, CommandParser, that takes some extra arguments to improve error formatting. These arguments are
Best mode: SEARCH
Plan it produced (competing whole-task routes, tried and then compared):
  1. Override add\_subparsers in CommandParser to return custom SubParsersAction that creates CommandParser subparsers with same kwargs
  2. Modify CommandParser.\_\_init\_\_ to store formatting kwargs and propagate them when subparsers are created
  3. Patch the subparsers action class so each added parser inherits the parent's error-formatting arguments

  [... examples 2, 3, 5, 6 omitted; same shape ...]

Task: $<$the GitHub issue text for the task being ranked$>$
\end{promptbox}

\end{document}